\documentclass{article}
\usepackage{PRIMEarxiv}
\usepackage[utf8]{inputenc}
\usepackage[T1]{fontenc}
\usepackage{hyperref}
\hypersetup{hidelinks=true}
\usepackage{url}
\usepackage{booktabs}
\usepackage{amsmath,amssymb,amsfonts}
\usepackage{nicefrac}
\usepackage{microtype}
\usepackage{fancyhdr}
\usepackage{graphicx}
\usepackage{longtable}
\usepackage{multirow}
\usepackage{array}
\usepackage{fvextra}
\usepackage{textcomp}
\usepackage{enumitem}
\graphicspath{{./}}
\DefineVerbatimEnvironment{PromptBlock}{Verbatim}{
  breaklines=true,
  breakanywhere=true,
  fontsize=\small
}
\numberwithin{equation}{section}

\title{Medical Heuristic Learning: An LLM-Driven Framework for Interpretable and Auditable Clinical Decision Rules}
\author{\parbox{0.94\textwidth}{\centering
\normalfont\bfseries Wei Xu$^{1,\dagger}$, Ke Yang$^{2,*\dagger}$, Gang Luo$^{1}$, Keli Zheng$^{3}$,  Lingyan Hu$^{4}$, Jing Wang$^{5,*}$, and Kefeng Li$^{1,*}$\\[0.5em]
\normalfont\footnotesize $^{1}$The Centre for Artificial Intelligence Driven Drug Discovery, Macao Polytechnic University, Macao SAR\\
\normalfont\footnotesize $^{2}$The Key Laboratory of Short-Range Radio Equipment Testing and Evaluation, Ministry of Industry and Information Technology,\\
\normalfont\footnotesize and the Terahertz Science Application Center (TSAC), Beijing Institute of Technology, Zhuhai, China\\
\normalfont\footnotesize $^{3}$The Faculty of Education, The University of Hong Kong, Hong Kong SAR\\
\normalfont\footnotesize $^{4}$The College of Information Engineering, Dalian University, Dalian, Liaoning, China\\
\normalfont\footnotesize $^{5}$The Department of Critical Care Medicine, Yantai Yuhuangding Hospital, Qingdao University,\\
\normalfont\footnotesize Yantai, Shandong, China\\[0.5em]
\normalfont\footnotesize $^{\dagger}$Wei Xu and Ke Yang contributed equally to this work.\\
\normalfont\footnotesize $^{*}$Correspondence: kefengl@mpu.edu.mo; jinga@qdu.edu.cn; yangke\_1208@163.com}}

\begin{document}
\maketitle

\begin{abstract}
Predictive modeling for clinical decision support requires both strong predictive performance and transparent, auditable, and human-reviewable decision logic. Although deep learning and tree-based ensemble methods can achieve high accuracy, their black-box nature remains a major obstacle to trustworthy clinical deployment. Moreover, clinical prediction often operates under practical constraints, including limited sample sizes, severe class imbalance, and feature evolution arising from changes in diagnostic criteria or clinical documentation practices. We propose Medical Heuristic Learning (MHL), a constrained paradigm for LLM-assisted rule learning. Rather than relying on updates to implicit model weights, MHL integrates statistical probes, medical knowledge probes, initial rule synthesis, and iterative rule optimization to construct an executable rule-based expert system. The resulting rule system is expressed entirely using the native logical and control-flow constructs of a programming language. Valid rule versions are recorded and retained along the search trajectory, making the decision logic explicit, interpretable, and auditable. MHL also supports continual learning by using previously validated rules as a starting point and iteratively revising them in response to updated feature information under data drift or feature evolution. MHL is not tied to any specific programming language. In this study, we instantiate it in Python and release the implementation on PyPI as a ready-to-use software package. Comprehensive experiments on medical datasets show that MHL achieves predictive performance comparable to that of state-of-the-art methods, performs favorably in small-sample and highly imbalanced settings, and supports the transfer and adaptive revision of validated rules under feature evolution. Overall, these findings suggest that non-gradient-based heuristic systems offer an approach to balancing predictive performance and transparency in clinical decision support. The PyPI package can be installed via \texttt{pip install medical-heuristic-learning}, and the source code is publicly available at \url{https://github.com/MPU-Li-OmicsLab/medical-heuristic-learning}.
\end{abstract}

\section{Introduction}
\label{sec:1}

Machine learning models are increasingly being explored for clinical decision support \cite{alowais2023revolutionizing}. In many applications, they promise earlier risk stratification, more efficient allocation of clinical resources, and timelier intervention. In medicine, however, predictive accuracy alone is insufficient. Clinical decisions are made in high-stakes settings, where erroneous predictions may lead to delayed treatment, inappropriate escalation of care, or unnecessary intervention. Clinicians therefore cannot reasonably rely on systems whose decision processes are opaque. Simpler models, such as linear models or shallow decision trees, are easier to inspect, but they are often perceived as less competitive on complex tasks. More intrinsically interpretable models can improve transparency, but their modeling capacity is still often constrained when complex nonlinear structure or heterogeneous interactions must be captured. By contrast, deep neural networks, ensemble tree models, and other black-box approaches often deliver stronger predictive performance, yet their post-hoc explanation tools, including SHAP and LIME, remain sensitive to modeling choices and input perturbations, limiting their ability to provide stable and clinically reliable insight \cite{lundberg2017unified,ribeiro2016trust,rudin2019stop,tjoa2021survey}. Consequently, clinical deployment continues to face a persistent trade-off. Models with strong predictive performance are often difficult to audit, whereas models that permit effective human oversight often struggle to maintain competitive performance.

Beyond interpretability, current high-performing models remain vulnerable to the statistical and temporal complexity of clinical data. Medical datasets are often small, imbalanced, and heterogeneous rather than large, clean, and balanced. Under such conditions, methods that rely mainly on gradient-based optimization or greedy partitioning may be sensitive to data limitations and unstable empirical regularities. Clinical practice is also inherently dynamic. As medical knowledge advances and diagnostic technologies evolve, both the set of measured variables and the criteria used to define disease may change over time \cite{kelly2019key,singer2016sepsis3}. When feature evolution occurs, model updating becomes not only a retraining problem but also a knowledge-maintenance problem. Previously validated decision logic must be revised to accommodate new variables and retired measurements without losing its auditability. These difficulties motivate the pursuit of a learning framework that can jointly support high predictive performance, strong adaptability, and high transparency.

To address these challenges, we introduce the learning beyond gradients paradigm, namely Heuristic Learning, into clinical prediction. Traditional heuristic algorithms typically perform iterative search in a candidate solution space through empirical rules or bio-inspired mechanisms to identify efficient and feasible solutions without relying on gradient information, as exemplified by particle swarm optimization and ant colony optimization \cite{kennedy1995particle,dorigo1996ant}. Heuristic Learning extends this perspective to executable rule-program optimization. It uses a large language model (LLM) to draft, revise, and evolve candidate rule modules, thereby shifting the optimization target from a numerical parameter space to an explicit program space \cite{weng2026learning_beyond_gradients}. Under this paradigm, the learned object is not a black-box parameter vector but an executable rule system whose structure can be inspected, tested, revised, and retained throughout the search trajectory. Heuristic Learning therefore provides a natural foundation for studying white-box predictive modeling as constrained search over discrete programs.

Building on this paradigm, we propose Medical Heuristic Learning (MHL), an LLM-assisted framework for learning explicit rule programs from structured medical data. Unlike earlier heuristic-learning explorations that relied heavily on open-ended coding agents, MHL is organized as a constrained and auditable workflow tailored to medical prediction, integrating two complementary evidence sources in which a statistical probe extracts distributional summaries and candidate predictive signals from the training data, whereas a medical knowledge probe derives clinically grounded priors and threshold suggestions from variable semantics. Conditioned on these inputs, the LLM generates an initial executable rule program and then revises it through small code-level edits guided by error and degradation feedback. In algorithmic terms, MHL operates as a trajectory-based single-solution metaheuristic over a discrete rule-program space, with prioritized validation criteria determining the returned program from the valid search trajectory. When feature evolution occurs, MHL reuses the previously validated rule program as a warm-start blueprint and updates it with renewed evidence, so adaptation takes the form of explicit rule maintenance rather than hidden-parameter transfer.

We evaluate MHL on multiple medical datasets, including UK Biobank (UKB)~\cite{sudlow2015ukbiobank}, the Critical Care Information Database (CCID), and the Medical Information Mart for Intensive Care (MIMIC)~\cite{johnson2016mimic}. The experiments show that MHL remains competitive with strong baseline learners overall while preserving explicit white-box decision logic. Its clearest empirical advantages appear in small-sample and severely imbalanced settings, where reliable rule induction is especially difficult, and in a simulated continual-adaptation setting involving feature evolution, where explicit inheritance and revision of validated rules improve adaptation to the new feature space. Taken together, these results position MHL as a search-based white-box alternative to conventional black-box learning for structured medical prediction, linking executable knowledge representation, constrained program generation, single-solution heuristic search, and explicit rule maintenance within one framework.

\section{Related Work}
\label{sec:2}

\subsection{Clinical Prediction and Interpretability}
\label{sec:2-1}

Clinical prediction and medical diagnostic modeling have long been shaped primarily by the pursuit of predictive performance. Early approaches relied on statistical models such as linear and logistic regression, together with decision trees, to relate structured clinical variables to diagnostic or prognostic outcomes \cite{cox1958regression,breiman1984classification}. As data availability and computational capacity increased, more complex methods emerged. Large boosted-tree ensembles such as XGBoost and LightGBM, as well as deep models for structured prediction, can capture nonlinear effects and high-order interactions that simpler models may miss \cite{chen2016xgboost,ke2017lightgbm,gorishniy2021revisiting}. This progression has expanded predictive capacity, but it has also made the learned decision process increasingly difficult to inspect directly.

From an interpretability perspective, many early models already provide intrinsic transparency. Sparse regression models expose variable contributions through their coefficients and link functions, while shallow decision trees express predictions as traceable paths of feature tests \cite{cox1958regression,breiman1984classification}. For complex models whose internal structures are not directly inspectable, explainable artificial intelligence has instead focused largely on post-hoc analysis. Methods such as LIME and SHAP seek to explain why a model produces a particular prediction for a given instance by estimating the local influence or contribution of each input feature through perturbation analysis or feature attribution \cite{ribeiro2016trust,lundberg2017unified}. Such post-hoc explanations can provide useful information about model behavior, but because they are generated after an opaque model has produced its output rather than exposing the underlying decision mechanism, explanation quality and stability do not necessarily translate into clinical usefulness \cite{tjoa2021survey,kelly2019key}.

These limitations have motivated the argument that intrinsically interpretable models should be preferred to post-hoc explanations of opaque systems, particularly in high-stakes applications \cite{rudin2019stop}. Generalized additive models (GAMs), GAMs with pairwise interactions (GA2Ms), Explainable Boosting Machines (EBMs), and Automatic Piecewise Linear Regression (APLR) provide more flexible forms of intrinsic interpretability \cite{lou2013ga2m,vonottenbreit2024automatic}. These models expose predictions through inspectable feature effects, optionally limited interactions, or piecewise linear components, allowing nonlinear relationships to be modeled while preserving substantial global transparency. Nevertheless, intrinsic interpretability alone does not ensure that the complete predictor is expressed as executable and maintainable decision logic. Coefficients, additive shape functions, piecewise linear components, and tree paths are inspectable, but they are not necessarily organized as versioned rule artifacts that can be validated and revised. This indicates a need to learn transparent predictors while preserving their rules as explicit and maintainable program structures.

\subsection{Rule-Based Clinical Reasoning}
\label{sec:2-2}

Early medical expert systems and subsequent clinical decision support systems established the importance of explicit knowledge representation and traceable reasoning in healthcare. MYCIN is a representative early example, showing how medical inference can be organized through explicit expert rules rather than implicit model parameters \cite{shortliffe1975inexact_reasoning}. Later research on clinical decision support further emphasized inspectable inference processes and integration with clinical workflows \cite{middleton2016cds_vision,sutton2020cdss_overview}. Taken together, this line of work established explicit rules and traceable inference as central design principles for clinical decision support. At the same time, many early systems relied heavily on manually specified knowledge bases, which constrained their scalability and adaptability in modern data-rich clinical settings.

Subsequent work on rule learning provided a formal framework for interpretable prediction by representing decisions as explicit rule structures. Representative studies such as Bayesian Rule Lists and CORELS show that compact rule lists can remain competitive while being substantially easier to inspect than many black-box alternatives \cite{letham2015bayesian_rule_lists,angelino2018corels}. These studies demonstrate that rules can function as predictive models rather than merely as explanatory surrogates. Nevertheless, most rule learning research still treats rules primarily as the endpoint of classification, with limited attention to how they are grounded in multiple evidence sources, iteratively revised, or maintained as versioned reasoning artifacts. What remains underdeveloped is a modern rule-based expert system workflow that combines data-driven construction with executable and auditable expert rule maintenance.

\subsection{LLMs for Rule Induction}
\label{sec:2-3}

The rise of large language models has created new opportunities for knowledge-guided reasoning, code synthesis, and programmatic inference. Transformer architectures enabled large-scale contextual representation learning, while advances in few-shot generalization substantially expanded the range of tasks that language models can address \cite{vaswani2017attention,brown2020few_shot}. In parallel, code-oriented language models demonstrated that executable programs can be synthesized from natural-language specifications \cite{chen2021llm_code}, and reasoning-and-acting frameworks such as ReAct showed that language models can support multi-step inference rather than only one-shot text generation \cite{yao2023react}. Together, these developments suggest that LLMs can serve not only as predictors but also as intermediaries that translate statistical evidence and domain knowledge into executable decision structures.

Much of the existing literature nevertheless uses LLMs to produce answers, explanatory text, or code fragments rather than to construct and maintain white-box rule systems \cite{chen2021llm_code,yao2023react}. In clinical settings, this distinction is important because deployment concerns extend beyond predictive correctness on isolated cases to accountability, verification, and sustainable maintenance \cite{kelly2019key}. Thus, although LLMs provide an important technical basis for knowledge-guided rule induction, prior work has not yet established a workflow in which LLMs help construct executable expert rules for clinical reasoning under explicit evidence constraints.

\subsection{Heuristics and Heuristic Learning}
\label{sec:2-4}

The idea of improving systems through search in a structured solution space predates modern gradient-based learning. Metaheuristic search can be broadly divided into single-solution and population-based strategies. Single-solution methods, exemplified by simulated annealing, iteratively modify one incumbent solution and use controlled stochastic acceptance to escape local optima \cite{kirkpatrick1983simulated_annealing}. Population-based methods, including particle swarm optimization and ant colony optimization, coordinate multiple candidate solutions or search agents during optimization \cite{kennedy1995particle,dorigo1996ant}. Taken together, these methods exemplify the broader metaheuristic view that candidate solutions can be iteratively improved through gradient-free exploration, selection, and adaptation \cite{blum2003metaheuristics}. Related work on evolutionary program synthesis further shows that executable programs themselves can be searched and optimized as the primary carriers of behavior \cite{sobania2023program_synthesis_survey}. Collectively, these studies provide a conceptual basis for learning in program space rather than only in parameter space.

Heuristic Learning advances this line of work into a distinct learning paradigm in which executable systems are iteratively refined rather than hidden weights repeatedly adjusted. In contrast to the gradient-based paradigm exemplified by neural representation learning \cite{rumelhart1986learning}, Learning Beyond Gradients proposes that learning can operate directly on programs, rules, and procedures that remain open to inspection and revision \cite{weng2026learning_beyond_gradients}. This perspective is well suited to clinical prediction and medical decision modeling, where interpretability, traceability, and maintainability may be as important as predictive performance. To the best of our knowledge, however, Heuristic Learning has not yet been instantiated in medical applications, particularly as an integrated framework that combines statistical evidence, knowledge-guided reasoning, initial rule synthesis, feedback-guided rule revision, version-aware maintenance, and continual adaptation under feature evolution. The framework proposed in this study is designed to address this gap.

\section{Methods}
\label{sec:3}

This section presents Medical Heuristic Learning (MHL) in detail, encompassing its formal mathematical formulation and the construction of its operational workflow. Section~\ref{sec:3-1} defines the classification problem and formulates constrained optimization over the rule-program space. Section~\ref{sec:3-2} presents the overall MHL workflow. Sections~\ref{sec:3-3} and~\ref{sec:3-4} explain how the initial statistical and medical knowledge probes acquire statistical evidence and LLM-generated medical-prior evidence, respectively. Sections~\ref{sec:3-5} and~\ref{sec:3-6} then formulate constrained initialization, feedback-guided program revision, version retention, and lexicographic selection. Section~\ref{sec:3-7} extends MHL to feature evolution. Section~\ref{sec:3-8} describes the Python implementation of MHL, and Section~\ref{sec:3-9} illustrates a representative end-to-end execution of MHL.

\subsection{Problem Definition}
\label{sec:3-1}

\subsubsection{Classification Task and Rule Program}
\label{sec:3-1-1}

Consider a classification dataset with \(m\) samples and \(d\) features:

\begin{equation}
\mathcal D
=
\left\{
(\mathbf x_i,y_i)
\right\}_{i=1}^{m}.
\end{equation}

Each sample comprises a feature vector and a class label:

\begin{equation}
\mathbf x_i
=
(x_{i1},x_{i2},\ldots,x_{id})
\in\mathcal X,
\qquad
y_i\in\mathcal Y.
\end{equation}

The joint feature space is

\begin{equation}
\mathcal X
=
\mathcal X_1
\times
\mathcal X_2
\times\cdots\times
\mathcal X_d
=
\prod_{j=1}^{d}\mathcal X_j,
\end{equation}

where \(\mathcal X_j\) is the domain of the \(j\)-th feature. Individual features may be continuous or categorical. The class space is

\begin{equation}
\mathcal Y
=
\{c_1,c_2,\ldots,c_C\},
\qquad
C\geq2.
\end{equation}

This formulation covers general multiclass classification. Without loss of generality, the experiments in this study consider the binary special case \(C=2\).

Rather than representing the learned classifier only through implicit parameters, we seek an explicit rule program \(P\). Executing this program defines the classification function

\begin{equation}
h_P:\mathcal X\rightarrow\mathcal Y.
\end{equation}

and produces the prediction

\begin{equation}
\widehat y_i
=
h_P(\mathbf x_i).
\end{equation}

Accordingly, \(P\) denotes the concrete rule structure that can be generated, inspected, modified, and stored, whereas \(h_P\) denotes the input--output relation induced by its execution. This distinction separates the program representation of the learned rule system from its prediction semantics.

To make this abstract program representation constructive, we next define the primitive predicates from which rule conditions are formed.

\subsubsection{Atomic Feature Predicates}
\label{sec:3-1-2}

We first define the atomic feature-predicate set \(\Phi\), whose elements are elementary Boolean tests on individual features. The Boolean domain is denoted by

\begin{equation}
\mathbb B=\{0,1\}
\end{equation}

where \(1\) indicates that a condition is satisfied and \(0\) indicates that it is not. An atomic feature predicate applies one elementary test to one feature and defines a Boolean function:

\begin{equation}
\phi:
\mathcal X
\rightarrow
\mathbb B.
\end{equation}

The feature indices are partitioned into continuous and categorical sets:

\begin{equation}
\mathcal J_{\mathrm{con}}
\cup
\mathcal J_{\mathrm{cat}}
=
\{1,2,\ldots,d\}.
\end{equation}

For a continuous feature \(x_j\), the predicate takes the form

\begin{equation}
\phi_{j,\bowtie,a}(\mathbf x)
=
\mathbb I[x_j\bowtie a],
\qquad
j\in\mathcal J_{\mathrm{con}},
\end{equation}

where

\begin{equation}
\bowtie\in\{<,\leq,=,\neq,\geq,>\},
\qquad
a\in\mathbb R,
\end{equation}

and \(\mathbb I[\cdot]\) is the indicator function. For a categorical feature \(x_j\), the corresponding predicate is

\begin{equation}
\phi_{j,A}(\mathbf x)
=
\mathbb I[x_j\in A],
\qquad
j\in\mathcal J_{\mathrm{cat}},
\quad
A\subseteq\mathcal X_j.
\end{equation}

When \(A\) contains a single category, the membership test reduces to equality. Collecting both predicate types gives

\begin{equation}
\Phi
=
\left\{
\phi_{j,\bowtie,a}
\right\}
\cup
\left\{
\phi_{j,A}
\right\}.
\end{equation}

Some typical examples take the following forms:

\begin{equation}
x_{\mathrm{age}}\geq65,
\qquad
x_{\mathrm{SBP}}<90,
\qquad
x_{\mathrm{diagnosis}}
=1
\end{equation}

each of which evaluates a single feature.

\subsubsection{Boolean Condition Expressions}
\label{sec:3-1-3}

Rule programs require conditions that can combine multiple atomic tests. We therefore construct a Boolean condition language \(\Psi\) from the atomic-predicate set \(\Phi\). First, let

\begin{equation}
\Phi^{\pm}
=
\Phi
\cup
\left\{
\neg\phi
\mid
\phi\in\Phi
\right\}.
\end{equation}

The condition language \(\Psi\) is defined as the smallest set satisfying

\begin{equation}
\Phi^{\pm}\subseteq\Psi,
\end{equation}

and, for any \(\psi_1,\psi_2\in\Psi\),

\begin{equation}
(\psi_1\land\psi_2)\in\Psi,
\qquad
(\psi_1\lor\psi_2)\in\Psi.
\end{equation}

This construction ensures that every expression \(\psi\in\Psi\) has Boolean semantics:

\begin{equation}
\psi:
\mathcal X
\rightarrow
\mathbb B,
\qquad
\psi\in\Psi.
\end{equation}

In particular, for a continuous feature \(x_j\), any bounds \(a,b\in\mathbb R\) with \(a<b\) define the interval condition

\begin{equation}
a\leq x_j<b,
\qquad
j\in\mathcal J_{\mathrm{con}},
\end{equation}

which is represented in \(\Psi\) by the conjunction of two atomic predicates:

\begin{equation}
(x_j\geq a)
\land
(x_j<b).
\end{equation}

Some typical composite conditions take the following forms:

\begin{equation}
(x_{\mathrm{age}}\geq65)
\land
(x_{\mathrm{SBP}}<90),
\qquad
55<x_{\mathrm{age}}<65.
\end{equation}

The first jointly evaluates predicates on two different features, whereas the second jointly evaluates lower and upper bounds on the same feature. Under the construction above, the latter is equivalent to \((x_{\mathrm{age}}>55)\land(x_{\mathrm{age}}<65)\). These examples illustrate how \(\Psi\) supports joint evaluation through the composition of distinct atomic predicates.

\subsubsection{Conditional-Branch Rule Programs}
\label{sec:3-1-4}

\begin{figure}[t]
\centering
\includegraphics[width=\textwidth]{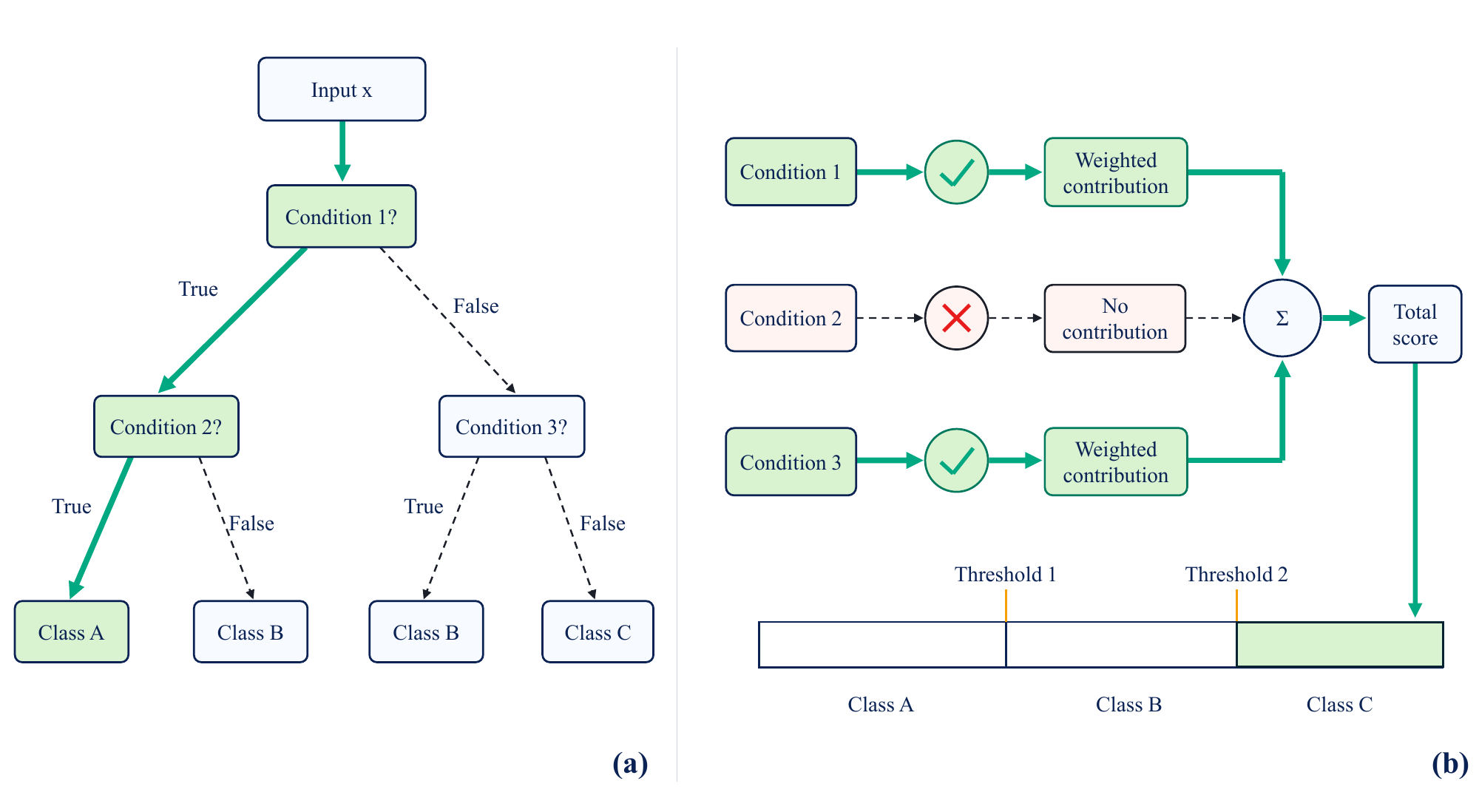}
\caption{Execution semantics of the two rule-program families: (a) a conditional-branch program follows one root-to-leaf path determined by successive Boolean outcomes; (b) an additive scoring program aggregates the weighted contributions of satisfied conditions and maps the total score to a class using ordered thresholds.}
\label{fig:results-3-1-4-rule-program-families}
\end{figure}

We consider two rule-program families: conditional-branch programs \(\mathcal P^{\mathrm{br}}\) and additive scoring programs \(\mathcal P^{\mathrm{sc}}\). As illustrated in Figure \ref{fig:results-3-1-4-rule-program-families}, conditional-branch programs route each input along a single root-to-leaf path selected by successive Boolean outcomes and return the class at the terminal leaf (subfigure (a)), whereas additive scoring programs sum the weights associated with satisfied conditions and map the total score to a class through ordered thresholds (subfigure (b)). This subsection formalizes \(\mathcal P^{\mathrm{br}}\). Section~\ref{sec:3-1-5} defines \(\mathcal P^{\mathrm{sc}}\).

A conditional-branch program

\begin{equation}
P\in\mathcal P^{\mathrm{br}}
\end{equation}

is represented as a finite rooted binary decision tree. Its internal decision nodes and leaves form disjoint sets:

\begin{equation}
V_P
=
V_P^{\mathrm{dec}}
\cup
V_P^{\mathrm{leaf}},
\qquad
V_P^{\mathrm{dec}}
\cap
V_P^{\mathrm{leaf}}
=
\varnothing.
\end{equation}

Each internal node \(v\in V_P^{\mathrm{dec}}\) is associated with a condition

\begin{equation}
\psi_v\in\Psi.
\end{equation}

Each internal node has true and false outgoing branches. Given \(\mathbf x\), execution follows the true branch when

\begin{equation}
\psi_v(\mathbf x)=1,
\end{equation}

and the false branch when

\begin{equation}
\psi_v(\mathbf x)=0.
\end{equation}

The condition therefore determines the node outcome, while the selected branch determines the subsequent execution path. Each leaf \(u\in V_P^{\mathrm{leaf}}\) carries a class label

\begin{equation}
c_u\in\mathcal Y.
\end{equation}

Execution begins at the root and evaluates the conditions along a single path. Because the tree is finite, every input reaches a unique leaf. Denote this leaf by

\begin{equation}
u_P(\mathbf x)
\in
V_P^{\mathrm{leaf}}
\end{equation}

The prediction defined by \(P\) is then

\begin{equation}
h_P(\mathbf x)
=
c_{u_P(\mathbf x)},
\end{equation}

which gives, for sample \(i\),

\begin{equation}
\widehat y_i
=
h_P(\mathbf x_i)
=
c_{u_P(\mathbf x_i)}.
\end{equation}

In executable form, internal nodes map naturally to native conditional statements and leaves to return statements. Nested nodes can express general binary trees, nested conditionals, and ordered decision lists. Importantly, this structural representation does not imply that the program was learned by a conventional decision-tree algorithm.

\subsubsection{Additive Scoring Rule Programs}
\label{sec:3-1-5}

The second family, \(\mathcal P^{\mathrm{sc}}\), contains additive scoring programs. As illustrated in Figure \ref{fig:results-3-1-4-rule-program-families}(b), these programs do not select a single execution path. Instead, every satisfied condition contributes its associated weight, the active contributions are summed, and the resulting total is located within a threshold-defined class interval. Formally, each program combines \(N_{\mathrm{sc}}\) condition--weight pairs with classification thresholds and a mapping from score intervals to classes:

\begin{equation}
P
=
\left(
\left\{
(\psi_r,w_r)
\right\}_{r=1}^{N_{\mathrm{sc}}},
\boldsymbol{\tau},
\boldsymbol c
\right)
\in
\mathcal P^{\mathrm{sc}},
\end{equation}

where

\begin{equation}
N_{\mathrm{sc}}\in\mathbb N_+,
\qquad
\psi_r\in\Psi,
\qquad
w_r\in\mathbb R.
\end{equation}

Here, \(N_{\mathrm{sc}}\) is the number of scoring conditions, and \(\psi_r\) may represent either a feature interval or a composite expression involving multiple features. The contribution of condition \(r\) is

\begin{equation}
q_r(\mathbf x)
=
w_r\psi_r(\mathbf x).
\end{equation}

Summing these contributions yields

\begin{equation}
s_P(\mathbf x)
=
\sum_{r=1}^{N_{\mathrm{sc}}}
w_r\psi_r(\mathbf x).
\end{equation}

Thus, every satisfied condition contributes to the final score. Scoring conditions may be mutually exclusive or overlapping. In the latter case, all activated terms contribute to the sum.

For a \(C\)-class task, the vector

\begin{equation}
\boldsymbol c
=
(c_{(1)},c_{(2)},\ldots,c_{(C)})
\end{equation}

specifies the class assigned to each score interval, while

\begin{equation}
\boldsymbol{\tau}
=
(\tau_1,\ldots,\tau_{C-1})
\end{equation}

contains the \(C-1\) finite thresholds. Together with \(\tau_0=-\infty\) and \(\tau_C=+\infty\), they satisfy

\begin{equation}
-\infty
=
\tau_0
<
\tau_1
<
\cdots
<
\tau_{C-1}
<
\tau_C
=
+\infty.
\end{equation}

The resulting classifier is

\begin{equation}
h_P(\mathbf x)
=
c_{(k)}
\quad\Longleftrightarrow\quad
\tau_{k-1}
\leq
s_P(\mathbf x)
<
\tau_k,
\qquad
k=1,\ldots,C.
\end{equation}

The ordering \((c_{(1)},\ldots,c_{(C)})\) refers only to the score-interval mapping and does not imply a natural clinical ordering of the classes. Binary classification requires only one threshold \(\tau\):

\begin{equation}
h_P(\mathbf x)
=
\begin{cases}
c_{-},
& s_P(\mathbf x)<\tau,\\[1mm]
c_{+},
& s_P(\mathbf x)\geq\tau.
\end{cases}
\end{equation}

This binary rule is precisely the \(C=2\) instance of the multiclass interval mapping.

\subsubsection{Unified Rule-Program Grammar and Program Space}
\label{sec:3-1-6}

The preceding definitions establish a common condition language for both program families. Atomic predicates in \(\Phi\) generate the Boolean expressions in \(\Psi\). Conditional-branch programs evaluate these expressions at internal nodes, whereas scoring programs associate them with numerical weights.

Let \(\Gamma\) collect the legal construction rules introduced in Sections~\ref{sec:3-1-2}--\ref{sec:3-1-5}. The induced rule-program space is

\begin{equation}
\mathcal P_\Gamma
=
\mathcal P^{\mathrm{br}}
\cup
\mathcal P^{\mathrm{sc}}.
\end{equation}

Every legal program

\begin{equation}
P\in\mathcal P_\Gamma,
\end{equation}

induces the classification function

\begin{equation}
h_P:\mathcal X\rightarrow\mathcal Y,
\end{equation}

and therefore predicts

\begin{equation}
\widehat y_i
=
h_P(\mathbf x_i).
\end{equation}

This abstraction specifies rule structure and execution semantics without committing to an implementation language. A legal program can therefore be instantiated using the native logical and control-flow constructs of any suitable general-purpose language.

\subsubsection{Rule-Program Optimization Objective}
\label{sec:3-1-7}

To accommodate multiple evaluation criteria, let

\begin{equation}
Q\in\mathbb N_+
\end{equation}

denote their number and define the associated cost functions

\begin{equation}
\ell_1(h_P;\mathcal D),
\ell_2(h_P;\mathcal D),
\ldots,
\ell_Q(h_P;\mathcal D),
\end{equation}

where \(\ell_q(h_P;\mathcal D)\in\mathbb R\), and every cost follows a common minimization orientation so that smaller values are always preferred.

Rather than collapsing the criteria into a weighted scalar, MHL assigns them the priority order

\begin{equation}
\ell_1
\succ
\ell_2
\succ
\cdots
\succ
\ell_Q,
\end{equation}

with \(\ell_1\) evaluated first. The corresponding ordered cost vector is

\begin{equation}
\mathbf L(P;\mathcal D)
=
\left(
\ell_1(h_P;\mathcal D),
\ell_2(h_P;\mathcal D),
\ldots,
\ell_Q(h_P;\mathcal D)
\right).
\end{equation}

Feasibility is represented by

\begin{equation}
C_{\mathrm{hard}}:
\mathcal P_\Gamma
\rightarrow
\{0,1\}
\end{equation}

which indicates whether a program satisfies the required syntax, input--output interface, execution, feature and operation restrictions, and output domain \(\mathcal Y\). For \(P\in\mathcal P_\Gamma\),

\begin{equation}
C_{\mathrm{hard}}(P)=1
\end{equation}

denotes a feasible program, whereas

\begin{equation}
C_{\mathrm{hard}}(P)=0
\end{equation}

indicates that at least one requirement is violated. Assuming that the feasible set is nonempty, the ideal objective is

\begin{equation}
\begin{aligned}
P^\dagger
\in
\operatorname*{arg\,lex\,min}_{P\in\mathcal P_\Gamma}
\quad&
\mathbf L(P;\mathcal D)
\\
\mathrm{s.t.}\quad&
C_{\mathrm{hard}}(P)=1.
\end{aligned}
\end{equation}

Lexicographic minimization first optimizes \(\ell_1\) and consults a lower-priority cost only when all preceding costs are tied. Consequently, improvement on a secondary criterion cannot compensate for deterioration on a higher-priority criterion.

\begin{figure}[t]
\centering
\includegraphics[width=\textwidth]{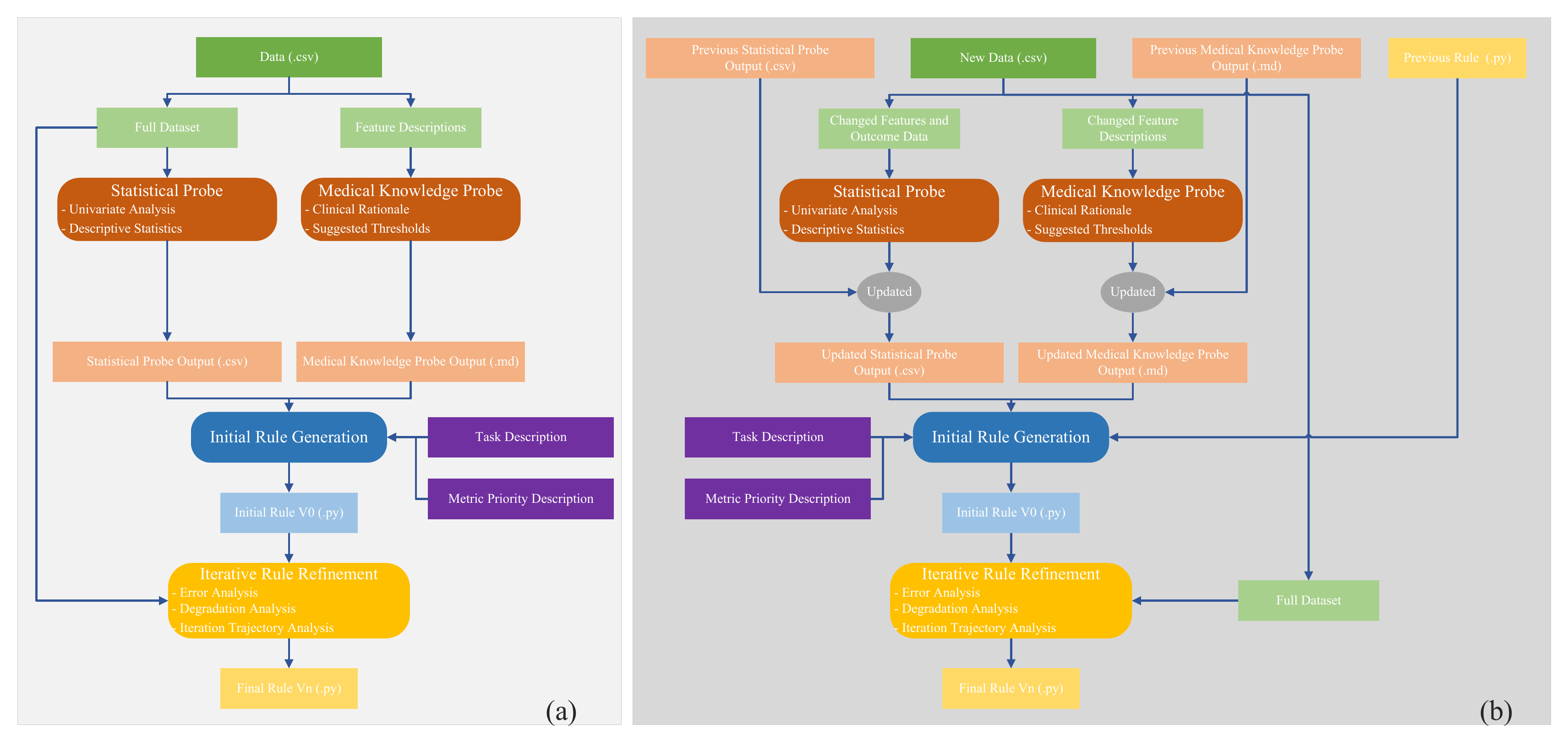}
\caption{Overview of the MHL framework: (a) the standard pipeline and (b) the continual-learning extension under feature evolution.}
\label{fig:results-2-1-system-architecture}
\end{figure}

\subsection{Overview of the MHL Framework}
\label{sec:3-2}

\(P^\dagger\) denotes an ideal solution over the full space of feasible rule programs. Medical Heuristic Learning (MHL) turns this objective into a practical solution procedure through constrained white-box search. Rather than exhaustively enumerating \(\mathcal P_\Gamma\), MHL records a finite sequence of feasible programs generated over successive iterations and selects \(P^\star\) from this record according to the prescribed cost priorities. The resulting \(P^\star\) constitutes a modern rule-based expert system for clinical prediction from structured medical data. As shown in Figure \ref{fig:results-2-1-system-architecture}, subfigure (a) presents the standard MHL pipeline, whereas subfigure (b) presents its continual-learning extension under feature evolution.

To specify how this workflow operates, we first distinguish the data and textual inputs used during learning. Following standard machine-learning practice, the dataset is partitioned into mutually disjoint training, validation, and test sets:

\begin{equation}
\mathcal D
=
\mathcal D_{\mathrm{tr}}
\mathbin{\dot\cup}
\mathcal D_{\mathrm{val}}
\mathbin{\dot\cup}
\mathcal D_{\mathrm{te}}.
\end{equation}

The three subsets have distinct responsibilities. The training set supplies data for statistical probing and provides sample-level error and degradation feedback during rule revision. During learning, the validation set evaluates each valid rule version as it is produced, thereby monitoring the version trajectory. After iteration, the prioritized validation criteria determine the selected program \(P^\star\). This is analogous to evaluating a deep model on a validation set throughout training and retaining the best-performing checkpoint. In MHL, validation results support monitoring and retrospective version selection but are not passed to the LLM as revision feedback. The test set remains held out until \(P^\star\) has been fixed and is then used once to estimate the generalization performance of \(h_{P^\star}\). These roles follow the standard machine-learning separation between model construction, model selection, and final evaluation.

In addition to the structured data, MHL receives three textual objects. The researcher-provided task description \(T_1\) specifies the clinical background, prediction objective, and necessary application context. The variable-information text \(T_2\) contains the feature and target column names and is automatically organized from the table schema and the designated target column. The evaluation-criteria text \(T_3\) specifies the evaluation metrics and their priority order as a natural-language description of \(\ell_1\succ\ell_2\succ\cdots\succ\ell_Q\). Section~\ref{sec:3-7} introduces the additional feature-evolution context required for continual learning.

The discrete rule-program space \(\mathcal P_\Gamma\) is structured, generally non-differentiable, and too large to enumerate. MHL therefore employs an LLM as a heuristic operator for generating and revising rule programs rather than as a black-box classifier. Once selected, the deterministic rule program itself performs inference without invoking the LLM. The standard workflow is

\begin{equation}
\begin{aligned}
S
&=
\operatorname{Probe}_{\mathrm{stat}}
(\mathcal D_{\mathrm{tr}}),\\
K
&=
\operatorname{Probe}_{\mathrm{know}}
(T_1,T_2),\\
(P_0,B_0)
&=
\operatorname{Init}_{\mathrm{LLM}}
(S,K,T_1,T_3),\\
\mathcal A_N
&=
\operatorname{Iter}_{\mathrm{LLM}}
(P_0,\mathcal D_{\mathrm{tr}},\mathcal D_{\mathrm{val}},T_1,T_3),\\
P^\star
&=
\operatorname{Select}_{\mathrm{lex}}
(\mathcal A_N,\mathcal D_{\mathrm{val}}).
\end{aligned}
\end{equation}

MHL begins by deriving two complementary forms of evidence from its structured and textual inputs. The statistical probe transforms \(\mathcal D_{\mathrm{tr}}\) into \(S\), a summary of empirical distributional patterns based on descriptive statistics and univariate associations. In parallel, the medical knowledge probe uses \(T_1\) and \(T_2\) to produce \(K\), which links feature semantics to clinically motivated interpretations and candidate thresholds. The initialization operator then conditions on \(S\), \(K\), the task description \(T_1\), and the prioritized evaluation criteria \(T_3\) to generate the initial rule program \(P_0\) and its design rationale \(B_0\). The knowledge captured by \(S\) and \(K\) persists through \(P_0\) and the subsequent program history, while bounded feedback on classification errors and performance degradation guides each revision.

Starting from \(P_0\), the iterative operator constructs the retained valid-version set

\begin{equation}
\mathcal A_N
=
\{P_0,P_1,\ldots,P_N\}.
\end{equation}

The programs in this retained set are generated along the ordered trajectory

\begin{equation}
P_0
\rightarrow
P_1
\rightarrow
\cdots
\rightarrow
P_N.
\end{equation}

The final output \(P^\star\) is the lexicographically best valid program selected from \(\mathcal A_N\) on \(\mathcal D_{\mathrm{val}}\). The rule programs produced by MHL are not restricted to any particular programming language. In this study, they are instantiated as executable Python rule functions.

This structure characterizes MHL as a single-solution metaheuristic. The program \(P_0\) is the initial solution, \(P_t\) is the current solution at iteration \(t\), and each successful iteration generates one valid successor \(P_{t+1}\). The trajectory need not improve monotonically, and \(\mathcal A_N\) records the generated solutions rather than representing a population whose members evolve in parallel or recombine. MHL is similar to single-solution metaheuristics such as simulated annealing \cite{kirkpatrick1983simulated_annealing} in that both begin with one solution, follow a single trajectory, and permit temporary fitness deterioration. Unlike simulated annealing, MHL uses neither a predefined neighborhood, a temperature schedule, nor probabilistic acceptance. Instead, an LLM conditioned on program history and bounded training feedback generates each successor, while retrospective validation-based selection over the retained versions determines the returned program.

Next, Sections~\ref{sec:3-3}--\ref{sec:3-6} describe the individual operators in turn, Section~\ref{sec:3-7} extends MHL to settings involving feature evolution and data drift, Section~\ref{sec:3-8} describes the Python implementation of MHL, and Section~\ref{sec:3-9} illustrates its end-to-end execution.

\subsection{Statistical Probe}
\label{sec:3-3}

The statistical probe extracts structured evidence from the training set for subsequent rule construction:

\begin{equation}
\begin{aligned}
S
&=
\operatorname{Probe}_{\mathrm{stat}}
\left(
\mathcal D_{\mathrm{tr}}
\right)
\\
&=
\left(
S_j
\right)_{j=1}^{d}.
\end{aligned}
\end{equation}

Specifically, for each feature \(x_j\), the probe constructs the feature-level record

\begin{equation}
S_j
=
\left(
x_j,
D_j,
A_j
\right),
\qquad
j=1,\ldots,d.
\end{equation}

Here, \(D_j\) summarizes the observed distribution of \(x_j\), whereas \(A_j\) summarizes its univariate association with the target. Their contents depend on the feature type and classification setting. Table \ref{tab:methods-4-1-1-statistical-probe-fields} lists the artifact fields reported for these summaries in this study, together with information on feature identity and type, data completeness, and ranking.

In the current implementation, records are ordered by predefined statistical criteria, and the highest-ranked entries are included in the LLM context. Thus, \(S\) supplies downstream rule synthesis with ranked candidate features, distributional summaries, and transparent univariate association signals.

\begin{table}[htbp]
\small
\centering
\caption{Fields exported by the statistical probe artifact.}
\label{tab:methods-4-1-1-statistical-probe-fields}
\begin{tabular}{>{\raggedright\arraybackslash}p{0.18\linewidth}>{\raggedright\arraybackslash}p{0.17\linewidth}>{\raggedright\arraybackslash}p{0.20\linewidth}>{\raggedright\arraybackslash}p{0.37\linewidth}}
\toprule
Category & Variable type & Field & Interpretation \\
\midrule
\multirow{13}{=}{\raggedright\arraybackslash Descriptive statistics} & \multirow{5}{=}{\raggedright\arraybackslash Continuous and discrete} & rank & Relevance-based rank order of the candidate feature in the exported probe table. \\
 &  & feature & Feature name. \\
 &  & feature\_type & Variable type assigned by the probe, such as continuous or binary. \\
 &  & n\_valid & Number of non-missing observations used for the feature. \\
 &  & missing\_rate & Proportion of missing values for the feature. \\
\cline{2-4}
 & \multirow{6}{=}{\raggedright\arraybackslash Continuous} & mean & Sample mean. \\
 &  & std & Sample standard deviation. \\
 &  & median & Sample median. \\
 &  & min & Minimum observed value. \\
 &  & max & Maximum observed value. \\
 &  & binned\_or\_q4\_rel\_to\_q1 & Optional quantile-based odds-ratio summary for higher bins relative to the first quartile. \\
\cline{2-4}
 & \multirow{2}{=}{\raggedright\arraybackslash Discrete} & n\_unique & Number of unique levels. \\
 &  & level\_counts & Per-level counts, typically stored as a serialized dictionary. \\
\cline{1-4}
\multirow{10}{=}{\raggedright\arraybackslash Univariate analysis} & \multirow{3}{=}{\raggedright\arraybackslash Continuous and discrete} & method & Statistical method selected for the feature, such as pointbiserial, mwu, or chi2. \\
 &  & statistic & Main test statistic returned by the selected method. \\
 &  & p\_value & Main p-value returned by the selected method. \\
\cline{2-4}
 & \multirow{5}{=}{\raggedright\arraybackslash Continuous} & direction & Direction of association when applicable. \\
 &  & pointbiserial\_r & Point-biserial correlation coefficient for continuous features in binary tasks. \\
 &  & pointbiserial\_p & P-value associated with the point-biserial correlation. \\
 &  & mwu\_u & Mann-Whitney U statistic. \\
 &  & mwu\_p & P-value associated with the Mann-Whitney U test. \\
\cline{2-4}
 & \multirow{2}{=}{\raggedright\arraybackslash Discrete} & chi2\_stat & Chi-square statistic for discrete features. \\
 &  & chi2\_p & P-value associated with the chi-square test. \\
\bottomrule
\end{tabular}
\end{table}

Nonlinear multivariate association analysis is omitted from the statistical probe because its interaction estimates can be unstable in small or class-imbalanced datasets, whereas relevant conditional structure can be proposed and inspected explicitly during rule synthesis. The probe likewise avoids hard feature selection because a feature with weak marginal association may still be informative in combination with others.

\subsection{Medical Knowledge Probe}
\label{sec:3-4}

Complementing the data-derived evidence, the medical knowledge probe elicits structured prior evidence from feature semantics and the task description:

\begin{equation}
\begin{aligned}
K
&=
\operatorname{Probe}_{\mathrm{know}}(T_1,T_2)
\\
&=
\operatorname{LLM}
\left(
\operatorname{Prompt}_{\mathrm{know}}(T_1,T_2)
\right)
\\
&=
\{K_j\}_{j=1}^{d}.
\end{aligned}
\end{equation}

The prompt combines a fixed instruction template with the task description \(T_1\) and the variable-information text \(T_2\), which supplies the feature and target column names. Specifically, for each feature \(x_j\), the probe produces a feature-level record

\begin{equation}
K_j
=
\left(
x_j,
u_j,
\xi_j,
\vartheta_j,
\kappa_j
\right),
\qquad
j=1,\ldots,d.
\end{equation}

Here, \(u_j\) summarizes the expected feature--outcome relationship, \(\xi_j\) records the stated medical rationale, \(\vartheta_j\) provides a candidate threshold or categorical trigger, and \(\kappa_j\) records the reported confidence level. The statistical and medical-prior evidence streams are generated independently: \(S\) and \(K\) remain separate rather than being reduced to a weighted score and are combined only when the initial rule is synthesized.

Table \ref{tab:methods-4-1-2-medical-knowledge-probe-fields} lists the fields generated by the probe.

\begin{table}[t]
\normalsize
\centering
\caption{Fields generated by the medical knowledge probe.}
\label{tab:methods-4-1-2-medical-knowledge-probe-fields}
\begin{tabular}{p{0.33\linewidth}p{0.56\linewidth}}
\toprule
Field & Function in downstream rule generation \\
\midrule
Feature & Identifies the candidate variable to be referenced in rule code. \\
Univariate signal (summary) & Provides an LLM-generated qualitative prior for the expected feature--outcome relationship, such as strong positive, weak to moderate, or mixed association. \\
Clinical rationale & Provides concise medical justification, including typical mechanisms, clinical context, or caveats when relevant. \\
Suggested threshold & Supplies a directly usable rule condition, which may be a numeric threshold or a categorical trigger such as Present (1). \\
Evidence confidence (high/medium/low) & Indicates the expected confidence level of the clinical prior during rule construction. \\
\bottomrule
\end{tabular}
\end{table}

The resulting table organizes medical knowledge elicited from the LLM as structured prior evidence for downstream rule generation, and Section~\ref{sec:3-9} presents an example.

\subsection{Constrained Initial Rule Generation}
\label{sec:3-5}

The two evidence streams jointly condition initial program synthesis:

\begin{equation}
\begin{aligned}
(P_0,B_0)
&=
\operatorname{Init}_{\mathrm{LLM}}
\left(
S,K,T_1,T_3
\right)\\
&=
\operatorname{LLM}
\left(
\operatorname{Prompt}_{\mathrm{init}}
\left(
S,K,T_1,T_3
\right)
\right).
\end{aligned}
\end{equation}

The initialization prompt combines fixed instructions for the model role, rule structure, generation procedure, and output contract with \(S\), \(K\), the task description \(T_1\), and the metric-priority description \(T_3\). Because \(S\) and \(K\) already encode the variable information supplied by \(T_2\), \(T_2\) is not passed separately. The two returned components serve complementary functions. \(P_0\in\mathcal P_\Gamma\) provides the initial solution for subsequent search, whereas \(B_0\) records its design rationale. After passing the required validity checks, \(P_0\) initializes the retained valid-version set

\begin{equation}
\mathcal A_0
=
\{P_0\}.
\end{equation}

Its prioritized validation cost is then recorded as

\begin{equation}
\mathbf L
\left(
P_0;
\mathcal D_{\mathrm{val}}
\right).
\end{equation}

The initial program may instantiate either program family defined in Sections~\ref{sec:3-1-4} and~\ref{sec:3-1-5}. The LLM therefore serves as a constrained program-synthesis operator, whereas \(P_0\) is the executable classifier used for prediction.

The framework-level output contract is language-agnostic. Initialization returns an executable program \(P_0\) together with its rationale \(B_0\). In the present implementation, a candidate enters \(\mathcal A_0\) only after passing implementation-level checks that operationalize part of \(C_{\mathrm{hard}}\) without exhaustively verifying every feature reference, execution path, or medical rationale.

\subsection{Feedback-Guided Rule-Program Update and Selection}
\label{sec:3-6}

Starting from \(P_0\), MHL learns by revising the explicit rule program itself. The iterative operator builds a retained valid-version set, and validation-based lexicographic selection then returns the final program:

\begin{equation}
\begin{aligned}
\mathcal A_N
&=
\operatorname{Iter}_{\mathrm{LLM}}
\left(
P_0,
\mathcal D_{\mathrm{tr}},
\mathcal D_{\mathrm{val}},
T_1,
T_3
\right),\\
P^\star
&=
\operatorname{Select}_{\mathrm{lex}}
\left(
\mathcal A_N,
\mathcal D_{\mathrm{val}}
\right).
\end{aligned}
\end{equation}

Let \(P_t\) denote the \(t\)-th valid program in the retained set, with

\begin{equation}
h_t
=
h_{P_t},
\qquad
t=0,1,\ldots.
\end{equation}

\begin{figure}[t]
\centering
\includegraphics[width=\textwidth]{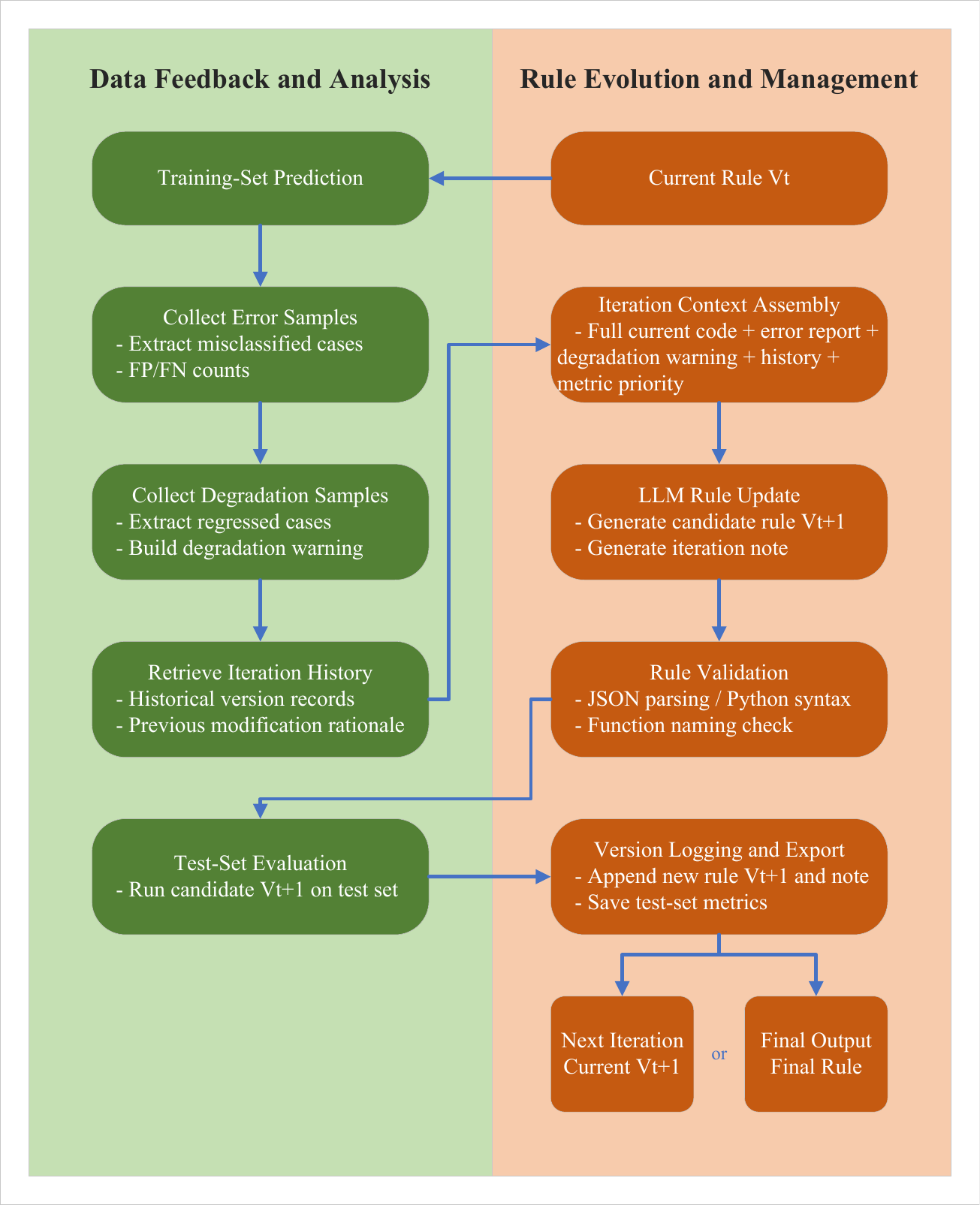}
\caption{Detailed swimlane view of the rule-program update stage in MHL, showing error analysis, degradation detection, proposal generation, validity checking, and version recording.}
\label{fig:methods-4-1-4-rule-iteration}
\end{figure}

At each round, training feedback and the complete program history condition the proposal of one successor. Validation costs are recorded for comparison across retained versions and do not impose round-by-round improvement. Figure \ref{fig:methods-4-1-4-rule-iteration} summarizes the interaction between feedback-guided revision and version management.

\subsubsection{Error and Degradation Feedback}
\label{sec:3-6-1}

For each training sample \((\mathbf x_i,y_i)\), version \(t\) predicts

\begin{equation}
\widehat y_i^{(t)}
=
h_t(\mathbf x_i).
\end{equation}

The samples misclassified by \(P_t\) form the error set

\begin{equation}
\mathcal E_t
=
\left\{
(\mathbf x_i,y_i)
\in
\mathcal D_{\mathrm{tr}}
\;\middle|\;
\widehat y_i^{(t)}
\neq
y_i
\right\}.
\end{equation}

The error set alone does not reveal whether the latest revision has damaged previously correct behavior. For \(t\geq1\), the degradation set therefore contains cases that were correct under \(P_{t-1}\) but become incorrect under \(P_t\):

\begin{equation}
\mathcal G_t
=
\left\{
(\mathbf x_i,y_i)
\in
\mathcal D_{\mathrm{tr}}
\;\middle|\;
\widehat y_i^{(t-1)}
=
y_i
\land
\widehat y_i^{(t)}
\neq
y_i
\right\},
\qquad
\mathcal G_0=\varnothing.
\end{equation}

To keep the prompt context bounded, let

\begin{equation}
n_{\mathrm{err}}^{\max}
\in
\mathbb N_+
\end{equation}

denote the maximum number of current-error examples and select a context subset satisfying

\begin{equation}
\mathcal E_t^{\mathrm{ctx}}
\subseteq
\mathcal E_t,
\qquad
\left|
\mathcal E_t^{\mathrm{ctx}}
\right|
=
\min
\left\{
n_{\mathrm{err}}^{\max},
|\mathcal E_t|
\right\}.
\end{equation}

When \(|\mathcal E_t|>n_{\mathrm{err}}^{\max}\), the subset is sampled uniformly without replacement. Otherwise, \(\mathcal E_t^{\mathrm{ctx}}=\mathcal E_t\). The degradation context is bounded analogously. Let

\begin{equation}
n_{\mathrm{deg}}^{\max}
\in
\mathbb N_+
\end{equation}

denote its maximum size and require

\begin{equation}
\mathcal G_t^{\mathrm{ctx}}
\subseteq
\mathcal G_t,
\qquad
\left|
\mathcal G_t^{\mathrm{ctx}}
\right|
=
\min
\left\{
n_{\mathrm{deg}}^{\max},
|\mathcal G_t|
\right\}.
\end{equation}

When the degradation set exceeds this limit, examples are again sampled uniformly without replacement. Otherwise, \(\mathcal G_t^{\mathrm{ctx}}=\mathcal G_t\), with \(\mathcal G_0^{\mathrm{ctx}}=\varnothing\). Thus, \(\mathcal E_t\) captures unresolved errors, whereas \(\mathcal G_t\) isolates regressions introduced by the latest revision. Their bounded subsets provide complementary feedback to the LLM. For binary tasks, the implementation also reports false-positive and false-negative counts.

\subsubsection{Program Proposal and Current-Solution Update}
\label{sec:3-6-2}

Program revision is conditioned on the ordered history

\begin{equation}
\mathcal H_t
=
\left(
P_s,
B_s
\right)_{s=0}^{t},
\end{equation}

where \(B_s\) records the rationale associated with \(P_s\). Because revision order carries information, \(\mathcal H_t\) is retained as a sequence rather than a set. Each \(P_s\) represents \(h_s=h_{P_s}\), so the stored programs expose the complete classifier trajectory to the LLM rather than only the current function \(h_t\). Relevant feature information is already present in the programs and feedback examples, and \(T_2\) is therefore not supplied again. The next version is proposed as

\begin{equation}
\left(
P_{t+1},
B_{t+1}
\right)
=
\operatorname{LLM}
\left(
\operatorname{Prompt}_{\mathrm{iter}}
\left(
\mathcal H_t,
\mathcal E_t^{\mathrm{ctx}},
\mathcal G_t^{\mathrm{ctx}},
T_1,
T_3
\right)
\right).
\end{equation}

The proposed program must satisfy

\begin{equation}
P_{t+1}
\in
\mathcal P_\Gamma,
\end{equation}

and \(B_{t+1}\) records the corresponding revision rationale. The update context therefore combines all preceding programs, unresolved errors, newly introduced regressions, the task description, and the metric priorities. At the framework level, this history-conditioned edit acts as the candidate-update operator without requiring an explicit program-distance metric or predefined neighborhood.

In the present implementation, each update is realized as a targeted edit to the current rule code. The proposal is constrained to small adjustments in thresholds, weights, or conditions, prioritizes regressed cases, and discourages near-constant predictions. This edit budget is an implementation-level search bias rather than part of the general grammar. A proposed version enters the retained trajectory only after satisfying the required implementation-level checks.

Once accepted, \(P_{t+1}\) is retained together with \(B_{t+1}\) and \(\mathbf L(P_{t+1};\mathcal D_{\mathrm{val}})\). Here, acceptance means that \(P_{t+1}\) satisfies the hard constraint \(C_{\mathrm{hard}}(P_{t+1})=1\), rather than achieving a performance improvement, so it may score worse than \(P_t\). The trajectory terminates when no valid proposal is obtained within the retry limit or when the configured iteration budget is exhausted.

\subsubsection{Retained Valid Versions and Final Selection}
\label{sec:3-6-3}

Let \(N\) index the last successfully generated and retained version. The iterative operator returns

\begin{equation}
\mathcal A_N
=
\{P_0,P_1,\ldots,P_N\}
\subseteq
\mathcal F_\Gamma,
\qquad
\mathcal F_\Gamma
=
\left\{
P\in\mathcal P_\Gamma
\mid
C_{\mathrm{hard}}(P)=1
\right\}.
\end{equation}

The retained valid-version set records a single search trajectory rather than a population. Each round adds at most one valid successor. Retained programs neither evolve in parallel nor undergo recombination. Accordingly, the validation-cost sequence

\begin{equation}
\left(
\mathbf L
\left(
P_t;
\mathcal D_{\mathrm{val}}
\right)
\right)_{t=0}^{N}
\end{equation}

need not improve monotonically. Every valid program remains available for retrospective comparison:

\begin{equation}
P^\star
\in
\operatorname*{arg\,lex\,min}_{P\in\mathcal A_N}
\mathbf L
\left(
P;
\mathcal D_{\mathrm{val}}
\right).
\end{equation}

The lexicographic selector realizes the priority order defined in Section~\ref{sec:3-1-7} by applying nested filtering to the retained set. The first level retains

\begin{equation}
\mathcal A_N^{(1)}
=
\operatorname*{arg\,min}_{P\in\mathcal A_N}
\ell_1
\left(
h_P;
\mathcal D_{\mathrm{val}}
\right).
\end{equation}

At each subsequent priority \(q=2,\ldots,Q\), comparison is restricted to the versions retained at the preceding level:

\begin{equation}
\mathcal A_N^{(q)}
=
\operatorname*{arg\,min}_{P\in\mathcal A_N^{(q-1)}}
\ell_q
\left(
h_P;
\mathcal D_{\mathrm{val}}
\right),
\qquad
q=2,\ldots,Q.
\end{equation}

This procedure yields the nested sequence of retained subsets

\begin{equation}
\mathcal A_N
\supseteq
\mathcal A_N^{(1)}
\supseteq
\cdots
\supseteq
\mathcal A_N^{(Q)},
\qquad
P^\star
\in
\mathcal A_N^{(Q)}.
\end{equation}

Thus, a lower-priority metric is considered only among versions tied on every higher-priority metric. The implementation performs the equivalent lexicographic maximization of the original performance metrics and retains the earliest retained version in the event of a complete tie. This deterministic tie-break makes \(\operatorname{Select}_{\mathrm{lex}}\) return a single program \(P^\star\), whose classification function is

\begin{equation}
h_{P^\star}:
\mathcal X
\rightarrow
\mathcal Y.
\end{equation}

The selected program is exported through a stable inference interface and is evaluated on \(\mathcal D_{\mathrm{te}}\) only after selection is complete. The notation distinguishes the ideal solution \(P^\dagger\), defined over the full feasible program space, from \(P^\star\), the best program found in the finite set of retained versions generated by MHL. Taken together, these mechanisms characterize MHL as a trajectory-based, single-solution metaheuristic over rule programs with retrospective selection over retained versions under hard constraints. The LLM proposes revisions from program history and sample-level feedback, training errors and regressions guide the search trajectory, and validation-set lexicographic fitness selects the final selected program without gradient-based or hidden-parameter updates.

\subsection{Continual Learning Under Feature Evolution}
\label{sec:3-7}

MHL handles feature-space evolution by carrying explicit evidence and rule artifacts across successive learning stages. Stage 1 follows Sections~\ref{sec:3-2}--\ref{sec:3-6} and retains \(S^{(1)}\), \(K^{(1)}\), and the selected program \(P^{\star,(1)}\). For Stage 2, the removed and added feature names are represented by

\begin{equation}
\Delta_{\mathrm{col}}^{(2)}
=
\left(
\mathcal C_{\mathrm{drop}}^{(2)},
\mathcal C_{\mathrm{add}}^{(2)}
\right).
\end{equation}

The two sets contain column names, not feature values or free-text descriptions. The Stage-2 task text \(T_1^{(2)}\) supplies the clinical context, prediction objective, and a natural-language account of the feature changes. \(T_2^{(2)}\) and \(T_3^{(2)}\) retain the variable-information and evaluation-priority definitions from Section~\ref{sec:3-2}. The evidence update is

\begin{equation}
\begin{aligned}
S^{(2)}
&=
\operatorname{Probe}_{\mathrm{stat}}^{\mathrm{CL}}
\left(
\mathcal D_{\mathrm{tr}}^{(2)},
S^{(1)};
\Delta_{\mathrm{col}}^{(2)}
\right),\\
K^{(2)}
&=
\operatorname{Probe}_{\mathrm{know}}^{\mathrm{CL}}
\left(
T_1^{(2)},
T_2^{(2)},
K^{(1)};
\Delta_{\mathrm{col}}^{(2)}
\right).
\end{aligned}
\end{equation}

In these operators, the semicolon separates the structured change marker \(\Delta_{\mathrm{col}}^{(2)}\) from the primary inputs. The statistical update removes records for dropped features, retains unaffected Stage-1 records, computes statistics for added features from \(\mathcal D_{\mathrm{tr}}^{(2)}\), and merges the results. No natural-language description enters this computation.

The knowledge update invokes the LLM only for newly added features. With \(T_{2,\mathrm{add}}^{(2)}\) denoting their variable-information text together with the target, the new records are

\begin{equation}
\begin{aligned}
K_{\mathrm{add}}^{(2)}
&=
\operatorname{Probe}_{\mathrm{know}}
\left(
T_1^{(2)},
T_{2,\mathrm{add}}^{(2)}
\right)\\
&=
\operatorname{LLM}
\left(
\operatorname{Prompt}_{\mathrm{know}}
\left(
T_1^{(2)},
T_{2,\mathrm{add}}^{(2)}
\right)
\right).
\end{aligned}
\end{equation}

After obsolete entries are removed, \(K_{\mathrm{add}}^{(2)}\) is merged with the retained knowledge records to form \(K^{(2)}\). Thus, natural-language task context enters the knowledge probe and later LLM operations but not the statistical probe.

The selected Stage-1 program then provides the warm-start blueprint for Stage-2 initialization:

\begin{equation}
\begin{aligned}
\left(
P_0^{(2)},
B_0^{(2)}
\right)
&=
\operatorname{Init}_{\mathrm{LLM}}^{\mathrm{CL}}
\left(
P^{\star,(1)},
S^{(2)},
K^{(2)},
\Delta_{\mathrm{col}}^{(2)},
T_1^{(2)},
T_3^{(2)}
\right)\\
&=
\operatorname{LLM}
\left(
\operatorname{Prompt}_{\mathrm{cont}}
\left(
P^{\star,(1)},
S^{(2)},
K^{(2)},
\Delta_{\mathrm{col}}^{(2)},
T_1^{(2)},
T_3^{(2)}
\right)
\right).
\end{aligned}
\end{equation}

The structured marker identifies the affected columns, while \(T_1^{(2)}\) supplies their task-level semantics. The generated program must satisfy

\begin{equation}
P_0^{(2)}
\in
\mathcal P_{\Gamma^{(2)}},
\end{equation}

where \(B_0^{(2)}\) explains the adaptation. Because \(S^{(2)}\) and \(K^{(2)}\) already encode the variable information, \(T_2^{(2)}\) is not repeated as a separate initialization input. The continual adaptation step uses the selected Stage-1 program as an explicit warm-start blueprint, removes obsolete feature references, and admits added features when supported by the updated evidence.

After initialization, Stage 2 reuses the error, degradation, version-retention, and validation-selection procedures from Section~\ref{sec:3-6}. If \(N_2\) indexes the last valid Stage-2 version, then

\begin{equation}
\begin{aligned}
\mathcal A_{N_2}
&=
\operatorname{Iter}_{\mathrm{LLM}}
\left(
P_0^{(2)},
\mathcal D_{\mathrm{tr}}^{(2)},
\mathcal D_{\mathrm{val}}^{(2)},
T_1^{(2)},
T_3^{(2)}
\right),\\
P^{\star,(2)}
&=
\operatorname{Select}_{\mathrm{lex}}
\left(
\mathcal A_{N_2},
\mathcal D_{\mathrm{val}}^{(2)}
\right).
\end{aligned}
\end{equation}

Stage 2 uses only its own training and validation rows. Stage-1 training records are not replayed. Continual learning is therefore realized by explicitly inheriting and revising evidence and rule artifacts, rather than by transferring hidden state or overwriting learned parameters.

\subsection{Python Reference Implementation and Software Infrastructure}
\label{sec:3-8}

Although the abstract MHL formulation is independent of any programming language, the workflow used in this study is instantiated in Python without altering the framework-level semantics introduced in Sections~\ref{sec:3-2}--\ref{sec:3-7}.

Each retained rule program \(P_t\) is represented as a self-contained pure-Python function that relies exclusively on the Python standard library. Initialization produces the first version \texttt{predict\_v0(features: dict) -> int}. Each accepted revision appends a new versioned function such as \texttt{predict\_v1} or \texttt{predict\_v2}. Every referenced feature is read from the input dictionary. Each function returns an integer class label. Each conditional branch includes a concise English rationale comment. After validation-based selection, the chosen program is exported through a stable inference interface.

The Python instantiation also specifies an explicit LLM interaction contract. Both initialization and iterative revision require a strict JSON response containing \texttt{version}, \texttt{error\_analysis}, and \texttt{new\_policy\_code}. The field \texttt{error\_analysis} stores the English rationale associated with the proposed program. The field \texttt{new\_policy\_code} contains the complete Python function for that version. Candidate responses are parsed automatically and checked for the required version identifier, function name, Python syntax, undefined Python names, and module-loading validity. These checks operationalize part of \(C_{\mathrm{hard}}\), but they do not exhaustively verify every possible execution path or the correctness of the medical rationale. Failed parsing or validation triggers another attempt. An accepted proposal is appended to the retained version history.

The abstract prompt operators introduced earlier are instantiated as explicit Python templates. \(\operatorname{Prompt}_{\mathrm{know}}\) corresponds to \texttt{get\_knowledge\_probe\_prompt(...)}. For newly added features during continual learning, the corresponding template is \texttt{get\_continuous\_knowledge\_probe\_prompt(...)}. The initialization and revision operators correspond to \texttt{get\_rule\_generation\_prompt(...)} and \texttt{get\_iteration\_prompt(...)}. For continual adaptation, \(\operatorname{Prompt}_{\mathrm{cont}}\) corresponds to \texttt{get\_continuous\_v0\_generation\_prompt(...)}. Subsequent Stage-2 revisions use \texttt{get\_continuous\_iteration\_prompt(...)}. The engineering implementation also supports explicit renaming pairs as schema mappings. In this setting, the warm start is realized from the previously exported final-model artifact, which preserves the selected version and its history. The corresponding prompt templates are provided in Appendix~\ref{app:prompt-templates}.

The implementation follows the PEP 8 coding conventions \cite{vanrossum2001pep8} and the packaging standards maintained by the Python Packaging Authority. Its build interface, build-system requirements, and project metadata follow PEP 517, PEP 518, and PEP 621, respectively \cite{smith2015pep517,cannon2016pep518,cannon2020pep621}. The software is distributed through the Python Package Index under the package name \texttt{medical-heuristic-learning}. A \texttt{pytest}-based test suite is executed through GitHub Actions as a continuous-integration pipeline. Upon a release trigger, the workflow builds and publishes the versioned distribution automatically.

\subsection{Case Study of Rule Generation and Iteration}
\label{sec:3-9}

\begin{figure}[t]
\centering
\includegraphics[width=\textwidth]{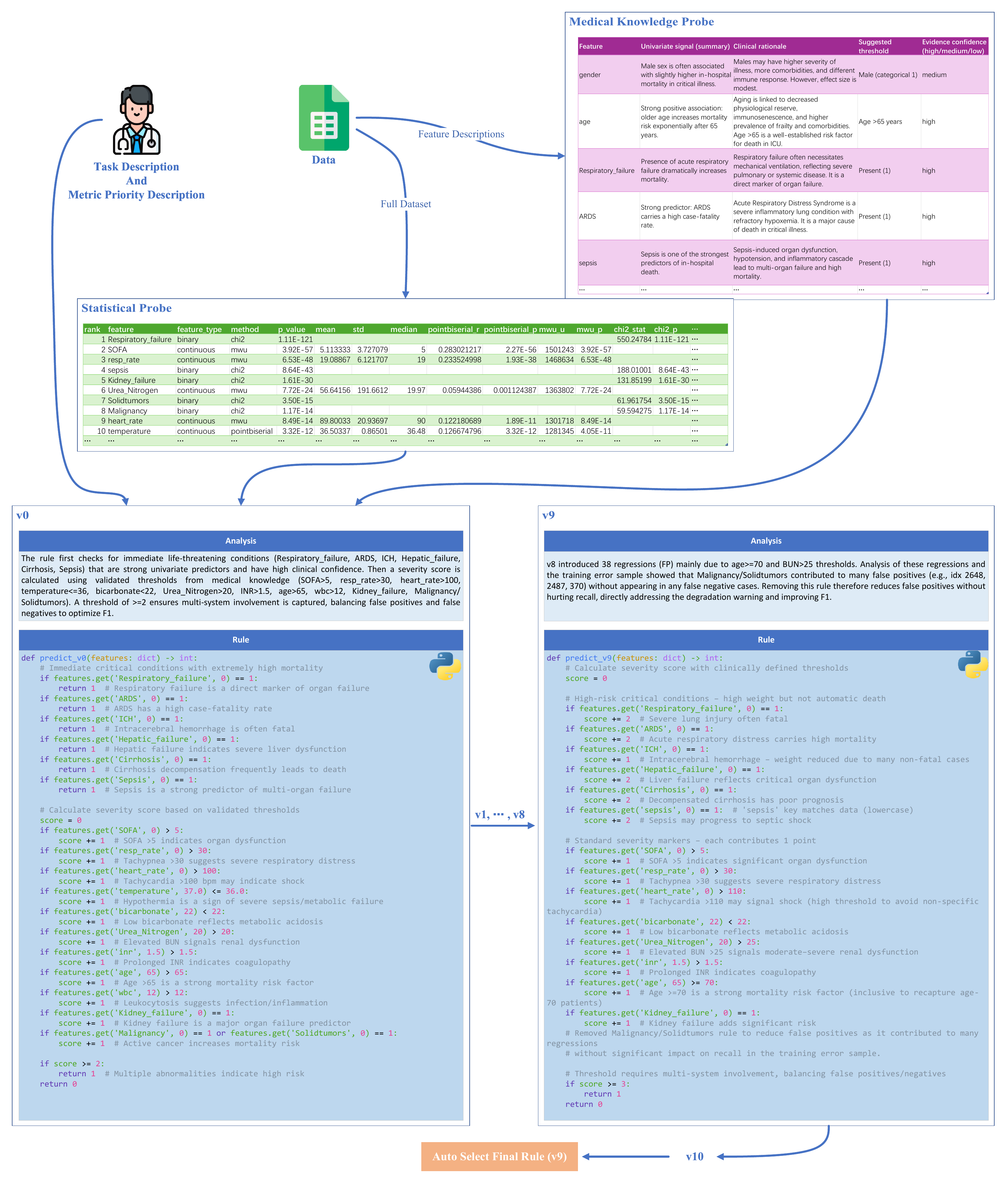}
\caption{Case study of MHL rule generation and revision on the CCID dataset, showing probe outputs, the initial rule, and the final selected rule with revision notes.}
\label{fig:results-2-2-case-study}
\end{figure}

To illustrate how the workflow materializes in practice, we present a representative case from the CCID dataset with 3000 training samples. Figure \ref{fig:results-2-2-case-study} shows a complete white-box artifact chain, including the task description, the metric priority description, excerpts from the statistical probe, excerpts from the medical knowledge probe, the initial rule \texttt{v0} with its generation rationale, and the final selected rule \texttt{v9} with its revision rationale. Taken together, these artifacts show how task requirements, evaluation priorities, empirical evidence, and clinical prior knowledge are organized into executable logic, and how validation feedback is subsequently converted into explicit grounds for rule revision.

\begin{figure}[t]
\centering
\includegraphics[width=0.72\textwidth]{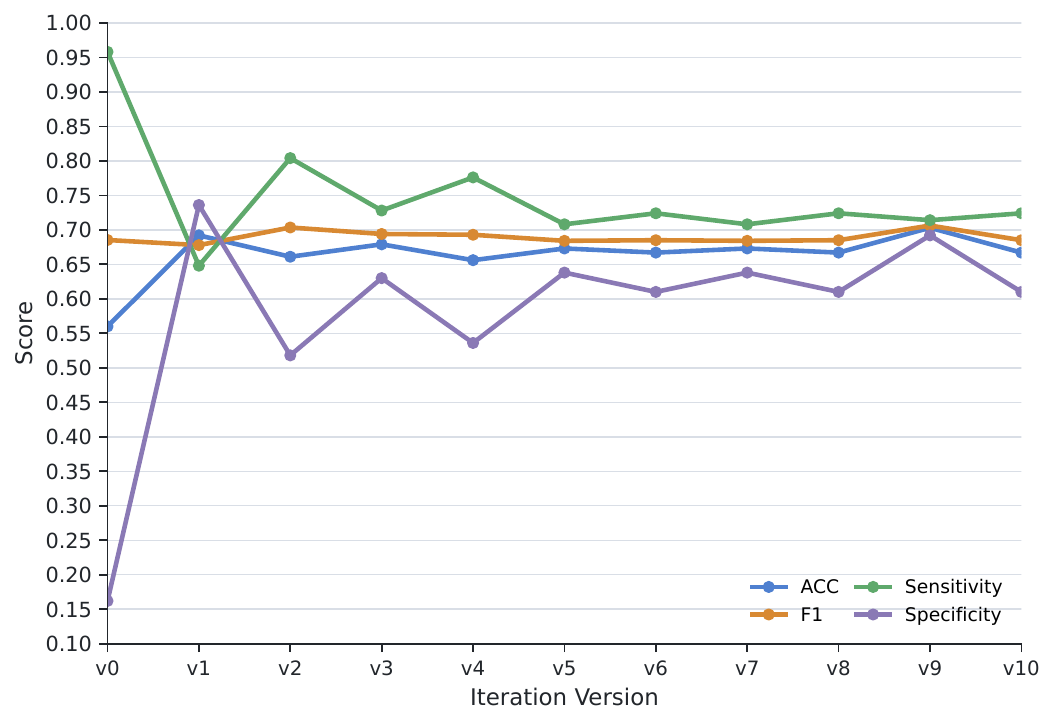}
\caption{Evolution of ACC, F1, precision, and recall during rule iteration for the CCID case study.}
\label{fig:results-2-2-rule-evolution}
\end{figure}

Figure \ref{fig:results-2-2-rule-evolution} further shows the performance trajectory across rule versions. In this case, the two probes initially yielded a rule with a strongly one-sided decision tendency. This pattern is visible in the starting metrics, where \texttt{v0} showed very high sensitivity of approximately 0.95 but very low specificity of approximately 0.15, while ACC remained modest. Subsequent iterations were then driven by validation feedback. The first few revisions mainly corrected this severe specificity deficit, bringing the rule back from an almost always-positive prediction pattern, whereas later revisions refined the balance among ACC, F1, sensitivity, and specificity through small and explicit code changes. By \texttt{v9}, the rule had reached a more appropriate overall trade-off, and the system selected it as the final version. This case clarifies how MHL functions as an interpretable and auditable rule-based expert system in practice and provides a concrete reference for the experiments reported in the Results section.

\section{Experimental Setup}
\label{sec:4}

\subsection{Datasets and Preprocessing}
\label{sec:4-1}

We evaluated Medical Heuristic Learning (MHL) on three medical datasets with structured predictor variables. UK Biobank (UKB) is a large public biomedical cohort resource, and in this study it was used for depression prediction from blood metabolomic measurements \cite{sudlow2015ukbiobank}. The Critical Care Information Database (CCID) is a private clinical dataset and serves as the in-hospital critical care prediction benchmark in this work \cite{wang2025predicting}. Because access to UKB requires paid subscription and a relatively complex application process, and because the open-release plan for CCID is still in progress, the LLMs used in our experiments had not been trained on either UKB or CCID data. The Medical Information Mart for Intensive Care (MIMIC) is a public intensive care database \cite{johnson2016mimic}, and in this study it was used primarily to evaluate continual adaptation under feature evolution. All tasks were binary classification problems.

For data handling, simple mean imputation was applied whenever a numerical predictor contained missing observations. This preprocessing procedure was applied uniformly to MHL and all baselines to ensure comparable inputs. Additional normalization was applied before training for Logistic Regression \cite{cox1958regression}, MLP \cite{rumelhart1986learning}, and FT-Transformer \cite{gorishniy2021revisiting} because these models are more sensitive to feature scale. The training, validation, and test sets were generated by one-time random sampling and then held fixed within each experimental configuration so that all methods were compared on identical splits. Different experiments then modified one factor at a time within this fixed-split design, such as training set size, class ratio, backend model, or stage-specific feature space.

\subsection{Evaluation Metrics}
\label{sec:4-2}

All experiments report Accuracy (ACC), F1, Sensitivity, and Specificity \cite{foody2023accuracy_metrics,sammut2011f1measure,akobeng2007diagnostic_tests}. ACC measures overall predictive correctness, while F1 summarizes the balance between precision and recall. Sensitivity measures the proportion of positive cases correctly detected, whereas Specificity measures the proportion of negative cases correctly identified. Together, these metrics characterize overall and class-specific behavior under small-sample, class-imbalance, and feature-evolution settings. The evaluation of robustness under severe class imbalance additionally reports the numbers of true positives (TP), false positives (FP), false negatives (FN), and true negatives (TN). Baseline classifiers convert positive-class probabilities to labels using a threshold of 0.5, whereas the selected MHL rule program directly outputs a binary prediction.

\subsection{Experimental Settings}
\label{sec:4-3}

All experiments were run under Linux on Windows Subsystem for Linux 2 (WSL2), specifically Ubuntu 24.04.2 LTS, with Python 3.11. Unless otherwise specified, experiments were repeated with random seeds \texttt{36}, \texttt{40}, and \texttt{42}. The LLM backend was fixed to DeepSeek-V4-Pro-Max \cite{deepseekai2026deepseekv4} in all experiments except the experiment assessing the impact of LLM backends. Outside the ablation study on probe mechanisms, MHL used both the statistical probe and the medical knowledge probe (S+K). The ablation varied only the availability of these two probes. All configurations otherwise followed the same initial rule-generation and iterative rule-revision procedures.

\subsection{Ablation Study on Probe Mechanisms}
\label{sec:4-4}

To assess the individual and joint contributions of statistical evidence and elicited medical knowledge to rule generation and iterative refinement, we conducted a systematic ablation study on UKB \cite{sudlow2015ukbiobank} and CCID \cite{wang2025predicting}. We compared four configurations: no probe (N), the medical knowledge probe only (K), the statistical probe only (S), and both probes (S+K). For each configuration, the validation and test sets each contained 1000 samples, while the training-set size was varied over 10, 100, 1000, and 3000. All splits were constructed with a 1:1 ratio of positive to negative samples. These scales were chosen to examine the role of the two probes from highly data-constrained settings to relatively data-sufficient ones, while the use of both UKB and CCID allowed the comparison to cover two different medical prediction scenarios.

\subsection{Impact of LLM Backends}
\label{sec:4-5}

The experiment on the impact of LLM backends evaluates the applicability of MHL across different LLM backends and investigates how backend choice influences the final white-box rule programs produced by the framework. The experiment uses UKB \cite{sudlow2015ukbiobank} and CCID \cite{wang2025predicting}. For each dataset, the training, validation, and test sets each contain 1000 samples and are sampled without replacement at a 1:1 ratio of positive to negative instances. Seven backend configurations are compared. Four DeepSeek configurations are formed by crossing the Pro and Flash model variants with High and Max reasoning-strength levels. These configurations are DeepSeek-V4-Pro-High, DeepSeek-V4-Pro-Max, DeepSeek-V4-Flash-High, and DeepSeek-V4-Flash-Max \cite{deepseekai2026deepseekv4}. The remaining backends are Qwen 3.7-Max \cite{yang2025qwen3}, Gemini 3.1-Pro-Preview \cite{googledeepmind2026gemini3pro}, and GPT-5.5 \cite{openai2025gpt5systemcard}.

\subsection{Model Comparison Design}
\label{sec:4-6}

The model-comparison experiments are conducted on UKB \cite{sudlow2015ukbiobank} and CCID \cite{wang2025predicting}. The ten baselines were selected to span several major methodological families in clinical tabular prediction. Logistic Regression \cite{cox1958regression} and Decision Tree \cite{breiman1984classification} serve as classical baselines that are transparent by construction, with their decision functions represented by explicit coefficients and tree-structured branching, respectively. XGBoost \cite{chen2016xgboost} and LightGBM \cite{ke2017lightgbm} represent widely used gradient-boosted decision-tree ensembles for tabular prediction. The neural baselines comprise MLP \cite{rumelhart1986learning} as a standard feed-forward architecture and the tabular ResNet and FT-Transformer architectures \cite{gorishniy2021revisiting} as modern deep models designed for tabular data. The intrinsically interpretable baselines include Explainable Boosting Machine (EBM) \cite{lou2013ga2m,nori2019interpretml} and Automatic Piecewise Linear Regression (APLR) \cite{vonottenbreit2024automatic}, which provide additive and piecewise-linear model structures, respectively. CORELS \cite{angelino2018corels} complements these methods by learning certifiably optimal rule lists. Taken together, the baselines span classical parametric and tree models, boosted ensembles, neural architectures for tabular data, glass-box additive and piecewise-linear models, and rule-list learners. Neither the baselines nor MHL receive automatic compensation for class frequencies through class weights, sample weights, weighted sampling, or related mechanisms.

The model comparison comprises three complementary experiments evaluating robustness across training-set sizes, robustness under severe class imbalance, and continual adaptation under feature evolution. Robustness across training-set sizes is assessed by varying the number of training instances over 10, 50, 100, 500, 1,000, and 3,000. The validation and test sets each contain 1,000 samples, and the training, validation, and test sets are all sampled with equal numbers of positive and negative instances. These scales span settings from extremely small samples to relatively data-rich conditions, allowing changes in model performance to be examined under otherwise matched data partitions.

Robustness under severe class imbalance is assessed by systematically varying the positive-to-negative ratio in the training set over 1:1, 1:2, 2:1, 1:5, 5:1, 1:10, 10:1, 1:50, and 50:1. The validation and test sets remain balanced at a 1:1 class ratio and each contain 1,000 samples. This comparison is conducted separately with training-set sizes of 1,000 and 3,000, thereby assessing the effect of class imbalance at two different data scales. The design evaluates whether each method retains meaningful discrimination of the minority class as the training distribution becomes increasingly skewed.

The experiment on continual adaptation under feature evolution examines whether MHL can transfer explicit rule knowledge learned and validated on a preceding task to a subsequent task whose feature space has evolved and for which only limited training data are available. In sepsis assessment, the Sepsis-3 consensus shifted the clinical criteria from SIRS-based definitions to SOFA-based assessment of organ dysfunction \cite{singer2016sepsis3}. One practical consequence of this transition is that some clinical units may discontinue recording SIRS-related variables and begin recording SOFA-related variables, thereby creating a realistic form of feature evolution in clinical data collection. We simulate this transition through a two-stage task using the Medical Information Mart for Intensive Care (MIMIC) \cite{johnson2016mimic}. In Stage 1, all methods are trained on the original feature set containing SIRS, using 1,000 training samples. Stage 2 represents the early period after feature evolution, when the new SOFA-based feature space has entered use but only limited data have accumulated under the updated measurement regime. The models must therefore adapt to the new feature set using only 40 training samples. In each stage, the validation and test sets contain 500 and 800 samples, respectively. In this comparison, each baseline starts from its Stage 1 model and continues training after being transferred to the Stage 2 feature space. MHL instead starts from its previously validated rule program and revises its rules under the updated feature definition. The experiment includes only methods for which cross-stage transfer is supported in the current implementation, namely MLP \cite{rumelhart1986learning}, XGBoost \cite{chen2016xgboost}, LightGBM \cite{ke2017lightgbm}, EBM \cite{lou2013ga2m,nori2019interpretml}, the FT-Transformer and tabular ResNet architectures described by Gorishniy et al. \cite{gorishniy2021revisiting}, and MHL. As a reference condition, each method is also trained from scratch using only the same Stage 2 data.

\section{Results and Analysis}
\label{sec:5}

\subsection{Ablation Study on Probe Mechanisms}
\label{sec:5-1}

\begin{figure}[h]
\centering
\includegraphics[width=0.49\textwidth]{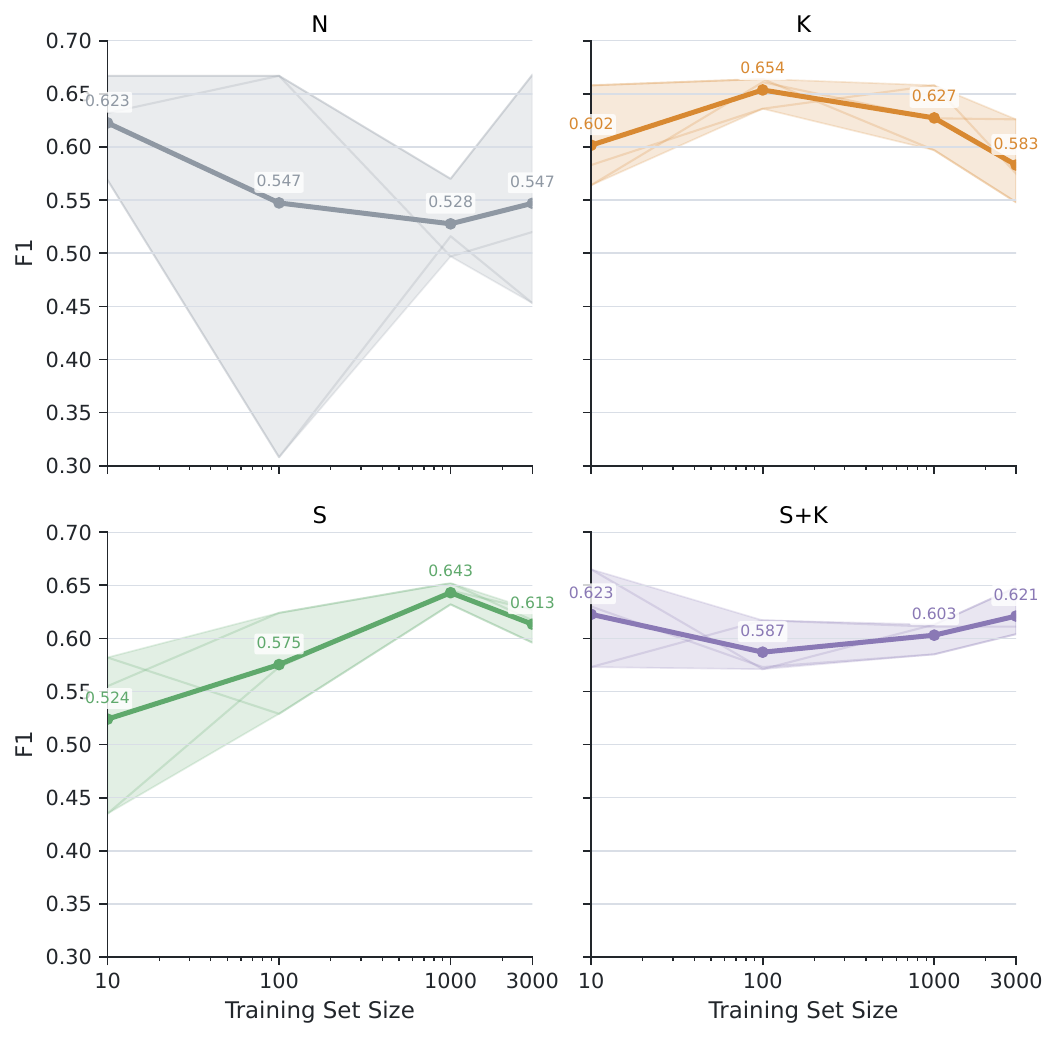}
\includegraphics[width=0.49\textwidth]{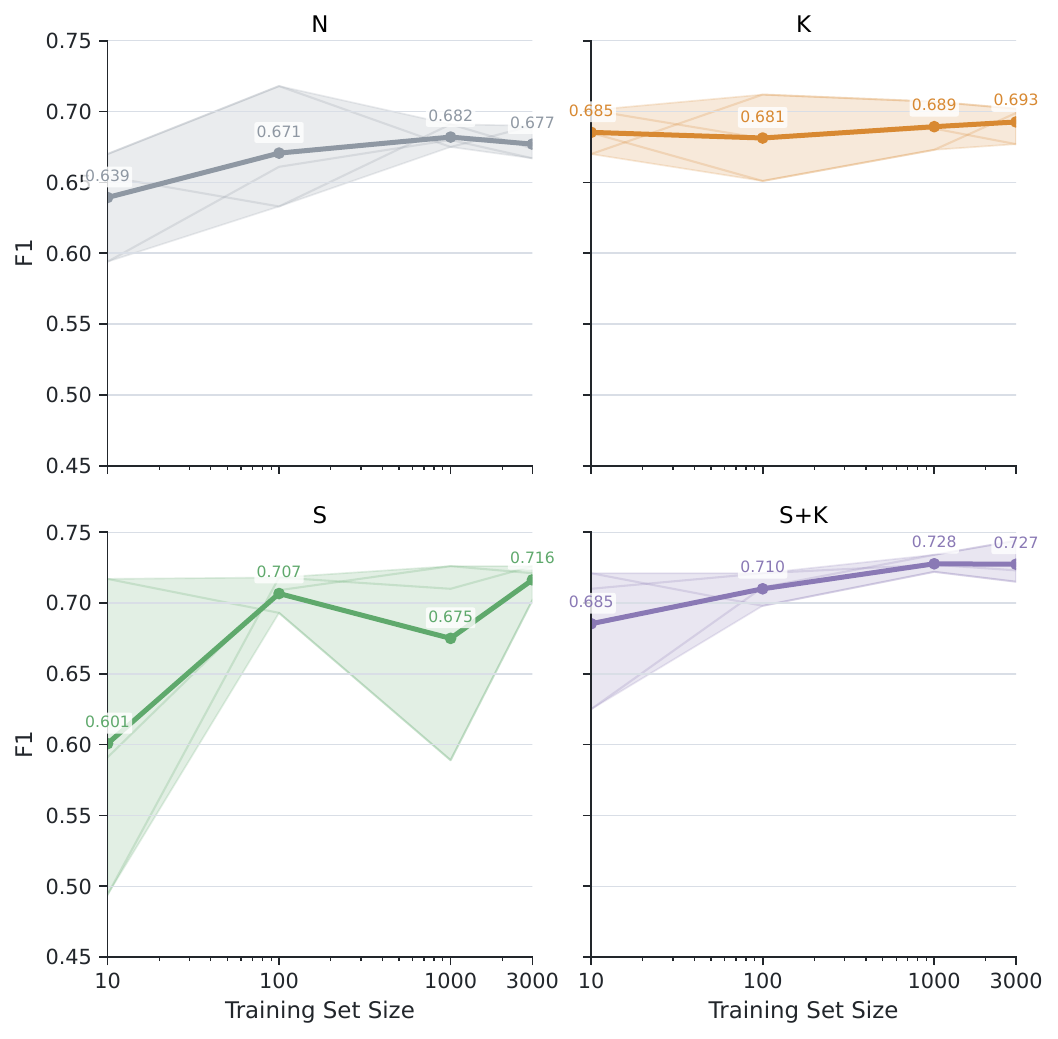}
\caption{F1 of MHL under four probe configurations on UKB (left) and CCID (right).}
\label{fig:results-2-3-ablation}
\end{figure}

The ablation results under different probe configurations are shown in Figure \ref{fig:results-2-3-ablation}, which summarizes the outcomes on UKB \cite{sudlow2015ukbiobank} (Left) and CCID \cite{wang2025predicting} (Right) across the four settings N, K, S, and S+K, with the x-axis shown on a logarithmic scale. Detailed metrics are reported in Table \ref{tab:results-2-3-ablation}. In the figure, thick lines denote mean trends across three random seeds, thin lines denote seed-specific trajectories, and shaded bands denote the range of variation across seeds.

Across settings, S+K did not always achieve the highest score at every operating point, but it delivered the most stable and least failure-prone performance overall. In Figure \ref{fig:results-2-3-ablation}, N and S exhibit greater dispersion across random seeds, with particularly wide shaded bands in the low-sample regime. This pattern is consistent with stronger sampling variability under small training sets. When the sampled subset is less representative of the underlying distribution, a rule generator that relies only on sparse empirical signals is more likely to encode local noise as if it were a stable pattern. The introduction of K reduced this variability by supplying clinical priors and threshold references, thereby limiting the extent to which rule generation followed accidental small-sample regularities. When empirical evidence and medical priors were combined, the S+K curves became smoother and the seed-to-seed range became narrower. Table \ref{tab:results-2-3-ablation} supports the same conclusion, showing that S+K maintained a consistently usable F1 level across the examined sample sizes.

The ablation results further suggest that different probe settings bias the model toward different sensitivity-specificity trade-offs. Overall, K alone more often shifts the rules toward higher sensitivity, usually at the cost of specificity, whereas S alone more often favors specificity in better-sampled settings and may reduce sensitivity. These patterns should be interpreted as general tendencies rather than rules that hold in every setting. By contrast, S+K delivered the most appropriate overall balance in most settings and was more consistent in both stability and aggregate performance. Taken together, the results indicate that the two probes are distinct but complementary. The statistical probe prioritizes features that stand out in the observed data, whereas the medical knowledge probe organizes thresholds and rule structure in a more clinically grounded manner. On this basis, S+K was adopted as the default configuration in the subsequent experiments.

\subsection{Impact of LLM Backends}
\label{sec:5-2}

\begin{figure}[h]
\centering
\includegraphics[width=0.4\textwidth]{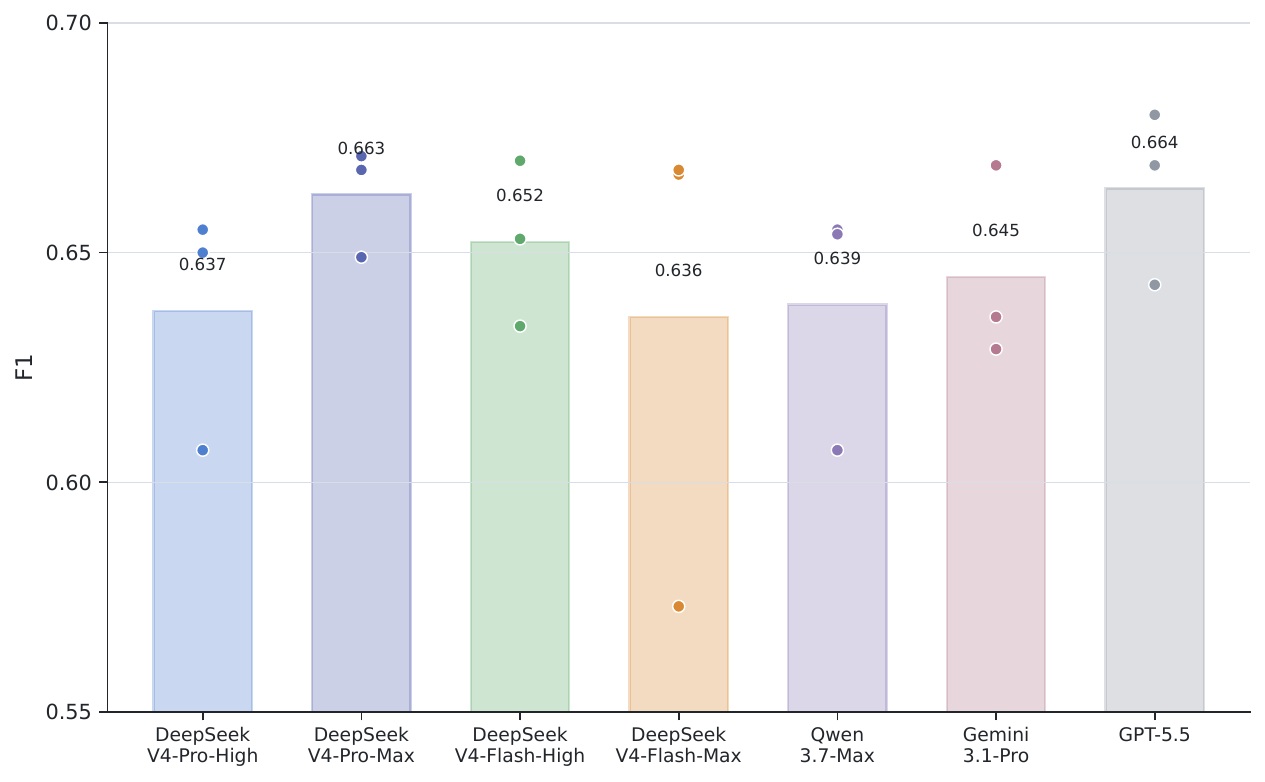}
\includegraphics[width=0.4\textwidth]{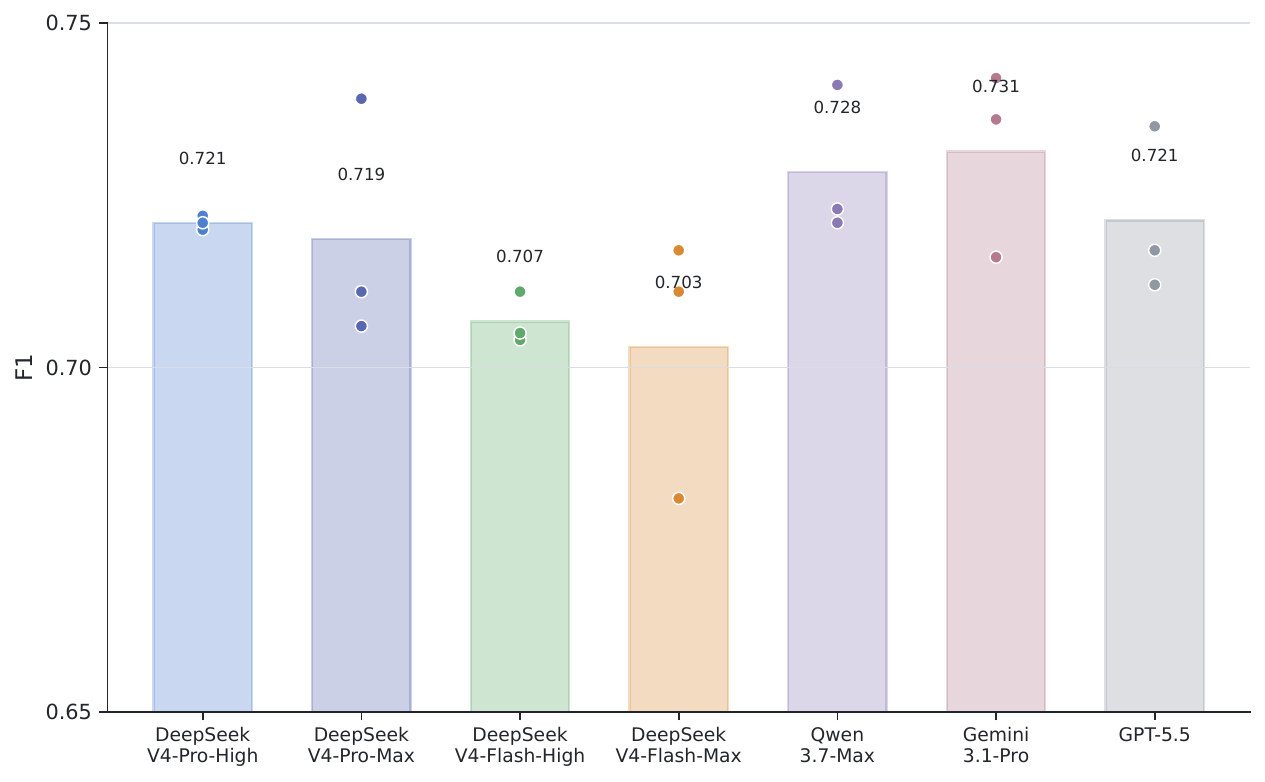}
\caption{F1 of MHL with seven LLM backend configurations for rule generation and revision on UKB (left) and CCID (right). Bars show means; points show the underlying observations.}
\label{fig:results-2-6-backends}
\end{figure}

Figure \ref{fig:results-2-6-backends} compares the final MHL rule programs generated and revised with seven LLM backend configurations on UKB \cite{sudlow2015ukbiobank} (left) and CCID \cite{wang2025predicting} (right). Bars show mean F1, and points show the underlying observations. The corresponding mean ACC, F1, Sensitivity, and Specificity are reported in Table \ref{tab:results-2-6-backends}. The LLMs do not classify test instances directly. They serve only as constrained operators during rule-program construction.

Mean F1 varied only modestly across configurations, but no configuration led on both datasets. On UKB, GPT-5.5 \cite{openai2025gpt5systemcard} achieved the highest mean F1 (0.664), narrowly exceeding DeepSeek-V4-Pro-Max \cite{deepseekai2026deepseekv4} (0.663). On CCID, Gemini 3.1-Pro \cite{googledeepmind2026gemini3pro} ranked first (0.731), followed by Qwen 3.7-Max \cite{yang2025qwen3} (0.728). The best-to-worst mean F1 spread was 0.028 on each dataset, while variation among the individual observations was comparable to several between-configuration differences. Increasing the configured reasoning strength from High to Max was likewise not uniformly beneficial. It improved the DeepSeek Pro variant only on UKB, while Flash-Max trailed Flash-High on both datasets. Backend ranking was therefore dataset dependent, with no universally preferable reasoning configuration.

However, the rule programs generated by different backends exhibited distinct decision styles. Gemini 3.1-Pro and Qwen 3.7-Max were the most consistently balanced. On both datasets, they preserved relatively high Sensitivity while attaining the two highest Specificity values. This balance translated into the two best ACC scores on both datasets and, on CCID, the two best F1 scores. GPT-5.5 was more recall-oriented and assigned positive labels more readily. This tendency was most pronounced on UKB, where the highest Sensitivity (0.899) coincided with the lowest Specificity (0.192), producing the highest F1 but the lowest ACC. This contrast shows that similar, or even leading, aggregate performance can conceal materially different behavior across the positive and negative classes. The DeepSeek configurations exhibited no single, stable style. Increasing the reasoning strength of the Pro variant consistently raised Sensitivity while reducing Specificity, whereas the direction of this change for the Flash variant depended on the dataset. Reasoning strength therefore does not by itself determine whether the resulting rules favor positive-case detection or negative-case exclusion. Rather, the decision style emerges from the interaction among the backend family, model configuration, and dataset. Because every configuration produced a non-degenerate executable rule program, the results support MHL's portability across backend families. Nevertheless, the backend remains an important model-design factor that should be controlled and reported, with selection guided by the relative costs of false negatives and false positives rather than F1 alone.

\subsection{Robustness Across Training-Set Sizes}
\label{sec:5-3}

\begin{figure}[h]
\centering
\includegraphics[width=0.49\textwidth]{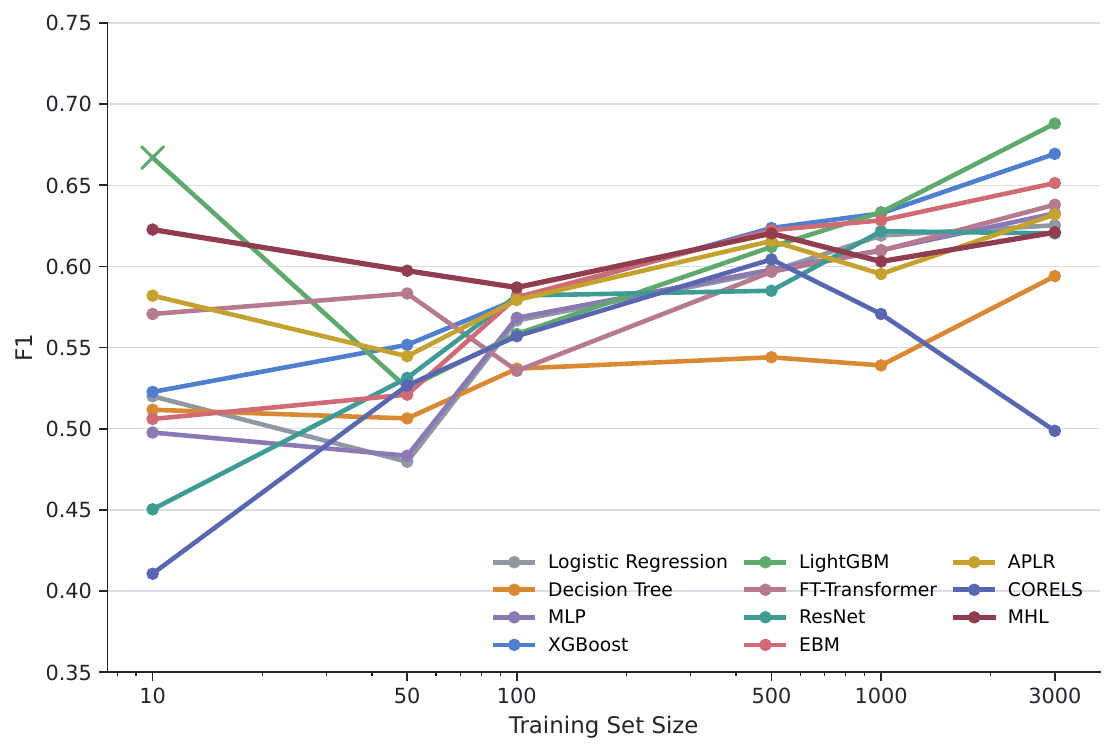}
\includegraphics[width=0.49\textwidth]{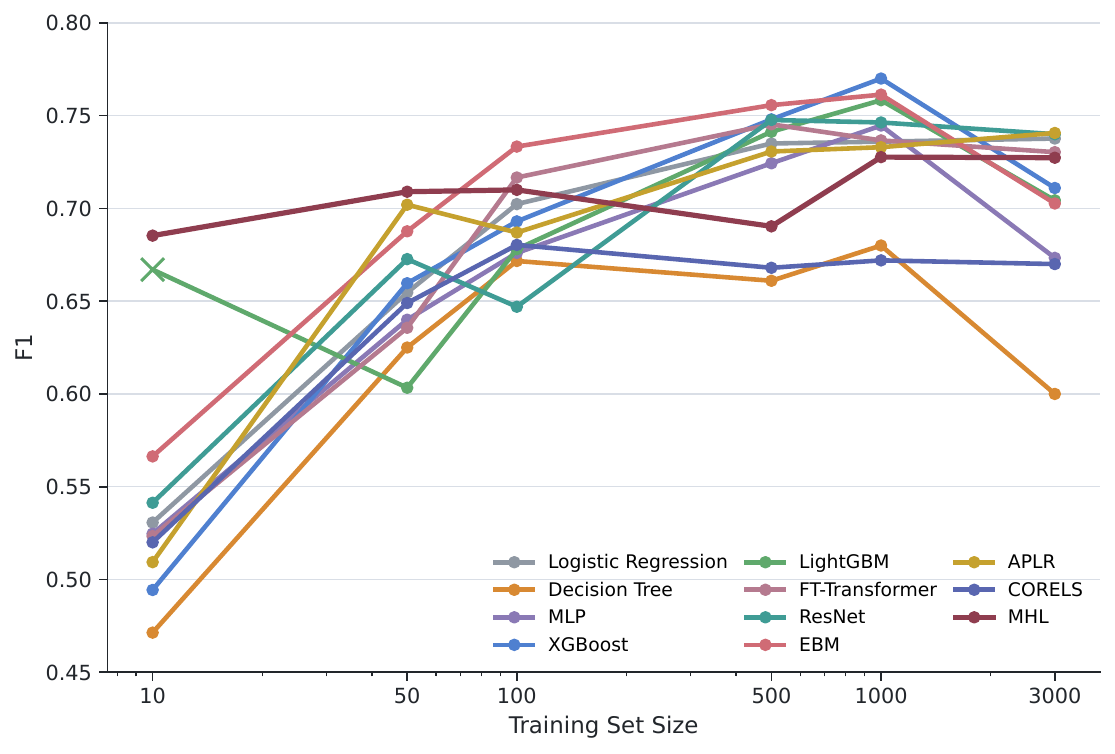}
\caption{Mean F1 at each training-set size on UKB (left) and CCID (right). Crosses denote LightGBM training collapse at $n=10$.}
\label{fig:results-2-4-sample-size}
\end{figure}

Figure \ref{fig:results-2-4-sample-size} shows mean F1 as the balanced training-set size increases from 10 to 3,000 on UKB \cite{sudlow2015ukbiobank} (left) and CCID \cite{wang2025predicting} (right). Full mean metrics are reported in Table \ref{tab:results-2-4-sample-size}. The crosses at $n=10$ denote LightGBM training collapse on both datasets. Each collapsed run assigned every test instance to the positive class, yielding Sensitivity $=1.000$ and Specificity $=0.000$. The corresponding F1 of 0.667 therefore reflects single-class failure rather than useful discrimination.

MHL showed its clearest advantage in the most data-constrained regimes. On UKB, it ranked first among the non-collapsed methods at $n=10$, 50, and 100. On CCID, it ranked first at $n=10$ and 50 and remained among the strongest methods at $n=100$. Its F1 remained comparatively stable across the full range of training sizes rather than depending on steep gains as more examples became available. The comparison with CORELS \cite{angelino2018corels} is especially informative because both methods produce explicit rule-based classifiers but construct them using different evidence and search processes. MHL achieved higher F1 than CORELS at every evaluated training size on both datasets. The difference was largest at $n=10$, when MHL and CORELS obtained 0.623 versus 0.411 on UKB and 0.685 versus 0.520 on CCID, respectively. Taken together, these results suggest that integrating statistical evidence, elicited medical knowledge, and feedback-guided revision provides MHL with useful structural guidance when the sampled data alone offer limited support for reliable empirical regularities. This guidance helps MHL construct effective explicit rules in low-data regimes while maintaining stable performance as data availability increases.

As the training set grew, several statistical and neural baselines benefited more substantially from the additional observations and surpassed MHL in some data-rich settings. MHL nevertheless varied within a comparatively narrow F1 range and remained competitive throughout. The results therefore characterize its principal empirical strengths as sample efficiency and stability across data scales rather than universal score dominance when abundant training data are available. Beyond this predictive behavior, MHL preserves white-box interpretability across all training sizes because the learned classifier remains an explicit rule program. Its predicates, thresholds, and decision paths can be inspected directly, making the basis of each prediction accessible without relying on a post hoc surrogate. This combination of cross-scale robustness and directly inspectable decision logic is a central advantage of MHL.

\subsection{Robustness under Severe Class Imbalance}
\label{sec:5-4}

\begin{figure}[h]
\centering
\includegraphics[width=0.49\textwidth]{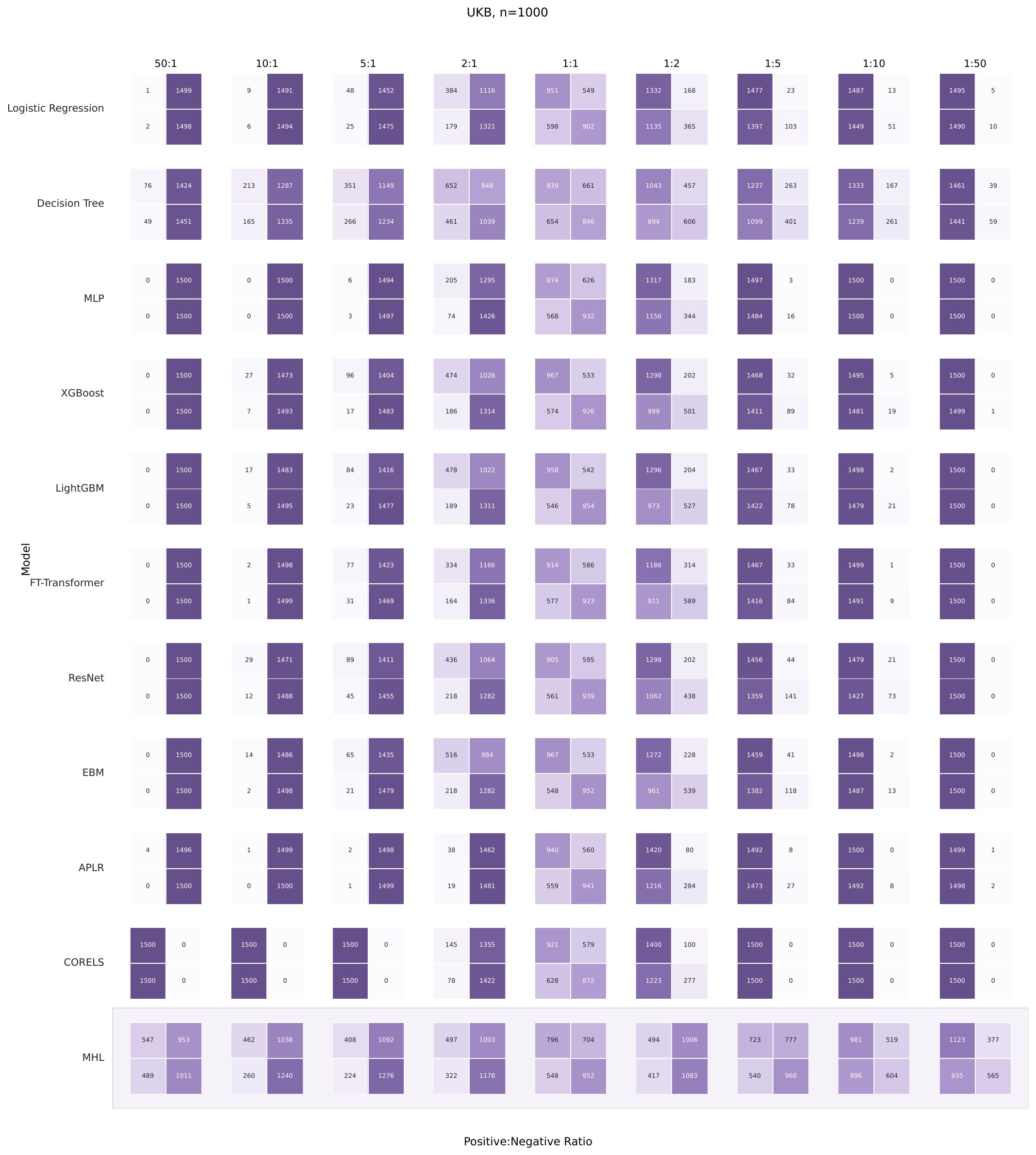}
\includegraphics[width=0.49\textwidth]{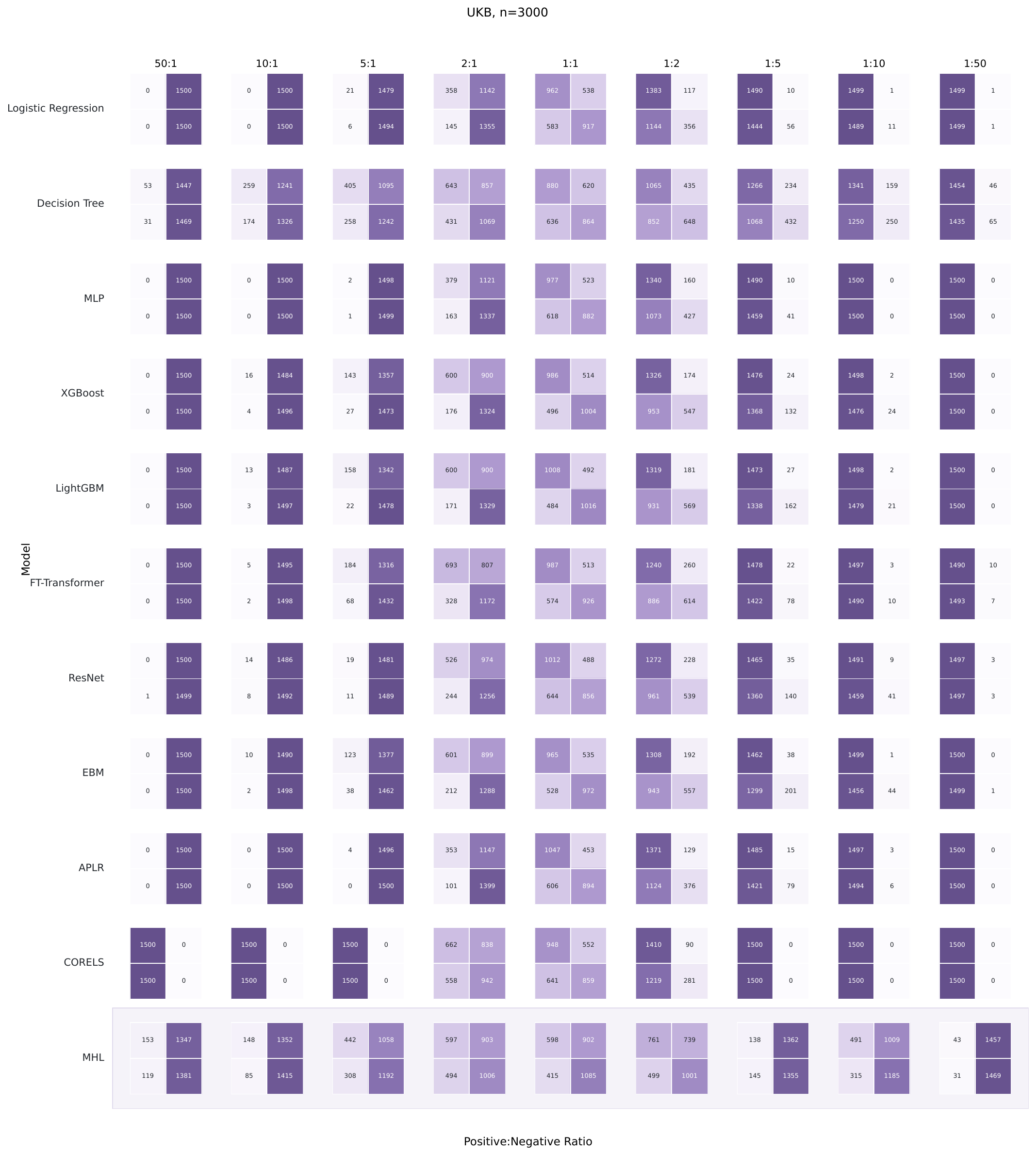}
\includegraphics[width=0.49\textwidth]{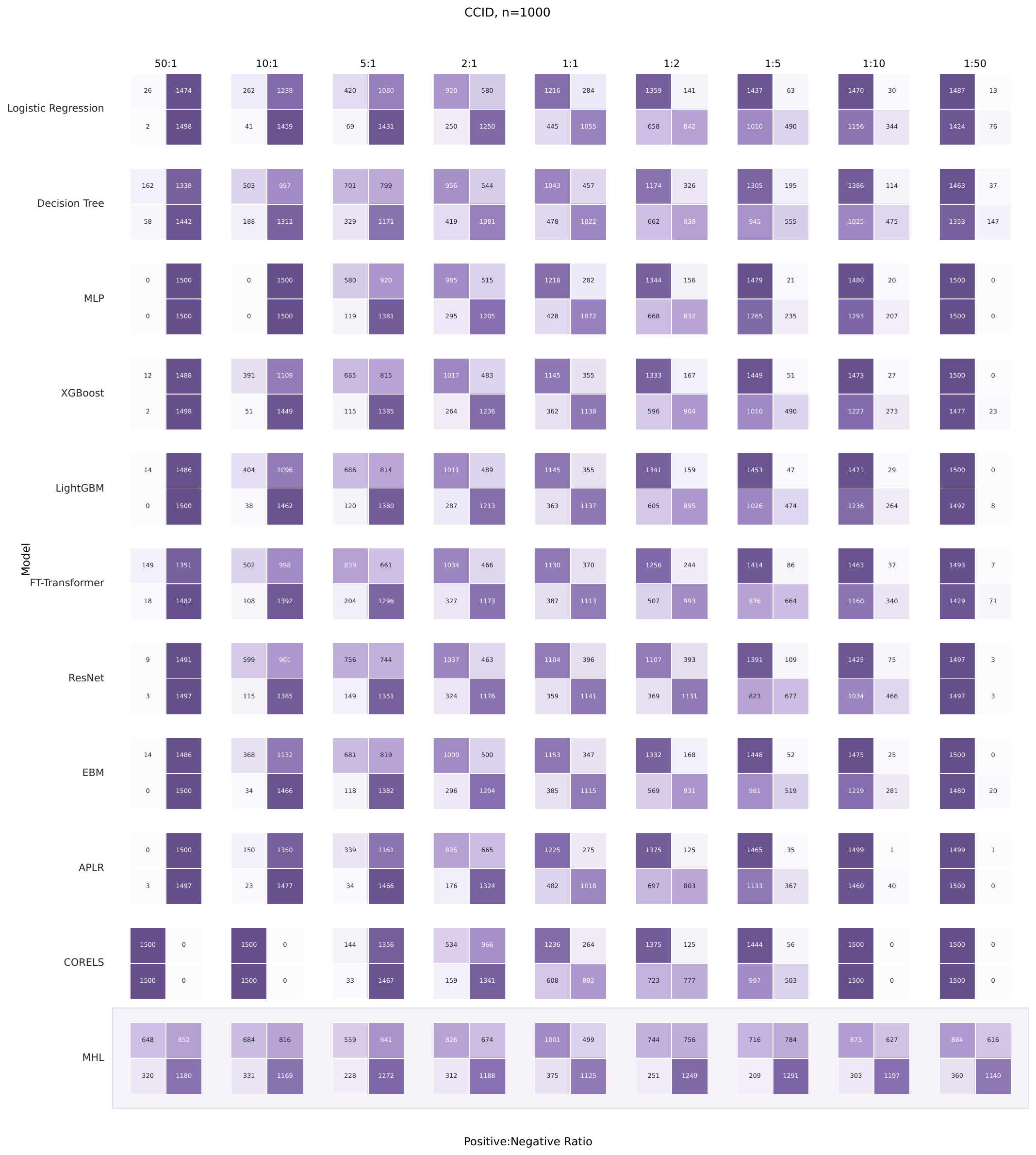}
\includegraphics[width=0.49\textwidth]{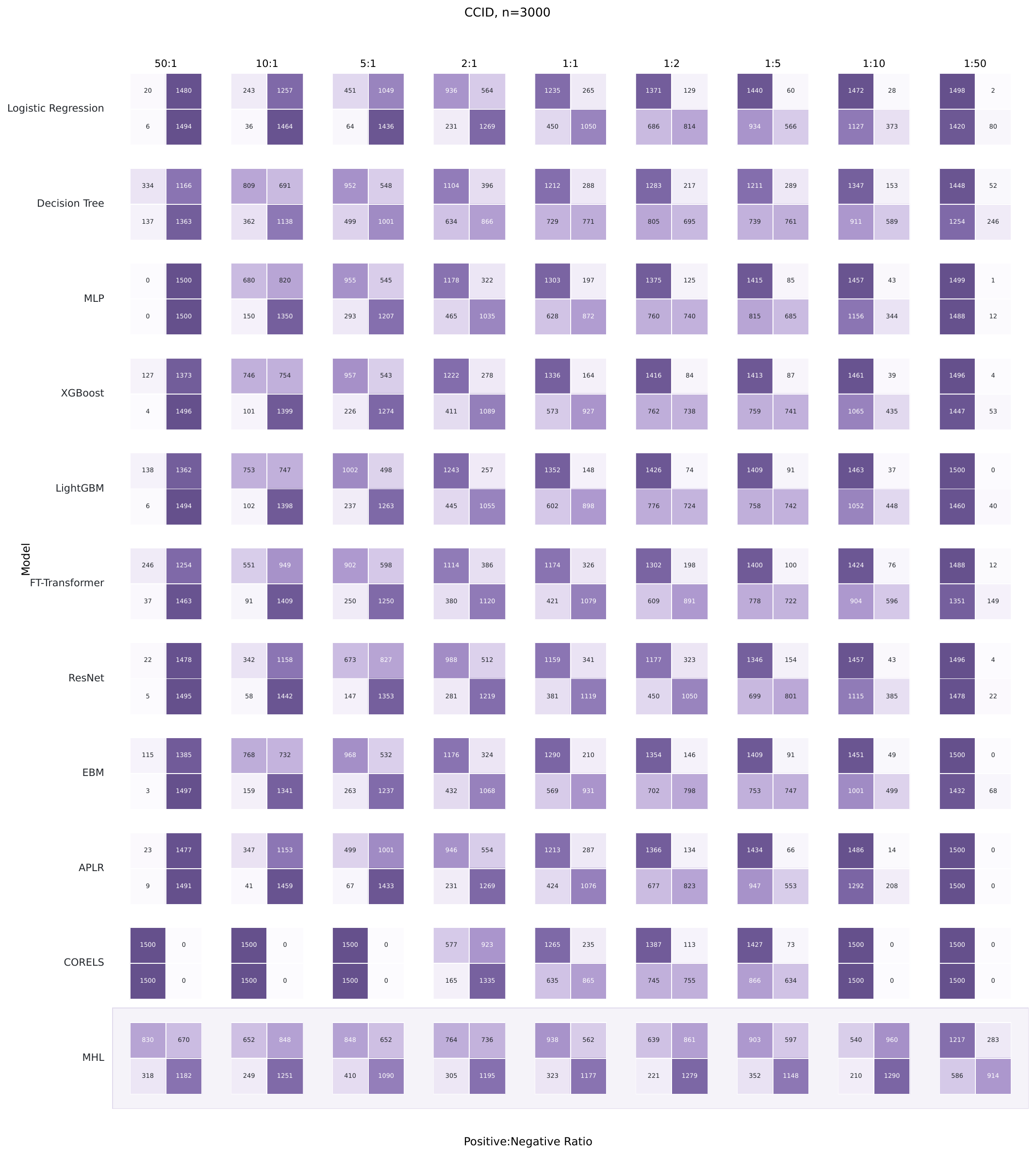}
\caption{Pooled confusion matrices under different positive-to-negative training ratios for UKB at $n=1{,}000$ and $n=3{,}000$ (top) and CCID at $n=1{,}000$ and $n=3{,}000$ (bottom). Each $2\times2$ block is ordered as $[[\mathrm{TN},\mathrm{FP}],[\mathrm{FN},\mathrm{TP}]].$}
\label{fig:results-2-5-imbalance}
\end{figure}

Figure \ref{fig:results-2-5-imbalance} reports pooled confusion matrices for UKB \cite{sudlow2015ukbiobank} and CCID \cite{wang2025predicting} at training sizes of 1,000 and 3,000. Table \ref{tab:results-2-5-imbalance} provides the corresponding mean metrics and pooled TP, FP, FN, and TN counts. For each model and positive-to-negative training ratio, the pooled counts comprise 1,500 positive and 1,500 negative test instances. Because class imbalance was imposed only during training and no automatic class correction was used, the balanced evaluation data isolate each learner's response to the altered training prior.

At the most extreme ratios, many baselines followed the dominant training class and collapsed toward single-class prediction, and the larger training set did not consistently prevent this behavior. Across the dataset and training-size conditions, most baselines kept both Sensitivity and Specificity above 0.4 only when the training ratio narrowed to between 1:2 and 2:1. MHL was substantially more resilient. On CCID, it was the only evaluated method to keep both Sensitivity and Specificity above 0.4 at 50:1 and 1:50 under both training sizes. On UKB with $n=1{,}000$, MHL likewise preserved nontrivial discrimination of both classes at the two extremes, whereas most baselines were already close to constant prediction. This pattern is consistent with MHL's evidence-guided rule search reducing susceptibility to severe shifts in the sampled class ratio. The main exception occurred on UKB with $n=3{,}000$, where MHL became strongly biased toward positive predictions at the extreme ratios, particularly at 1:50, underscoring the need for class-specific validation of every learned rule program. Overall, MHL maintained two-sided discrimination across a broader range of imbalance conditions than the competing methods, indicating that its explicit rule-construction process can remain effective even when the observed class distribution is highly skewed.

\subsection{Continual Adaptation under Feature Evolution}
\label{sec:5-5}

\begin{figure}[h]
\centering
\includegraphics[width=0.65\textwidth]{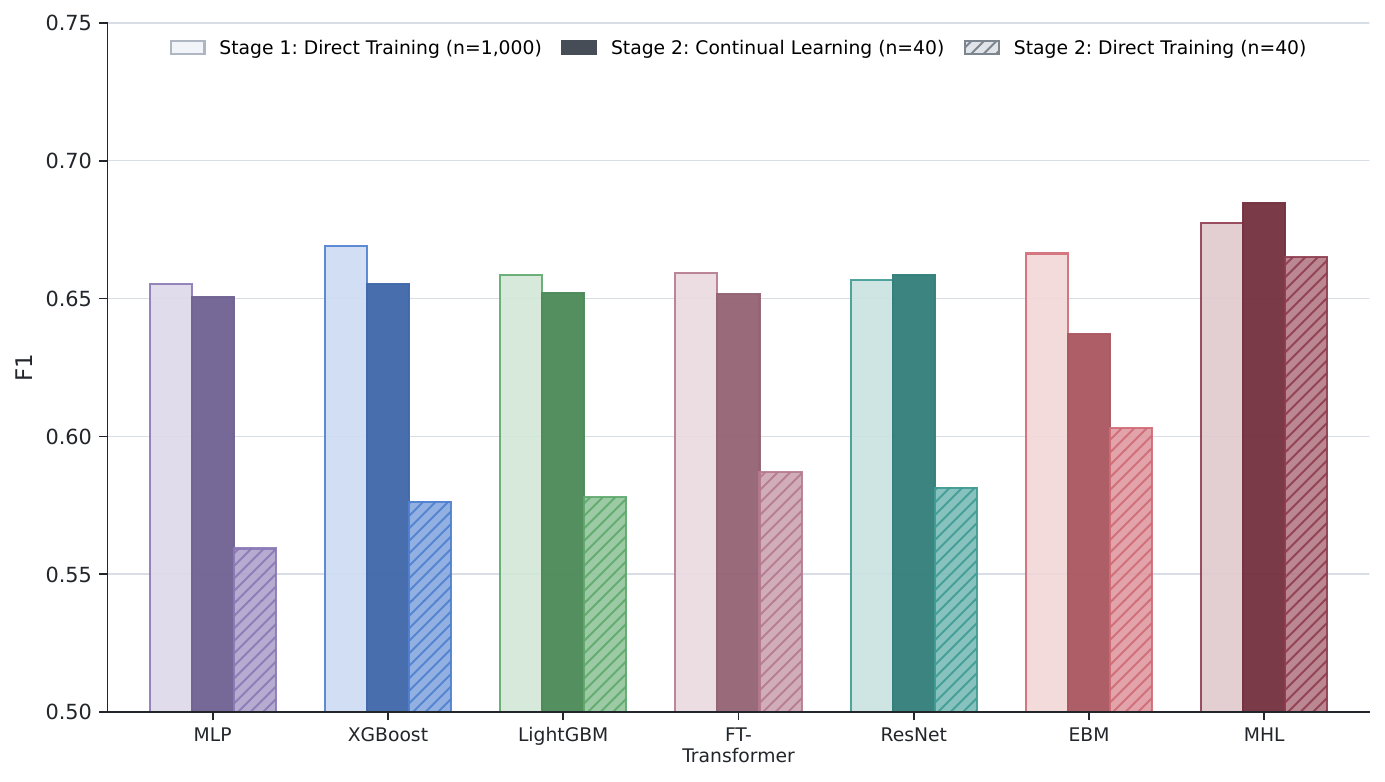}
\caption{F1 on MIMIC at the Stage 1 direct-training, Stage 2 continual-learning, and Stage 2 direct-training endpoints.}
\label{fig:results-2-7-continual-learning}
\end{figure}

Figure \ref{fig:results-2-7-continual-learning} compares three endpoints on MIMIC \cite{johnson2016mimic}. These endpoints comprise Stage 1 direct training with 1,000 samples, Stage 2 continual adaptation from the Stage 1 state with 40 new samples, and Stage 2 direct training from scratch with the same 40 samples. The continual branch inherits the Stage 1 model state but does not reuse any Stage 1 training samples during Stage 2 adaptation. Table \ref{tab:results-2-7-continual-learning} reports the mean ACC, F1, Sensitivity, and Specificity.

Continual adaptation improved F1 over Stage 2 direct training for every evaluated method. MHL achieved the highest continual-adaptation F1 of 0.685 and improved by 0.020 over direct Stage 2 MHL training. Following the removal of SIRS and introduction of SOFA in Stage 2, the Stage 1 rule and probe outputs provided an effective warm-start blueprint when only 40 new samples were available. This result highlights MHL's ability to maintain an explicit rule program rather than reconstruct its decision logic entirely from scratch.

The cross-stage pattern provides complementary evidence about knowledge retention. From Stage 1 to the continual Stage 2 endpoint, ACC decreased for every baseline and remained nearly unchanged for MHL. F1 decreased for five baselines and changed only marginally for ResNet, whereas MHL increased from 0.677 to 0.685. Although Stage 1 and Stage 2 use different test sets, these shifts cannot serve as direct measures of forgetting. Nevertheless, the contrasting pattern is consistent with MHL retaining and transferring previously learned rule knowledge more effectively through feature evolution. Together with its improvement over training from scratch on the shared Stage 2 test set, this result suggests that explicit inheritance of the rule program and probe evidence helps limit forgetting while supporting adaptation to the new SOFA-based feature space.

The Stage 2 gain was class asymmetric. Continual MHL increased Sensitivity from 0.812 to 0.891 but reduced Specificity from 0.370 to 0.287, leaving ACC essentially unchanged at 0.589 versus 0.591 for direct Stage 2 training. Its higher F1 therefore reflects a more recall-oriented rule program rather than uniform improvement across classes. Even with this qualification, MHL attained the highest F1 in the controlled Stage 2 comparison while preserving a traceable code-level revision path from SIRS-based to SOFA-based evidence. The combination of predictive performance, evidence of prior-knowledge retention, and explicit knowledge maintenance is the principal advantage demonstrated here.

\section{Discussion}
\label{sec:6}

Across the probe-ablation study, the LLM-backend study, the robustness studies on training-set size and severe class imbalance, and the continual-adaptation study under feature evolution, Medical Heuristic Learning achieved performance comparable to strong state-of-the-art baselines while preserving a fully white-box, interpretable, and auditable model form. This finding is important because it challenges a common assumption in clinical prediction that strong predictive performance necessarily requires black-box models, whereas interpretable systems must sacrifice expressive power. The advantage of MHL is most apparent in the settings that are often most difficult in practice, namely small training samples and severely imbalanced class distributions. Under these conditions, strong baselines such as XGBoost~\cite{chen2016xgboost}, LightGBM~\cite{ke2017lightgbm}, FT-Transformer~\cite{gorishniy2021revisiting}, and MLP~\cite{rumelhart1986learning} were more prone to performance degradation or one-sided prediction behavior, whereas MHL retained a more usable predictive profile. In larger and more balanced training settings, conventional models remained highly competitive and in some cases achieved stronger numerical results. Even so, MHL remained within a similar performance range while maintaining a substantially more interpretable model form. Its value therefore lies not only in predictive accuracy, but also in the fact that the final model remains readable, executable, and open to direct inspection.

This behavior is closely related to the way MHL formulates learning as constrained search over explicit rule programs rather than as parameter fitting in a hidden representation space. As defined in the Methods section, the target object is an executable rule program with a prescribed grammar and prediction semantics, while the LLM functions only as a constrained operator for initialization and local revision. The statistical probe and the medical knowledge probe provide complementary evidence for constructing the initial program, and error feedback, degradation feedback, and validation-based lexicographic selection govern how later versions are revised and retained. Such a design is especially relevant in low-sample and class-imbalanced regimes, where models driven mainly by gradient optimization or greedy partitioning may overfit accidental correlations or become dominated by majority-class signals. In MHL, the statistical probe anchors rule construction in observed distributional regularities, whereas the medical knowledge probe contributes clinical semantics, candidate thresholds, and prior expectations about plausible feature-outcome relations. Because each accepted update is a bounded modification to the current valid program, MHL operates as a trajectory-based single-solution metaheuristic with retrospective version selection over a discrete rule-program space rather than as a population-based search procedure or a monolithic rewrite process. The resulting decision logic remains explicit throughout the version trajectory, making local bias, failure modes, and regression-inducing edits easier to inspect and correct. The robustness observed in the experiments therefore appears to arise from the interaction among explicit rule-program representation, complementary evidence acquisition, constrained trajectory-based search, and retrospective selection over retained versions, rather than from any single component in isolation.

The interpretability of MHL also differs from both post-hoc explanation methods and other inherently interpretable models. SHAP and LIME explain a trained black-box model after the fact \cite{lundberg2017unified,ribeiro2016trust}. Although useful, these methods do not expose the executable logic of the predictor itself and may depend on background distributions, perturbation strategies, and modeling assumptions \cite{rudin2019stop,tjoa2021survey}. Linear models, shallow decision trees, EBM \cite{lou2013ga2m,nori2019interpretml}, and APLR \cite{vonottenbreit2024automatic} are transparent in form, but they mainly expose coefficients, splits, or additive effects rather than a clinically grounded rule-construction and revision process. Rule-list learners such as CORELS \cite{angelino2018corels} are closer to MHL because they also return symbolic logic, yet MHL adds medically informed threshold suggestions, explicit error and degradation feedback, and versioned rule refinement. This difference is especially valuable in medical tasks, where clinically meaningful thresholds, asymmetric error costs, and evolving measurement conventions make model governance as important as one-shot prediction. The sample-size results are consistent with this advantage, since MHL outperformed CORELS in the most data-constrained settings. MHL therefore offers more than a readable final model. It provides an executable rule program together with the statistical evidence, medical rationale, error analyses, degradation warnings, and version history that explain why the rule entered the model and how it changed over time. In this sense, the model's reasoning and maintenance process is exposed to scrutiny in a way that post-hoc explanations, additive glass-box models, and conventional rule-list learners do not fully provide.

This same workflow also serves as a practical safeguard against LLM hallucination. MHL does not treat unconstrained textual reasoning as the final product. Instead, each candidate rule must be realized as deterministic, executable code, instantiated as Python rule code in this study, informed by observed statistical summaries and clinically motivated threshold suggestions, and then subjected to repeated evaluation through execution. In this sense, the framework relies on empirical guardrails derived from execution feedback. Thresholds or rationales that are clinically implausible, poorly calibrated, or weakly supported by the data are unlikely to remain concealed, because they typically appear as concentrated training errors, systematically one sided predictions, or clear degradation relative to the previously accepted version. These failures are captured in structured error reports and degradation warnings, which directly inform the next constrained revision step. The requirement that each revision remain small further limits uncontrolled drift by favoring local adjustments over wholesale rewrites. Overall, the current safeguard in MHL is an auditable closed loop in which candidate logic must withstand structured empirical scrutiny before it is retained, thereby reducing the likelihood that clinically unreasonable hallucinations persist unnoticed across versions.

The continual learning experiment extends this argument from static prediction to model governance under feature evolution. Clinical data environments are inherently dynamic. Measurements, scoring systems, disease definitions, practice guidelines, and feature distributions may all change as medical knowledge and care delivery evolve. For conventional models, feature removal, feature addition, and distributional shift together create a difficult adaptation problem. Even when a model can be retrained or transferred, it is often unclear how prior knowledge has been preserved, how newly introduced variables alter decision logic, or which parts of the previous model have effectively been overwritten. MHL offers a more transparent adaptation pathway. It begins from the previous rule base, explicitly records removed, added, and retained features, and incorporates new evidence through renewed probes and code-level rule revision. This mechanism helps explain why MHL can retain useful performance under feature evolution and avoid performance collapse even under an abrupt feature transition with very limited new samples, because adaptation is guided simultaneously by previously validated logic and by newly observed statistical and clinical signals. At the same time, the adapted model remains interpretable because changes appear as visible modifications to rule code rather than as hidden shifts in parameters. In high-stakes clinical settings, this distinction matters because model updating is not only a question of predictive performance, but also one of version control, accountability, and risk review.

Several boundaries of MHL should also be made explicit. First, the present evaluation focused on binary prediction tasks with structured clinical variables. Whether the same workflow can be extended to imaging, free text, multimodal prediction, multiclass outcomes, survival analysis, or complex longitudinal modeling remains to be established. Second, explicit rule systems may face scalability challenges when the feature space becomes very high-dimensional or when clinically relevant interactions are highly complex. In such settings, rule combinations may become difficult to read, maintain, or validate. Third, although MHL does not use the LLM as a direct classifier, the choice of LLM backend can still influence candidate rules, decision style, and revision trajectories, as indicated by the backend comparison experiment. Finally, the evidence reported here is based on offline evaluation, retrospective datasets, and a simulated feature evolution setting. These findings therefore cannot substitute for prospective, multicenter validation in real clinical workflows. White-box auditability increases the possibility of review, but it does not in itself guarantee causal validity, clinical safety, or readiness for deployment.

Overall, MHL suggests an LLM-based alternative to conventional black-box learning for clinical prediction, one that integrates medical priors, data evidence, executable rules, and explicit updating within a unified white-box framework. Its significance is not limited to interpretability alone. More broadly, it addresses several requirements that are central to high-risk digital medicine, including competitive performance, auditability, and the capacity to adapt as clinical data structures evolve. Future work should evaluate this framework across more complex tasks, external cohorts, and prospective clinical settings, while also developing principled methods for controlling rule complexity and governing rule updates over time. In the longer term, LLMs should remain assistive components that help generate auditable rule candidates rather than substitutes for clinical or statistical judgment. The broader implication of this study is that, in medical prediction from structured data, performance, interpretability, auditability, and adaptability need not be treated as mutually exclusive goals. They can be pursued together within a white-box learning framework designed for clinical accountability.

\section{Conclusion}
\label{sec:7}

In this study, we proposed Medical Heuristic Learning (MHL), an LLM-driven white-box framework for constructing interpretable and auditable clinical decision rules. Rather than relying on hidden parameter updates, MHL integrates statistical probes, medical knowledge probes, constrained rule synthesis, and iterative code-level refinement into a unified workflow in which the final model is represented as executable expert rules. In this way, the framework instantiates the learning beyond gradients paradigm in a clinically grounded setting and links modern LLM-assisted generation with rule-based expert system design.

Across multiple medical prediction tasks, MHL achieved performance comparable to strong black-box baselines while preserving a transparent and directly inspectable model form. Its advantages were especially evident in practically challenging settings, including limited training data, severe class imbalance, and feature evolution over time. Under these conditions, conventional models were more susceptible to unstable fitting, majority-class bias, and difficult adaptation after shifts in the available feature space, whereas MHL maintained a more usable modeling process through explicit rule construction and controlled rule revision. This study shows that predictive performance and interpretability need not be treated as mutually exclusive goals in high-risk medical prediction. More broadly, the contribution of this work lies not only in a medical application case, but also in a paradigm for building and maintaining highly interpretable modern expert systems under real-world constraints such as limited samples, class imbalance, and feature evolution.

\section*{Acknowledgments}
This work is supported by Macao Polytechnic University (RP/FCA-14/2023), the Macao Science and Technology Development Fund (FDCT, 0033/2023/RIB2, 0062/2025/RIB2), Joint Fund between FDCT and the Department of Science and Technology of Guangdong Province (FDCT-GDST, 0009/2024/AGJ), and Joint Fund between FDCT and the Ministry of Science and Technology of China (FDCT-MOST, 0106/2025/AMJ). The Macao Polytechnic University submission approval ID is fca.f673.ffe3.8.

The co-corresponding author in this study had obtained certification through the Collaborative Institutional Training Initiative (CITI program) for international cooperation and training. This research has been conducted using the UK Biobank Resource under Application Number 99946. The study protocol has been reviewed and approved by the Institutional Review Board (IRB) of Yantai Yuhuangding Hospital (file number: 2024-018).

\section*{Declaration of generative AI and AI-assisted technologies in the manuscript preparation process}
During the preparation of this work, the authors used GPT-5.4 for language polishing. After using this tool, the authors reviewed and edited the content as needed and take full responsibility for the content of the published article.

\appendix
\section{Supplementary Tables}
\label{app:supplementary-tables}
% Auto-generated from hl-data-process/ablation/ablation_summary.csv.
% The internal dataset identifier YHD is rendered as CCID to match the manuscript terminology.
\begingroup
\small
\setlength{\LTleft}{0pt}
\setlength{\LTright}{0pt}
\begin{longtable}{@{}lllrrrr@{}}
\caption{Ablation summary across probe settings.}\label{tab:results-2-3-ablation} \\
\toprule
Dataset & Probe & Train $n$ & ACC & F1 & Sens. & Spec. \\
\midrule
\endfirsthead
\multicolumn{7}{l}{\textit{Continued from previous page}} \\
\toprule
Dataset & Probe & Train $n$ & ACC & F1 & Sens. & Spec. \\
\midrule
\endhead
\midrule
\multicolumn{7}{r}{\textit{Continued on next page}} \\
\endfoot
\bottomrule
\endlastfoot
UKB & N & 10 & 0.503 & 0.623 & 0.837 & 0.169 \\
UKB & N & 100 & 0.511 & 0.547 & 0.736 & 0.286 \\
UKB & N & 1000 & 0.531 & 0.528 & 0.522 & 0.540 \\
UKB & N & 3000 & 0.525 & 0.547 & 0.629 & 0.420 \\
UKB & K & 10 & 0.556 & 0.602 & 0.683 & 0.429 \\
UKB & K & 100 & 0.529 & 0.654 & 0.891 & 0.167 \\
UKB & K & 1000 & 0.515 & 0.627 & 0.823 & 0.207 \\
UKB & K & 3000 & 0.547 & 0.583 & 0.643 & 0.451 \\
UKB & S & 10 & 0.526 & 0.524 & 0.533 & 0.520 \\
UKB & S & 100 & 0.561 & 0.575 & 0.597 & 0.525 \\
UKB & S & 1000 & 0.573 & 0.643 & 0.769 & 0.377 \\
UKB & S & 3000 & 0.585 & 0.613 & 0.657 & 0.513 \\
UKB & S+K & 10 & 0.533 & 0.623 & 0.781 & 0.285 \\
UKB & S+K & 100 & 0.548 & 0.587 & 0.645 & 0.451 \\
UKB & S+K & 1000 & 0.583 & 0.603 & 0.635 & 0.531 \\
UKB & S+K & 3000 & 0.561 & 0.621 & 0.723 & 0.399 \\
CCID & N & 10 & 0.570 & 0.639 & 0.765 & 0.375 \\
CCID & N & 100 & 0.667 & 0.671 & 0.676 & 0.658 \\
CCID & N & 1000 & 0.558 & 0.682 & 0.948 & 0.168 \\
CCID & N & 3000 & 0.554 & 0.677 & 0.931 & 0.177 \\
CCID & K & 10 & 0.583 & 0.685 & 0.907 & 0.259 \\
CCID & K & 100 & 0.618 & 0.681 & 0.819 & 0.418 \\
CCID & K & 1000 & 0.594 & 0.689 & 0.900 & 0.287 \\
CCID & K & 3000 & 0.629 & 0.693 & 0.832 & 0.427 \\
CCID & S & 10 & 0.571 & 0.601 & 0.660 & 0.482 \\
CCID & S & 100 & 0.665 & 0.707 & 0.811 & 0.518 \\
CCID & S & 1000 & 0.707 & 0.675 & 0.627 & 0.787 \\
CCID & S & 3000 & 0.712 & 0.716 & 0.726 & 0.699 \\
CCID & S+K & 10 & 0.641 & 0.685 & 0.786 & 0.497 \\
CCID & S+K & 100 & 0.700 & 0.710 & 0.735 & 0.665 \\
CCID & S+K & 1000 & 0.708 & 0.728 & 0.783 & 0.632 \\
CCID & S+K & 3000 & 0.705 & 0.727 & 0.785 & 0.625 \\
\end{longtable}
\endgroup

% Auto-generated from hl-data-process/contrast1/contrast1_summary.csv.
% Regenerate with hl-data-process/generate_results_latex_tables.py; do not edit numeric values manually.
% The internal CSV dataset identifier YHD is rendered as CCID to match the manuscript terminology.
\begingroup
\small
\setlength{\LTleft}{0pt}
\setlength{\LTright}{0pt}
\begin{longtable}{@{}llrrrrr@{}}
\caption{Mean test performance across random seeds 36, 40, and 42 for different training-set sizes. LightGBM at $n=10$ collapsed to all-positive prediction on both datasets; its reported metrics describe that failure rather than a valid competitive solution.}\label{tab:results-2-4-sample-size} \\
\toprule
Dataset & Model & Train $n$ & ACC & F1 & Sens. & Spec. \\
\midrule
\endfirsthead
\multicolumn{7}{l}{\textit{Continued from previous page}} \\
\toprule
Dataset & Model & Train $n$ & ACC & F1 & Sens. & Spec. \\
\midrule
\endhead
\midrule
\multicolumn{7}{r}{\textit{Continued on next page}} \\
\endfoot
\bottomrule
\endlastfoot
UKB & Logistic Regression & 10 & 0.555 & 0.520 & 0.506 & 0.604 \\
UKB & Logistic Regression & 50 & 0.524 & 0.480 & 0.441 & 0.607 \\
UKB & Logistic Regression & 100 & 0.559 & 0.567 & 0.578 & 0.540 \\
UKB & Logistic Regression & 500 & 0.596 & 0.597 & 0.600 & 0.591 \\
UKB & Logistic Regression & 1000 & 0.619 & 0.619 & 0.621 & 0.616 \\
UKB & Logistic Regression & 3000 & 0.633 & 0.625 & 0.613 & 0.652 \\
UKB & Decision Tree & 10 & 0.524 & 0.512 & 0.507 & 0.540 \\
UKB & Decision Tree & 50 & 0.537 & 0.506 & 0.480 & 0.594 \\
UKB & Decision Tree & 100 & 0.525 & 0.537 & 0.553 & 0.496 \\
UKB & Decision Tree & 500 & 0.543 & 0.544 & 0.545 & 0.541 \\
UKB & Decision Tree & 1000 & 0.542 & 0.539 & 0.536 & 0.549 \\
UKB & Decision Tree & 3000 & 0.591 & 0.594 & 0.599 & 0.584 \\
UKB & MLP & 10 & 0.547 & 0.498 & 0.483 & 0.611 \\
UKB & MLP & 50 & 0.547 & 0.483 & 0.429 & 0.665 \\
UKB & MLP & 100 & 0.535 & 0.568 & 0.612 & 0.457 \\
UKB & MLP & 500 & 0.592 & 0.598 & 0.609 & 0.575 \\
UKB & MLP & 1000 & 0.594 & 0.610 & 0.635 & 0.552 \\
UKB & MLP & 3000 & 0.633 & 0.633 & 0.632 & 0.635 \\
UKB & XGBoost & 10 & 0.529 & 0.523 & 0.531 & 0.527 \\
UKB & XGBoost & 50 & 0.567 & 0.552 & 0.534 & 0.601 \\
UKB & XGBoost & 100 & 0.577 & 0.580 & 0.586 & 0.569 \\
UKB & XGBoost & 500 & 0.620 & 0.624 & 0.629 & 0.612 \\
UKB & XGBoost & 1000 & 0.633 & 0.633 & 0.633 & 0.633 \\
UKB & XGBoost & 3000 & 0.671 & 0.669 & 0.667 & 0.675 \\
UKB & LightGBM & 10 & 0.500 & 0.667 & 1.000 & 0.000 \\
UKB & LightGBM & 50 & 0.548 & 0.525 & 0.501 & 0.595 \\
UKB & LightGBM & 100 & 0.562 & 0.558 & 0.556 & 0.567 \\
UKB & LightGBM & 500 & 0.605 & 0.612 & 0.622 & 0.589 \\
UKB & LightGBM & 1000 & 0.634 & 0.633 & 0.633 & 0.635 \\
UKB & LightGBM & 3000 & 0.688 & 0.688 & 0.687 & 0.689 \\
UKB & FT-Transformer & 10 & 0.533 & 0.571 & 0.622 & 0.444 \\
UKB & FT-Transformer & 50 & 0.568 & 0.583 & 0.606 & 0.530 \\
UKB & FT-Transformer & 100 & 0.547 & 0.536 & 0.523 & 0.572 \\
UKB & FT-Transformer & 500 & 0.605 & 0.597 & 0.595 & 0.615 \\
UKB & FT-Transformer & 1000 & 0.608 & 0.610 & 0.616 & 0.599 \\
UKB & FT-Transformer & 3000 & 0.643 & 0.638 & 0.629 & 0.656 \\
UKB & ResNet & 10 & 0.525 & 0.450 & 0.410 & 0.639 \\
UKB & ResNet & 50 & 0.566 & 0.531 & 0.499 & 0.634 \\
UKB & ResNet & 100 & 0.560 & 0.582 & 0.615 & 0.505 \\
UKB & ResNet & 500 & 0.595 & 0.585 & 0.571 & 0.620 \\
UKB & ResNet & 1000 & 0.627 & 0.622 & 0.614 & 0.639 \\
UKB & ResNet & 3000 & 0.630 & 0.620 & 0.608 & 0.652 \\
UKB & EBM & 10 & 0.531 & 0.506 & 0.521 & 0.542 \\
UKB & EBM & 50 & 0.548 & 0.521 & 0.494 & 0.603 \\
UKB & EBM & 100 & 0.570 & 0.581 & 0.598 & 0.543 \\
UKB & EBM & 500 & 0.624 & 0.622 & 0.619 & 0.629 \\
UKB & EBM & 1000 & 0.630 & 0.628 & 0.627 & 0.632 \\
UKB & EBM & 3000 & 0.652 & 0.651 & 0.651 & 0.653 \\
UKB & APLR & 10 & 0.532 & 0.582 & 0.664 & 0.401 \\
UKB & APLR & 50 & 0.558 & 0.545 & 0.529 & 0.587 \\
UKB & APLR & 100 & 0.558 & 0.579 & 0.615 & 0.501 \\
UKB & APLR & 500 & 0.599 & 0.616 & 0.643 & 0.555 \\
UKB & APLR & 1000 & 0.620 & 0.595 & 0.564 & 0.676 \\
UKB & APLR & 3000 & 0.652 & 0.632 & 0.599 & 0.705 \\
UKB & CORELS & 10 & 0.478 & 0.411 & 0.376 & 0.581 \\
UKB & CORELS & 50 & 0.535 & 0.527 & 0.539 & 0.531 \\
UKB & CORELS & 100 & 0.570 & 0.557 & 0.544 & 0.595 \\
UKB & CORELS & 500 & 0.591 & 0.604 & 0.631 & 0.551 \\
UKB & CORELS & 1000 & 0.602 & 0.571 & 0.533 & 0.671 \\
UKB & CORELS & 3000 & 0.598 & 0.499 & 0.400 & 0.795 \\
UKB & MHL & 10 & 0.533 & 0.623 & 0.781 & 0.285 \\
UKB & MHL & 50 & 0.534 & 0.597 & 0.716 & 0.351 \\
UKB & MHL & 100 & 0.548 & 0.587 & 0.645 & 0.451 \\
UKB & MHL & 500 & 0.537 & 0.620 & 0.765 & 0.308 \\
UKB & MHL & 1000 & 0.583 & 0.603 & 0.635 & 0.531 \\
UKB & MHL & 3000 & 0.561 & 0.621 & 0.723 & 0.399 \\
CCID & Logistic Regression & 10 & 0.573 & 0.531 & 0.495 & 0.651 \\
CCID & Logistic Regression & 50 & 0.671 & 0.655 & 0.627 & 0.715 \\
CCID & Logistic Regression & 100 & 0.704 & 0.702 & 0.698 & 0.711 \\
CCID & Logistic Regression & 500 & 0.746 & 0.735 & 0.703 & 0.789 \\
CCID & Logistic Regression & 1000 & 0.752 & 0.736 & 0.694 & 0.809 \\
CCID & Logistic Regression & 3000 & 0.758 & 0.738 & 0.681 & 0.835 \\
CCID & Decision Tree & 10 & 0.537 & 0.471 & 0.428 & 0.646 \\
CCID & Decision Tree & 50 & 0.644 & 0.625 & 0.603 & 0.684 \\
CCID & Decision Tree & 100 & 0.667 & 0.672 & 0.678 & 0.657 \\
CCID & Decision Tree & 500 & 0.657 & 0.661 & 0.669 & 0.646 \\
CCID & Decision Tree & 1000 & 0.674 & 0.680 & 0.695 & 0.653 \\
CCID & Decision Tree & 3000 & 0.657 & 0.600 & 0.515 & 0.800 \\
CCID & MLP & 10 & 0.565 & 0.525 & 0.490 & 0.640 \\
CCID & MLP & 50 & 0.662 & 0.640 & 0.602 & 0.723 \\
CCID & MLP & 100 & 0.673 & 0.676 & 0.686 & 0.660 \\
CCID & MLP & 500 & 0.699 & 0.724 & 0.785 & 0.614 \\
CCID & MLP & 1000 & 0.757 & 0.745 & 0.712 & 0.801 \\
CCID & MLP & 3000 & 0.720 & 0.673 & 0.579 & 0.860 \\
CCID & XGBoost & 10 & 0.519 & 0.494 & 0.489 & 0.548 \\
CCID & XGBoost & 50 & 0.666 & 0.660 & 0.654 & 0.678 \\
CCID & XGBoost & 100 & 0.697 & 0.693 & 0.688 & 0.705 \\
CCID & XGBoost & 500 & 0.746 & 0.748 & 0.753 & 0.739 \\
CCID & XGBoost & 1000 & 0.769 & 0.770 & 0.772 & 0.767 \\
CCID & XGBoost & 3000 & 0.752 & 0.711 & 0.611 & 0.893 \\
CCID & LightGBM & 10 & 0.500 & 0.667 & 1.000 & 0.000 \\
CCID & LightGBM & 50 & 0.603 & 0.603 & 0.605 & 0.601 \\
CCID & LightGBM & 100 & 0.685 & 0.678 & 0.667 & 0.702 \\
CCID & LightGBM & 500 & 0.740 & 0.741 & 0.745 & 0.736 \\
CCID & LightGBM & 1000 & 0.762 & 0.758 & 0.749 & 0.775 \\
CCID & LightGBM & 3000 & 0.749 & 0.704 & 0.599 & 0.899 \\
CCID & FT-Transformer & 10 & 0.547 & 0.523 & 0.499 & 0.595 \\
CCID & FT-Transformer & 50 & 0.661 & 0.636 & 0.599 & 0.723 \\
CCID & FT-Transformer & 100 & 0.712 & 0.717 & 0.727 & 0.698 \\
CCID & FT-Transformer & 500 & 0.747 & 0.745 & 0.741 & 0.752 \\
CCID & FT-Transformer & 1000 & 0.743 & 0.737 & 0.720 & 0.765 \\
CCID & FT-Transformer & 3000 & 0.744 & 0.730 & 0.695 & 0.793 \\
CCID & ResNet & 10 & 0.545 & 0.541 & 0.571 & 0.518 \\
CCID & ResNet & 50 & 0.681 & 0.673 & 0.657 & 0.706 \\
CCID & ResNet & 100 & 0.661 & 0.647 & 0.622 & 0.700 \\
CCID & ResNet & 500 & 0.744 & 0.748 & 0.760 & 0.728 \\
CCID & ResNet & 1000 & 0.751 & 0.746 & 0.738 & 0.763 \\
CCID & ResNet & 3000 & 0.744 & 0.740 & 0.733 & 0.755 \\
CCID & EBM & 10 & 0.582 & 0.566 & 0.551 & 0.613 \\
CCID & EBM & 50 & 0.703 & 0.688 & 0.657 & 0.748 \\
CCID & EBM & 100 & 0.741 & 0.733 & 0.710 & 0.772 \\
CCID & EBM & 500 & 0.756 & 0.756 & 0.752 & 0.761 \\
CCID & EBM & 1000 & 0.765 & 0.761 & 0.751 & 0.779 \\
CCID & EBM & 3000 & 0.739 & 0.703 & 0.617 & 0.862 \\
CCID & APLR & 10 & 0.574 & 0.509 & 0.453 & 0.694 \\
CCID & APLR & 50 & 0.717 & 0.702 & 0.671 & 0.763 \\
CCID & APLR & 100 & 0.715 & 0.687 & 0.628 & 0.802 \\
CCID & APLR & 500 & 0.741 & 0.731 & 0.705 & 0.778 \\
CCID & APLR & 1000 & 0.753 & 0.733 & 0.681 & 0.825 \\
CCID & APLR & 3000 & 0.758 & 0.741 & 0.694 & 0.821 \\
CCID & CORELS & 10 & 0.509 & 0.520 & 0.557 & 0.460 \\
CCID & CORELS & 50 & 0.659 & 0.649 & 0.627 & 0.691 \\
CCID & CORELS & 100 & 0.708 & 0.680 & 0.621 & 0.795 \\
CCID & CORELS & 500 & 0.712 & 0.668 & 0.585 & 0.839 \\
CCID & CORELS & 1000 & 0.712 & 0.672 & 0.593 & 0.831 \\
CCID & CORELS & 3000 & 0.713 & 0.670 & 0.584 & 0.842 \\
CCID & MHL & 10 & 0.641 & 0.685 & 0.786 & 0.497 \\
CCID & MHL & 50 & 0.680 & 0.709 & 0.780 & 0.579 \\
CCID & MHL & 100 & 0.700 & 0.710 & 0.735 & 0.665 \\
CCID & MHL & 500 & 0.608 & 0.690 & 0.871 & 0.344 \\
CCID & MHL & 1000 & 0.708 & 0.728 & 0.783 & 0.632 \\
CCID & MHL & 3000 & 0.705 & 0.727 & 0.785 & 0.625 \\
\end{longtable}
\endgroup

% Auto-generated from hl-data-process/contrast2_confusion_matrix/contrast2_confusion_matrix_summary.csv.
% Regenerate with hl-data-process/generate_results_latex_tables.py; do not edit numeric values manually.
% The internal CSV dataset identifier YHD is rendered as CCID to match the manuscript terminology.
\begingroup
\scriptsize
\setlength{\LTleft}{0pt}
\setlength{\LTright}{0pt}
\setlength{\tabcolsep}{2.8pt}
\begin{longtable}{@{}lrlcrrrrrrrr@{}}
\caption{Mean test metrics and pooled confusion counts across random seeds 36, 40, and 42 under different training-set class ratios. Each seed uses a balanced 1,000-sample test set; TP, FP, FN, and TN are therefore pooled over 3,000 test predictions.}\label{tab:results-2-5-imbalance} \\
\toprule
Dataset & Train $n$ & Model & Pos:Neg & ACC & F1 & Sens. & Spec. & TP & FP & FN & TN \\
\midrule
\endfirsthead
\multicolumn{12}{l}{\textit{Continued from previous page}} \\
\toprule
Dataset & Train $n$ & Model & Pos:Neg & ACC & F1 & Sens. & Spec. & TP & FP & FN & TN \\
\midrule
\endhead
\midrule
\multicolumn{12}{r}{\textit{Continued on next page}} \\
\endfoot
\bottomrule
\endlastfoot
UKB & 1000 & Logistic Regression & 50:1 & 0.500 & 0.666 & 0.999 & 0.001 & 1498 & 1499 & 2 & 1 \\
UKB & 1000 & Logistic Regression & 10:1 & 0.501 & 0.666 & 0.996 & 0.006 & 1494 & 1491 & 6 & 9 \\
UKB & 1000 & Logistic Regression & 5:1 & 0.508 & 0.666 & 0.983 & 0.032 & 1475 & 1452 & 25 & 48 \\
UKB & 1000 & Logistic Regression & 2:1 & 0.568 & 0.671 & 0.881 & 0.256 & 1321 & 1116 & 179 & 384 \\
UKB & 1000 & Logistic Regression & 1:1 & 0.618 & 0.611 & 0.601 & 0.634 & 902 & 549 & 598 & 951 \\
UKB & 1000 & Logistic Regression & 1:2 & 0.566 & 0.357 & 0.243 & 0.888 & 365 & 168 & 1135 & 1332 \\
UKB & 1000 & Logistic Regression & 1:5 & 0.527 & 0.126 & 0.069 & 0.985 & 103 & 23 & 1397 & 1477 \\
UKB & 1000 & Logistic Regression & 1:10 & 0.513 & 0.065 & 0.034 & 0.991 & 51 & 13 & 1449 & 1487 \\
UKB & 1000 & Logistic Regression & 1:50 & 0.502 & 0.013 & 0.007 & 0.997 & 10 & 5 & 1490 & 1495 \\
UKB & 1000 & Decision Tree & 50:1 & 0.509 & 0.663 & 0.967 & 0.051 & 1451 & 1424 & 49 & 76 \\
UKB & 1000 & Decision Tree & 10:1 & 0.516 & 0.648 & 0.890 & 0.142 & 1335 & 1287 & 165 & 213 \\
UKB & 1000 & Decision Tree & 5:1 & 0.528 & 0.635 & 0.823 & 0.234 & 1234 & 1149 & 266 & 351 \\
UKB & 1000 & Decision Tree & 2:1 & 0.564 & 0.613 & 0.693 & 0.435 & 1039 & 848 & 461 & 652 \\
UKB & 1000 & Decision Tree & 1:1 & 0.562 & 0.562 & 0.564 & 0.559 & 846 & 661 & 654 & 839 \\
UKB & 1000 & Decision Tree & 1:2 & 0.550 & 0.473 & 0.404 & 0.695 & 606 & 457 & 894 & 1043 \\
UKB & 1000 & Decision Tree & 1:5 & 0.546 & 0.369 & 0.267 & 0.825 & 401 & 263 & 1099 & 1237 \\
UKB & 1000 & Decision Tree & 1:10 & 0.531 & 0.271 & 0.174 & 0.889 & 261 & 167 & 1239 & 1333 \\
UKB & 1000 & Decision Tree & 1:50 & 0.507 & 0.074 & 0.039 & 0.974 & 59 & 39 & 1441 & 1461 \\
UKB & 1000 & MLP & 50:1 & 0.500 & 0.667 & 1.000 & 0.000 & 1500 & 1500 & 0 & 0 \\
UKB & 1000 & MLP & 10:1 & 0.500 & 0.667 & 1.000 & 0.000 & 1500 & 1500 & 0 & 0 \\
UKB & 1000 & MLP & 5:1 & 0.501 & 0.667 & 0.998 & 0.004 & 1497 & 1494 & 3 & 6 \\
UKB & 1000 & MLP & 2:1 & 0.544 & 0.676 & 0.951 & 0.137 & 1426 & 1295 & 74 & 205 \\
UKB & 1000 & MLP & 1:1 & 0.602 & 0.609 & 0.621 & 0.583 & 932 & 626 & 568 & 874 \\
UKB & 1000 & MLP & 1:2 & 0.554 & 0.319 & 0.229 & 0.878 & 344 & 183 & 1156 & 1317 \\
UKB & 1000 & MLP & 1:5 & 0.504 & 0.021 & 0.011 & 0.998 & 16 & 3 & 1484 & 1497 \\
UKB & 1000 & MLP & 1:10 & 0.500 & 0.000 & 0.000 & 1.000 & 0 & 0 & 1500 & 1500 \\
UKB & 1000 & MLP & 1:50 & 0.500 & 0.000 & 0.000 & 1.000 & 0 & 0 & 1500 & 1500 \\
UKB & 1000 & XGBoost & 50:1 & 0.500 & 0.667 & 1.000 & 0.000 & 1500 & 1500 & 0 & 0 \\
UKB & 1000 & XGBoost & 10:1 & 0.507 & 0.669 & 0.995 & 0.018 & 1493 & 1473 & 7 & 27 \\
UKB & 1000 & XGBoost & 5:1 & 0.526 & 0.676 & 0.989 & 0.064 & 1483 & 1404 & 17 & 96 \\
UKB & 1000 & XGBoost & 2:1 & 0.596 & 0.684 & 0.876 & 0.316 & 1314 & 1026 & 186 & 474 \\
UKB & 1000 & XGBoost & 1:1 & 0.631 & 0.626 & 0.617 & 0.645 & 926 & 533 & 574 & 967 \\
UKB & 1000 & XGBoost & 1:2 & 0.600 & 0.455 & 0.334 & 0.865 & 501 & 202 & 999 & 1298 \\
UKB & 1000 & XGBoost & 1:5 & 0.519 & 0.109 & 0.059 & 0.979 & 89 & 32 & 1411 & 1468 \\
UKB & 1000 & XGBoost & 1:10 & 0.505 & 0.025 & 0.013 & 0.997 & 19 & 5 & 1481 & 1495 \\
UKB & 1000 & XGBoost & 1:50 & 0.500 & 0.001 & 0.001 & 1.000 & 1 & 0 & 1499 & 1500 \\
UKB & 1000 & LightGBM & 50:1 & 0.500 & 0.667 & 1.000 & 0.000 & 1500 & 1500 & 0 & 0 \\
UKB & 1000 & LightGBM & 10:1 & 0.504 & 0.668 & 0.997 & 0.011 & 1495 & 1483 & 5 & 17 \\
UKB & 1000 & LightGBM & 5:1 & 0.520 & 0.672 & 0.985 & 0.056 & 1477 & 1416 & 23 & 84 \\
UKB & 1000 & LightGBM & 2:1 & 0.596 & 0.684 & 0.874 & 0.319 & 1311 & 1022 & 189 & 478 \\
UKB & 1000 & LightGBM & 1:1 & 0.637 & 0.637 & 0.636 & 0.639 & 954 & 542 & 546 & 958 \\
UKB & 1000 & LightGBM & 1:2 & 0.608 & 0.472 & 0.351 & 0.864 & 527 & 204 & 973 & 1296 \\
UKB & 1000 & LightGBM & 1:5 & 0.515 & 0.097 & 0.052 & 0.978 & 78 & 33 & 1422 & 1467 \\
UKB & 1000 & LightGBM & 1:10 & 0.506 & 0.028 & 0.014 & 0.999 & 21 & 2 & 1479 & 1498 \\
UKB & 1000 & LightGBM & 1:50 & 0.500 & 0.000 & 0.000 & 1.000 & 0 & 0 & 1500 & 1500 \\
UKB & 1000 & FT-Transformer & 50:1 & 0.500 & 0.667 & 1.000 & 0.000 & 1500 & 1500 & 0 & 0 \\
UKB & 1000 & FT-Transformer & 10:1 & 0.500 & 0.667 & 0.999 & 0.001 & 1499 & 1498 & 1 & 2 \\
UKB & 1000 & FT-Transformer & 5:1 & 0.515 & 0.669 & 0.979 & 0.051 & 1469 & 1423 & 31 & 77 \\
UKB & 1000 & FT-Transformer & 2:1 & 0.557 & 0.668 & 0.891 & 0.223 & 1336 & 1166 & 164 & 334 \\
UKB & 1000 & FT-Transformer & 1:1 & 0.612 & 0.613 & 0.615 & 0.609 & 923 & 586 & 577 & 914 \\
UKB & 1000 & FT-Transformer & 1:2 & 0.592 & 0.490 & 0.393 & 0.791 & 589 & 314 & 911 & 1186 \\
UKB & 1000 & FT-Transformer & 1:5 & 0.517 & 0.095 & 0.056 & 0.978 & 84 & 33 & 1416 & 1467 \\
UKB & 1000 & FT-Transformer & 1:10 & 0.503 & 0.012 & 0.006 & 0.999 & 9 & 1 & 1491 & 1499 \\
UKB & 1000 & FT-Transformer & 1:50 & 0.500 & 0.000 & 0.000 & 1.000 & 0 & 0 & 1500 & 1500 \\
UKB & 1000 & ResNet & 50:1 & 0.500 & 0.667 & 1.000 & 0.000 & 1500 & 1500 & 0 & 0 \\
UKB & 1000 & ResNet & 10:1 & 0.506 & 0.667 & 0.992 & 0.019 & 1488 & 1471 & 12 & 29 \\
UKB & 1000 & ResNet & 5:1 & 0.515 & 0.667 & 0.970 & 0.059 & 1455 & 1411 & 45 & 89 \\
UKB & 1000 & ResNet & 2:1 & 0.573 & 0.667 & 0.855 & 0.291 & 1282 & 1064 & 218 & 436 \\
UKB & 1000 & ResNet & 1:1 & 0.615 & 0.618 & 0.626 & 0.603 & 939 & 595 & 561 & 905 \\
UKB & 1000 & ResNet & 1:2 & 0.579 & 0.408 & 0.292 & 0.865 & 438 & 202 & 1062 & 1298 \\
UKB & 1000 & ResNet & 1:5 & 0.532 & 0.166 & 0.094 & 0.971 & 141 & 44 & 1359 & 1456 \\
UKB & 1000 & ResNet & 1:10 & 0.517 & 0.091 & 0.049 & 0.986 & 73 & 21 & 1427 & 1479 \\
UKB & 1000 & ResNet & 1:50 & 0.500 & 0.000 & 0.000 & 1.000 & 0 & 0 & 1500 & 1500 \\
UKB & 1000 & EBM & 50:1 & 0.500 & 0.667 & 1.000 & 0.000 & 1500 & 1500 & 0 & 0 \\
UKB & 1000 & EBM & 10:1 & 0.504 & 0.668 & 0.999 & 0.009 & 1498 & 1486 & 2 & 14 \\
UKB & 1000 & EBM & 5:1 & 0.515 & 0.670 & 0.986 & 0.043 & 1479 & 1435 & 21 & 65 \\
UKB & 1000 & EBM & 2:1 & 0.599 & 0.681 & 0.855 & 0.344 & 1282 & 984 & 218 & 516 \\
UKB & 1000 & EBM & 1:1 & 0.640 & 0.637 & 0.635 & 0.645 & 952 & 533 & 548 & 967 \\
UKB & 1000 & EBM & 1:2 & 0.604 & 0.475 & 0.359 & 0.848 & 539 & 228 & 961 & 1272 \\
UKB & 1000 & EBM & 1:5 & 0.526 & 0.142 & 0.079 & 0.973 & 118 & 41 & 1382 & 1459 \\
UKB & 1000 & EBM & 1:10 & 0.504 & 0.017 & 0.009 & 0.999 & 13 & 2 & 1487 & 1498 \\
UKB & 1000 & EBM & 1:50 & 0.500 & 0.000 & 0.000 & 1.000 & 0 & 0 & 1500 & 1500 \\
UKB & 1000 & APLR & 50:1 & 0.501 & 0.667 & 1.000 & 0.003 & 1500 & 1496 & 0 & 4 \\
UKB & 1000 & APLR & 10:1 & 0.500 & 0.667 & 1.000 & 0.001 & 1500 & 1499 & 0 & 1 \\
UKB & 1000 & APLR & 5:1 & 0.500 & 0.667 & 0.999 & 0.001 & 1499 & 1498 & 1 & 2 \\
UKB & 1000 & APLR & 2:1 & 0.506 & 0.667 & 0.987 & 0.025 & 1481 & 1462 & 19 & 38 \\
UKB & 1000 & APLR & 1:1 & 0.627 & 0.625 & 0.627 & 0.627 & 941 & 560 & 559 & 940 \\
UKB & 1000 & APLR & 1:2 & 0.568 & 0.303 & 0.189 & 0.947 & 284 & 80 & 1216 & 1420 \\
UKB & 1000 & APLR & 1:5 & 0.506 & 0.034 & 0.018 & 0.995 & 27 & 8 & 1473 & 1492 \\
UKB & 1000 & APLR & 1:10 & 0.503 & 0.011 & 0.005 & 1.000 & 8 & 0 & 1492 & 1500 \\
UKB & 1000 & APLR & 1:50 & 0.500 & 0.003 & 0.001 & 0.999 & 2 & 1 & 1498 & 1499 \\
UKB & 1000 & CORELS & 50:1 & 0.500 & 0.000 & 0.000 & 1.000 & 0 & 0 & 1500 & 1500 \\
UKB & 1000 & CORELS & 10:1 & 0.500 & 0.000 & 0.000 & 1.000 & 0 & 0 & 1500 & 1500 \\
UKB & 1000 & CORELS & 5:1 & 0.500 & 0.000 & 0.000 & 1.000 & 0 & 0 & 1500 & 1500 \\
UKB & 1000 & CORELS & 2:1 & 0.522 & 0.665 & 0.948 & 0.097 & 1422 & 1355 & 78 & 145 \\
UKB & 1000 & CORELS & 1:1 & 0.598 & 0.590 & 0.581 & 0.614 & 872 & 579 & 628 & 921 \\
UKB & 1000 & CORELS & 1:2 & 0.559 & 0.287 & 0.185 & 0.933 & 277 & 100 & 1223 & 1400 \\
UKB & 1000 & CORELS & 1:5 & 0.500 & 0.000 & 0.000 & 1.000 & 0 & 0 & 1500 & 1500 \\
UKB & 1000 & CORELS & 1:10 & 0.500 & 0.000 & 0.000 & 1.000 & 0 & 0 & 1500 & 1500 \\
UKB & 1000 & CORELS & 1:50 & 0.500 & 0.000 & 0.000 & 1.000 & 0 & 0 & 1500 & 1500 \\
UKB & 1000 & MHL & 50:1 & 0.519 & 0.573 & 0.674 & 0.365 & 1011 & 953 & 489 & 547 \\
UKB & 1000 & MHL & 10:1 & 0.567 & 0.656 & 0.827 & 0.308 & 1240 & 1038 & 260 & 462 \\
UKB & 1000 & MHL & 5:1 & 0.561 & 0.661 & 0.851 & 0.272 & 1276 & 1092 & 224 & 408 \\
UKB & 1000 & MHL & 2:1 & 0.558 & 0.639 & 0.785 & 0.331 & 1178 & 1003 & 322 & 497 \\
UKB & 1000 & MHL & 1:1 & 0.583 & 0.603 & 0.635 & 0.531 & 952 & 704 & 548 & 796 \\
UKB & 1000 & MHL & 1:2 & 0.526 & 0.598 & 0.722 & 0.329 & 1083 & 1006 & 417 & 494 \\
UKB & 1000 & MHL & 1:5 & 0.561 & 0.579 & 0.640 & 0.482 & 960 & 777 & 540 & 723 \\
UKB & 1000 & MHL & 1:10 & 0.528 & 0.428 & 0.403 & 0.654 & 604 & 519 & 896 & 981 \\
UKB & 1000 & MHL & 1:50 & 0.563 & 0.442 & 0.377 & 0.749 & 565 & 377 & 935 & 1123 \\
UKB & 3000 & Logistic Regression & 50:1 & 0.500 & 0.667 & 1.000 & 0.000 & 1500 & 1500 & 0 & 0 \\
UKB & 3000 & Logistic Regression & 10:1 & 0.500 & 0.667 & 1.000 & 0.000 & 1500 & 1500 & 0 & 0 \\
UKB & 3000 & Logistic Regression & 5:1 & 0.505 & 0.668 & 0.996 & 0.014 & 1494 & 1479 & 6 & 21 \\
UKB & 3000 & Logistic Regression & 2:1 & 0.571 & 0.678 & 0.903 & 0.239 & 1355 & 1142 & 145 & 358 \\
UKB & 3000 & Logistic Regression & 1:1 & 0.626 & 0.621 & 0.611 & 0.641 & 917 & 538 & 583 & 962 \\
UKB & 3000 & Logistic Regression & 1:2 & 0.580 & 0.361 & 0.237 & 0.922 & 356 & 117 & 1144 & 1383 \\
UKB & 3000 & Logistic Regression & 1:5 & 0.515 & 0.071 & 0.037 & 0.993 & 56 & 10 & 1444 & 1490 \\
UKB & 3000 & Logistic Regression & 1:10 & 0.503 & 0.015 & 0.007 & 0.999 & 11 & 1 & 1489 & 1499 \\
UKB & 3000 & Logistic Regression & 1:50 & 0.500 & 0.001 & 0.001 & 0.999 & 1 & 1 & 1499 & 1499 \\
UKB & 3000 & Decision Tree & 50:1 & 0.507 & 0.665 & 0.979 & 0.035 & 1469 & 1447 & 31 & 53 \\
UKB & 3000 & Decision Tree & 10:1 & 0.528 & 0.652 & 0.884 & 0.173 & 1326 & 1241 & 174 & 259 \\
UKB & 3000 & Decision Tree & 5:1 & 0.549 & 0.647 & 0.828 & 0.270 & 1242 & 1095 & 258 & 405 \\
UKB & 3000 & Decision Tree & 2:1 & 0.571 & 0.624 & 0.713 & 0.429 & 1069 & 857 & 431 & 643 \\
UKB & 3000 & Decision Tree & 1:1 & 0.581 & 0.579 & 0.576 & 0.587 & 864 & 620 & 636 & 880 \\
UKB & 3000 & Decision Tree & 1:2 & 0.571 & 0.502 & 0.432 & 0.710 & 648 & 435 & 852 & 1065 \\
UKB & 3000 & Decision Tree & 1:5 & 0.566 & 0.399 & 0.288 & 0.844 & 432 & 234 & 1068 & 1266 \\
UKB & 3000 & Decision Tree & 1:10 & 0.530 & 0.262 & 0.167 & 0.894 & 250 & 159 & 1250 & 1341 \\
UKB & 3000 & Decision Tree & 1:50 & 0.506 & 0.080 & 0.043 & 0.969 & 65 & 46 & 1435 & 1454 \\
UKB & 3000 & MLP & 50:1 & 0.500 & 0.667 & 1.000 & 0.000 & 1500 & 1500 & 0 & 0 \\
UKB & 3000 & MLP & 10:1 & 0.500 & 0.667 & 1.000 & 0.000 & 1500 & 1500 & 0 & 0 \\
UKB & 3000 & MLP & 5:1 & 0.500 & 0.667 & 0.999 & 0.001 & 1499 & 1498 & 1 & 2 \\
UKB & 3000 & MLP & 2:1 & 0.572 & 0.675 & 0.891 & 0.253 & 1337 & 1121 & 163 & 379 \\
UKB & 3000 & MLP & 1:1 & 0.620 & 0.607 & 0.588 & 0.651 & 882 & 523 & 618 & 977 \\
UKB & 3000 & MLP & 1:2 & 0.589 & 0.402 & 0.285 & 0.893 & 427 & 160 & 1073 & 1340 \\
UKB & 3000 & MLP & 1:5 & 0.510 & 0.052 & 0.027 & 0.993 & 41 & 10 & 1459 & 1490 \\
UKB & 3000 & MLP & 1:10 & 0.500 & 0.000 & 0.000 & 1.000 & 0 & 0 & 1500 & 1500 \\
UKB & 3000 & MLP & 1:50 & 0.500 & 0.000 & 0.000 & 1.000 & 0 & 0 & 1500 & 1500 \\
UKB & 3000 & XGBoost & 50:1 & 0.500 & 0.667 & 1.000 & 0.000 & 1500 & 1500 & 0 & 0 \\
UKB & 3000 & XGBoost & 10:1 & 0.504 & 0.668 & 0.997 & 0.011 & 1496 & 1484 & 4 & 16 \\
UKB & 3000 & XGBoost & 5:1 & 0.539 & 0.680 & 0.982 & 0.095 & 1473 & 1357 & 27 & 143 \\
UKB & 3000 & XGBoost & 2:1 & 0.641 & 0.711 & 0.883 & 0.400 & 1324 & 900 & 176 & 600 \\
UKB & 3000 & XGBoost & 1:1 & 0.663 & 0.665 & 0.669 & 0.657 & 1004 & 514 & 496 & 986 \\
UKB & 3000 & XGBoost & 1:2 & 0.624 & 0.492 & 0.365 & 0.884 & 547 & 174 & 953 & 1326 \\
UKB & 3000 & XGBoost & 1:5 & 0.536 & 0.159 & 0.088 & 0.984 & 132 & 24 & 1368 & 1476 \\
UKB & 3000 & XGBoost & 1:10 & 0.507 & 0.031 & 0.016 & 0.999 & 24 & 2 & 1476 & 1498 \\
UKB & 3000 & XGBoost & 1:50 & 0.500 & 0.000 & 0.000 & 1.000 & 0 & 0 & 1500 & 1500 \\
UKB & 3000 & LightGBM & 50:1 & 0.500 & 0.667 & 1.000 & 0.000 & 1500 & 1500 & 0 & 0 \\
UKB & 3000 & LightGBM & 10:1 & 0.503 & 0.667 & 0.998 & 0.009 & 1497 & 1487 & 3 & 13 \\
UKB & 3000 & LightGBM & 5:1 & 0.545 & 0.684 & 0.985 & 0.105 & 1478 & 1342 & 22 & 158 \\
UKB & 3000 & LightGBM & 2:1 & 0.643 & 0.713 & 0.886 & 0.400 & 1329 & 900 & 171 & 600 \\
UKB & 3000 & LightGBM & 1:1 & 0.675 & 0.675 & 0.677 & 0.672 & 1016 & 492 & 484 & 1008 \\
UKB & 3000 & LightGBM & 1:2 & 0.629 & 0.505 & 0.379 & 0.879 & 569 & 181 & 931 & 1319 \\
UKB & 3000 & LightGBM & 1:5 & 0.545 & 0.192 & 0.108 & 0.982 & 162 & 27 & 1338 & 1473 \\
UKB & 3000 & LightGBM & 1:10 & 0.506 & 0.027 & 0.014 & 0.999 & 21 & 2 & 1479 & 1498 \\
UKB & 3000 & LightGBM & 1:50 & 0.500 & 0.000 & 0.000 & 1.000 & 0 & 0 & 1500 & 1500 \\
UKB & 3000 & FT-Transformer & 50:1 & 0.500 & 0.667 & 1.000 & 0.000 & 1500 & 1500 & 0 & 0 \\
UKB & 3000 & FT-Transformer & 10:1 & 0.501 & 0.667 & 0.999 & 0.003 & 1498 & 1495 & 2 & 5 \\
UKB & 3000 & FT-Transformer & 5:1 & 0.539 & 0.674 & 0.955 & 0.123 & 1432 & 1316 & 68 & 184 \\
UKB & 3000 & FT-Transformer & 2:1 & 0.622 & 0.674 & 0.781 & 0.462 & 1172 & 807 & 328 & 693 \\
UKB & 3000 & FT-Transformer & 1:1 & 0.638 & 0.630 & 0.617 & 0.658 & 926 & 513 & 574 & 987 \\
UKB & 3000 & FT-Transformer & 1:2 & 0.618 & 0.515 & 0.409 & 0.827 & 614 & 260 & 886 & 1240 \\
UKB & 3000 & FT-Transformer & 1:5 & 0.519 & 0.094 & 0.052 & 0.985 & 78 & 22 & 1422 & 1478 \\
UKB & 3000 & FT-Transformer & 1:10 & 0.502 & 0.013 & 0.007 & 0.998 & 10 & 3 & 1490 & 1497 \\
UKB & 3000 & FT-Transformer & 1:50 & 0.499 & 0.009 & 0.005 & 0.993 & 7 & 10 & 1493 & 1490 \\
UKB & 3000 & ResNet & 50:1 & 0.500 & 0.667 & 0.999 & 0.000 & 1499 & 1500 & 1 & 0 \\
UKB & 3000 & ResNet & 10:1 & 0.502 & 0.666 & 0.995 & 0.009 & 1492 & 1486 & 8 & 14 \\
UKB & 3000 & ResNet & 5:1 & 0.503 & 0.666 & 0.993 & 0.013 & 1489 & 1481 & 11 & 19 \\
UKB & 3000 & ResNet & 2:1 & 0.594 & 0.673 & 0.837 & 0.351 & 1256 & 974 & 244 & 526 \\
UKB & 3000 & ResNet & 1:1 & 0.623 & 0.602 & 0.571 & 0.675 & 856 & 488 & 644 & 1012 \\
UKB & 3000 & ResNet & 1:2 & 0.604 & 0.474 & 0.359 & 0.848 & 539 & 228 & 961 & 1272 \\
UKB & 3000 & ResNet & 1:5 & 0.535 & 0.165 & 0.093 & 0.977 & 140 & 35 & 1360 & 1465 \\
UKB & 3000 & ResNet & 1:10 & 0.511 & 0.053 & 0.027 & 0.994 & 41 & 9 & 1459 & 1491 \\
UKB & 3000 & ResNet & 1:50 & 0.500 & 0.004 & 0.002 & 0.998 & 3 & 3 & 1497 & 1497 \\
UKB & 3000 & EBM & 50:1 & 0.500 & 0.667 & 1.000 & 0.000 & 1500 & 1500 & 0 & 0 \\
UKB & 3000 & EBM & 10:1 & 0.503 & 0.668 & 0.999 & 0.007 & 1498 & 1490 & 2 & 10 \\
UKB & 3000 & EBM & 5:1 & 0.528 & 0.674 & 0.975 & 0.082 & 1462 & 1377 & 38 & 123 \\
UKB & 3000 & EBM & 2:1 & 0.630 & 0.699 & 0.859 & 0.401 & 1288 & 899 & 212 & 601 \\
UKB & 3000 & EBM & 1:1 & 0.646 & 0.646 & 0.648 & 0.643 & 972 & 535 & 528 & 965 \\
UKB & 3000 & EBM & 1:2 & 0.622 & 0.495 & 0.371 & 0.872 & 557 & 192 & 943 & 1308 \\
UKB & 3000 & EBM & 1:5 & 0.554 & 0.231 & 0.134 & 0.975 & 201 & 38 & 1299 & 1462 \\
UKB & 3000 & EBM & 1:10 & 0.514 & 0.057 & 0.029 & 0.999 & 44 & 1 & 1456 & 1499 \\
UKB & 3000 & EBM & 1:50 & 0.500 & 0.001 & 0.001 & 1.000 & 1 & 0 & 1499 & 1500 \\
UKB & 3000 & APLR & 50:1 & 0.500 & 0.667 & 1.000 & 0.000 & 1500 & 1500 & 0 & 0 \\
UKB & 3000 & APLR & 10:1 & 0.500 & 0.667 & 1.000 & 0.000 & 1500 & 1500 & 0 & 0 \\
UKB & 3000 & APLR & 5:1 & 0.501 & 0.668 & 1.000 & 0.003 & 1500 & 1496 & 0 & 4 \\
UKB & 3000 & APLR & 2:1 & 0.584 & 0.692 & 0.933 & 0.235 & 1399 & 1147 & 101 & 353 \\
UKB & 3000 & APLR & 1:1 & 0.647 & 0.627 & 0.596 & 0.698 & 894 & 453 & 606 & 1047 \\
UKB & 3000 & APLR & 1:2 & 0.582 & 0.375 & 0.251 & 0.914 & 376 & 129 & 1124 & 1371 \\
UKB & 3000 & APLR & 1:5 & 0.521 & 0.099 & 0.053 & 0.990 & 79 & 15 & 1421 & 1485 \\
UKB & 3000 & APLR & 1:10 & 0.501 & 0.008 & 0.004 & 0.998 & 6 & 3 & 1494 & 1497 \\
UKB & 3000 & APLR & 1:50 & 0.500 & 0.000 & 0.000 & 1.000 & 0 & 0 & 1500 & 1500 \\
UKB & 3000 & CORELS & 50:1 & 0.500 & 0.000 & 0.000 & 1.000 & 0 & 0 & 1500 & 1500 \\
UKB & 3000 & CORELS & 10:1 & 0.500 & 0.000 & 0.000 & 1.000 & 0 & 0 & 1500 & 1500 \\
UKB & 3000 & CORELS & 5:1 & 0.500 & 0.000 & 0.000 & 1.000 & 0 & 0 & 1500 & 1500 \\
UKB & 3000 & CORELS & 2:1 & 0.535 & 0.452 & 0.628 & 0.441 & 942 & 838 & 558 & 662 \\
UKB & 3000 & CORELS & 1:1 & 0.602 & 0.590 & 0.573 & 0.632 & 859 & 552 & 641 & 948 \\
UKB & 3000 & CORELS & 1:2 & 0.564 & 0.300 & 0.187 & 0.940 & 281 & 90 & 1219 & 1410 \\
UKB & 3000 & CORELS & 1:5 & 0.500 & 0.000 & 0.000 & 1.000 & 0 & 0 & 1500 & 1500 \\
UKB & 3000 & CORELS & 1:10 & 0.500 & 0.000 & 0.000 & 1.000 & 0 & 0 & 1500 & 1500 \\
UKB & 3000 & CORELS & 1:50 & 0.500 & 0.000 & 0.000 & 1.000 & 0 & 0 & 1500 & 1500 \\
UKB & 3000 & MHL & 50:1 & 0.511 & 0.653 & 0.921 & 0.102 & 1381 & 1347 & 119 & 153 \\
UKB & 3000 & MHL & 10:1 & 0.521 & 0.663 & 0.943 & 0.099 & 1415 & 1352 & 85 & 148 \\
UKB & 3000 & MHL & 5:1 & 0.545 & 0.628 & 0.795 & 0.295 & 1192 & 1058 & 308 & 442 \\
UKB & 3000 & MHL & 2:1 & 0.534 & 0.582 & 0.671 & 0.398 & 1006 & 903 & 494 & 597 \\
UKB & 3000 & MHL & 1:1 & 0.561 & 0.621 & 0.723 & 0.399 & 1085 & 902 & 415 & 598 \\
UKB & 3000 & MHL & 1:2 & 0.587 & 0.613 & 0.667 & 0.507 & 1001 & 739 & 499 & 761 \\
UKB & 3000 & MHL & 1:5 & 0.498 & 0.640 & 0.903 & 0.092 & 1355 & 1362 & 145 & 138 \\
UKB & 3000 & MHL & 1:10 & 0.559 & 0.642 & 0.790 & 0.327 & 1185 & 1009 & 315 & 491 \\
UKB & 3000 & MHL & 1:50 & 0.504 & 0.664 & 0.979 & 0.029 & 1469 & 1457 & 31 & 43 \\
CCID & 1000 & Logistic Regression & 50:1 & 0.508 & 0.670 & 0.999 & 0.017 & 1498 & 1474 & 2 & 26 \\
CCID & 1000 & Logistic Regression & 10:1 & 0.574 & 0.695 & 0.973 & 0.175 & 1459 & 1238 & 41 & 262 \\
CCID & 1000 & Logistic Regression & 5:1 & 0.617 & 0.714 & 0.954 & 0.280 & 1431 & 1080 & 69 & 420 \\
CCID & 1000 & Logistic Regression & 2:1 & 0.723 & 0.751 & 0.833 & 0.613 & 1250 & 580 & 250 & 920 \\
CCID & 1000 & Logistic Regression & 1:1 & 0.757 & 0.743 & 0.703 & 0.811 & 1055 & 284 & 445 & 1216 \\
CCID & 1000 & Logistic Regression & 1:2 & 0.734 & 0.678 & 0.561 & 0.906 & 842 & 141 & 658 & 1359 \\
CCID & 1000 & Logistic Regression & 1:5 & 0.642 & 0.476 & 0.327 & 0.958 & 490 & 63 & 1010 & 1437 \\
CCID & 1000 & Logistic Regression & 1:10 & 0.605 & 0.366 & 0.229 & 0.980 & 344 & 30 & 1156 & 1470 \\
CCID & 1000 & Logistic Regression & 1:50 & 0.521 & 0.096 & 0.051 & 0.991 & 76 & 13 & 1424 & 1487 \\
CCID & 1000 & Decision Tree & 50:1 & 0.535 & 0.674 & 0.961 & 0.108 & 1442 & 1338 & 58 & 162 \\
CCID & 1000 & Decision Tree & 10:1 & 0.605 & 0.689 & 0.875 & 0.335 & 1312 & 997 & 188 & 503 \\
CCID & 1000 & Decision Tree & 5:1 & 0.624 & 0.675 & 0.781 & 0.467 & 1171 & 799 & 329 & 701 \\
CCID & 1000 & Decision Tree & 2:1 & 0.679 & 0.691 & 0.721 & 0.637 & 1081 & 544 & 419 & 956 \\
CCID & 1000 & Decision Tree & 1:1 & 0.688 & 0.686 & 0.681 & 0.695 & 1022 & 457 & 478 & 1043 \\
CCID & 1000 & Decision Tree & 1:2 & 0.671 & 0.629 & 0.559 & 0.783 & 838 & 326 & 662 & 1174 \\
CCID & 1000 & Decision Tree & 1:5 & 0.620 & 0.493 & 0.370 & 0.870 & 555 & 195 & 945 & 1305 \\
CCID & 1000 & Decision Tree & 1:10 & 0.620 & 0.454 & 0.317 & 0.924 & 475 & 114 & 1025 & 1386 \\
CCID & 1000 & Decision Tree & 1:50 & 0.537 & 0.173 & 0.098 & 0.975 & 147 & 37 & 1353 & 1463 \\
CCID & 1000 & MLP & 50:1 & 0.500 & 0.667 & 1.000 & 0.000 & 1500 & 1500 & 0 & 0 \\
CCID & 1000 & MLP & 10:1 & 0.500 & 0.667 & 1.000 & 0.000 & 1500 & 1500 & 0 & 0 \\
CCID & 1000 & MLP & 5:1 & 0.654 & 0.727 & 0.921 & 0.387 & 1381 & 920 & 119 & 580 \\
CCID & 1000 & MLP & 2:1 & 0.730 & 0.748 & 0.803 & 0.657 & 1205 & 515 & 295 & 985 \\
CCID & 1000 & MLP & 1:1 & 0.763 & 0.751 & 0.715 & 0.812 & 1072 & 282 & 428 & 1218 \\
CCID & 1000 & MLP & 1:2 & 0.725 & 0.669 & 0.555 & 0.896 & 832 & 156 & 668 & 1344 \\
CCID & 1000 & MLP & 1:5 & 0.571 & 0.230 & 0.157 & 0.986 & 235 & 21 & 1265 & 1479 \\
CCID & 1000 & MLP & 1:10 & 0.562 & 0.225 & 0.138 & 0.987 & 207 & 20 & 1293 & 1480 \\
CCID & 1000 & MLP & 1:50 & 0.500 & 0.000 & 0.000 & 1.000 & 0 & 0 & 1500 & 1500 \\
CCID & 1000 & XGBoost & 50:1 & 0.503 & 0.668 & 0.999 & 0.008 & 1498 & 1488 & 2 & 12 \\
CCID & 1000 & XGBoost & 10:1 & 0.613 & 0.714 & 0.966 & 0.261 & 1449 & 1109 & 51 & 391 \\
CCID & 1000 & XGBoost & 5:1 & 0.690 & 0.749 & 0.923 & 0.457 & 1385 & 815 & 115 & 685 \\
CCID & 1000 & XGBoost & 2:1 & 0.751 & 0.768 & 0.824 & 0.678 & 1236 & 483 & 264 & 1017 \\
CCID & 1000 & XGBoost & 1:1 & 0.761 & 0.760 & 0.759 & 0.763 & 1138 & 355 & 362 & 1145 \\
CCID & 1000 & XGBoost & 1:2 & 0.746 & 0.703 & 0.603 & 0.889 & 904 & 167 & 596 & 1333 \\
CCID & 1000 & XGBoost & 1:5 & 0.646 & 0.480 & 0.327 & 0.966 & 490 & 51 & 1010 & 1449 \\
CCID & 1000 & XGBoost & 1:10 & 0.582 & 0.303 & 0.182 & 0.982 & 273 & 27 & 1227 & 1473 \\
CCID & 1000 & XGBoost & 1:50 & 0.508 & 0.030 & 0.015 & 1.000 & 23 & 0 & 1477 & 1500 \\
CCID & 1000 & LightGBM & 50:1 & 0.505 & 0.669 & 1.000 & 0.009 & 1500 & 1486 & 0 & 14 \\
CCID & 1000 & LightGBM & 10:1 & 0.622 & 0.721 & 0.975 & 0.269 & 1462 & 1096 & 38 & 404 \\
CCID & 1000 & LightGBM & 5:1 & 0.689 & 0.747 & 0.920 & 0.457 & 1380 & 814 & 120 & 686 \\
CCID & 1000 & LightGBM & 2:1 & 0.741 & 0.757 & 0.809 & 0.674 & 1213 & 489 & 287 & 1011 \\
CCID & 1000 & LightGBM & 1:1 & 0.761 & 0.760 & 0.758 & 0.763 & 1137 & 355 & 363 & 1145 \\
CCID & 1000 & LightGBM & 1:2 & 0.745 & 0.701 & 0.597 & 0.894 & 895 & 159 & 605 & 1341 \\
CCID & 1000 & LightGBM & 1:5 & 0.642 & 0.469 & 0.316 & 0.969 & 474 & 47 & 1026 & 1453 \\
CCID & 1000 & LightGBM & 1:10 & 0.578 & 0.295 & 0.176 & 0.981 & 264 & 29 & 1236 & 1471 \\
CCID & 1000 & LightGBM & 1:50 & 0.503 & 0.011 & 0.005 & 1.000 & 8 & 0 & 1492 & 1500 \\
CCID & 1000 & FT-Transformer & 50:1 & 0.544 & 0.685 & 0.988 & 0.099 & 1482 & 1351 & 18 & 149 \\
CCID & 1000 & FT-Transformer & 10:1 & 0.631 & 0.718 & 0.928 & 0.335 & 1392 & 998 & 108 & 502 \\
CCID & 1000 & FT-Transformer & 5:1 & 0.712 & 0.750 & 0.864 & 0.559 & 1296 & 661 & 204 & 839 \\
CCID & 1000 & FT-Transformer & 2:1 & 0.736 & 0.747 & 0.782 & 0.689 & 1173 & 466 & 327 & 1034 \\
CCID & 1000 & FT-Transformer & 1:1 & 0.748 & 0.746 & 0.742 & 0.753 & 1113 & 370 & 387 & 1130 \\
CCID & 1000 & FT-Transformer & 1:2 & 0.750 & 0.726 & 0.662 & 0.837 & 993 & 244 & 507 & 1256 \\
CCID & 1000 & FT-Transformer & 1:5 & 0.693 & 0.588 & 0.443 & 0.943 & 664 & 86 & 836 & 1414 \\
CCID & 1000 & FT-Transformer & 1:10 & 0.601 & 0.330 & 0.227 & 0.975 & 340 & 37 & 1160 & 1463 \\
CCID & 1000 & FT-Transformer & 1:50 & 0.521 & 0.088 & 0.047 & 0.995 & 71 & 7 & 1429 & 1493 \\
CCID & 1000 & ResNet & 50:1 & 0.502 & 0.667 & 0.998 & 0.006 & 1497 & 1491 & 3 & 9 \\
CCID & 1000 & ResNet & 10:1 & 0.661 & 0.732 & 0.923 & 0.399 & 1385 & 901 & 115 & 599 \\
CCID & 1000 & ResNet & 5:1 & 0.702 & 0.752 & 0.901 & 0.504 & 1351 & 744 & 149 & 756 \\
CCID & 1000 & ResNet & 2:1 & 0.738 & 0.749 & 0.784 & 0.691 & 1176 & 463 & 324 & 1037 \\
CCID & 1000 & ResNet & 1:1 & 0.748 & 0.751 & 0.761 & 0.736 & 1141 & 396 & 359 & 1104 \\
CCID & 1000 & ResNet & 1:2 & 0.746 & 0.748 & 0.754 & 0.738 & 1131 & 393 & 369 & 1107 \\
CCID & 1000 & ResNet & 1:5 & 0.689 & 0.590 & 0.451 & 0.927 & 677 & 109 & 823 & 1391 \\
CCID & 1000 & ResNet & 1:10 & 0.630 & 0.448 & 0.311 & 0.950 & 466 & 75 & 1034 & 1425 \\
CCID & 1000 & ResNet & 1:50 & 0.500 & 0.004 & 0.002 & 0.998 & 3 & 3 & 1497 & 1497 \\
CCID & 1000 & EBM & 50:1 & 0.505 & 0.669 & 1.000 & 0.009 & 1500 & 1486 & 0 & 14 \\
CCID & 1000 & EBM & 10:1 & 0.611 & 0.715 & 0.977 & 0.245 & 1466 & 1132 & 34 & 368 \\
CCID & 1000 & EBM & 5:1 & 0.688 & 0.747 & 0.921 & 0.454 & 1382 & 819 & 118 & 681 \\
CCID & 1000 & EBM & 2:1 & 0.735 & 0.752 & 0.803 & 0.667 & 1204 & 500 & 296 & 1000 \\
CCID & 1000 & EBM & 1:1 & 0.756 & 0.753 & 0.743 & 0.769 & 1115 & 347 & 385 & 1153 \\
CCID & 1000 & EBM & 1:2 & 0.754 & 0.716 & 0.621 & 0.888 & 931 & 168 & 569 & 1332 \\
CCID & 1000 & EBM & 1:5 & 0.656 & 0.499 & 0.346 & 0.965 & 519 & 52 & 981 & 1448 \\
CCID & 1000 & EBM & 1:10 & 0.585 & 0.310 & 0.187 & 0.983 & 281 & 25 & 1219 & 1475 \\
CCID & 1000 & EBM & 1:50 & 0.507 & 0.026 & 0.013 & 1.000 & 20 & 0 & 1480 & 1500 \\
CCID & 1000 & APLR & 50:1 & 0.499 & 0.666 & 0.998 & 0.000 & 1497 & 1500 & 3 & 0 \\
CCID & 1000 & APLR & 10:1 & 0.542 & 0.683 & 0.985 & 0.100 & 1477 & 1350 & 23 & 150 \\
CCID & 1000 & APLR & 5:1 & 0.602 & 0.711 & 0.977 & 0.226 & 1466 & 1161 & 34 & 339 \\
CCID & 1000 & APLR & 2:1 & 0.720 & 0.759 & 0.883 & 0.557 & 1324 & 665 & 176 & 835 \\
CCID & 1000 & APLR & 1:1 & 0.748 & 0.728 & 0.679 & 0.817 & 1018 & 275 & 482 & 1225 \\
CCID & 1000 & APLR & 1:2 & 0.726 & 0.661 & 0.535 & 0.917 & 803 & 125 & 697 & 1375 \\
CCID & 1000 & APLR & 1:5 & 0.611 & 0.379 & 0.245 & 0.977 & 367 & 35 & 1133 & 1465 \\
CCID & 1000 & APLR & 1:10 & 0.513 & 0.050 & 0.027 & 0.999 & 40 & 1 & 1460 & 1499 \\
CCID & 1000 & APLR & 1:50 & 0.500 & 0.000 & 0.000 & 0.999 & 0 & 1 & 1500 & 1499 \\
CCID & 1000 & CORELS & 50:1 & 0.500 & 0.000 & 0.000 & 1.000 & 0 & 0 & 1500 & 1500 \\
CCID & 1000 & CORELS & 10:1 & 0.500 & 0.000 & 0.000 & 1.000 & 0 & 0 & 1500 & 1500 \\
CCID & 1000 & CORELS & 5:1 & 0.537 & 0.679 & 0.978 & 0.096 & 1467 & 1356 & 33 & 144 \\
CCID & 1000 & CORELS & 2:1 & 0.625 & 0.705 & 0.894 & 0.356 & 1341 & 966 & 159 & 534 \\
CCID & 1000 & CORELS & 1:1 & 0.709 & 0.671 & 0.595 & 0.824 & 892 & 264 & 608 & 1236 \\
CCID & 1000 & CORELS & 1:2 & 0.717 & 0.647 & 0.518 & 0.917 & 777 & 125 & 723 & 1375 \\
CCID & 1000 & CORELS & 1:5 & 0.649 & 0.482 & 0.335 & 0.963 & 503 & 56 & 997 & 1444 \\
CCID & 1000 & CORELS & 1:10 & 0.500 & 0.000 & 0.000 & 1.000 & 0 & 0 & 1500 & 1500 \\
CCID & 1000 & CORELS & 1:50 & 0.500 & 0.000 & 0.000 & 1.000 & 0 & 0 & 1500 & 1500 \\
CCID & 1000 & MHL & 50:1 & 0.609 & 0.667 & 0.787 & 0.432 & 1180 & 852 & 320 & 648 \\
CCID & 1000 & MHL & 10:1 & 0.618 & 0.670 & 0.779 & 0.456 & 1169 & 816 & 331 & 684 \\
CCID & 1000 & MHL & 5:1 & 0.610 & 0.686 & 0.848 & 0.373 & 1272 & 941 & 228 & 559 \\
CCID & 1000 & MHL & 2:1 & 0.671 & 0.706 & 0.792 & 0.551 & 1188 & 674 & 312 & 826 \\
CCID & 1000 & MHL & 1:1 & 0.708 & 0.728 & 0.783 & 0.632 & 1125 & 499 & 375 & 1001 \\
CCID & 1000 & MHL & 1:2 & 0.664 & 0.714 & 0.833 & 0.496 & 1249 & 756 & 251 & 744 \\
CCID & 1000 & MHL & 1:5 & 0.669 & 0.723 & 0.861 & 0.477 & 1291 & 784 & 209 & 716 \\
CCID & 1000 & MHL & 1:10 & 0.690 & 0.719 & 0.798 & 0.582 & 1197 & 627 & 303 & 873 \\
CCID & 1000 & MHL & 1:50 & 0.675 & 0.700 & 0.760 & 0.589 & 1140 & 616 & 360 & 884 \\
CCID & 3000 & Logistic Regression & 50:1 & 0.505 & 0.668 & 0.996 & 0.013 & 1494 & 1480 & 6 & 20 \\
CCID & 3000 & Logistic Regression & 10:1 & 0.569 & 0.694 & 0.976 & 0.162 & 1464 & 1257 & 36 & 243 \\
CCID & 3000 & Logistic Regression & 5:1 & 0.629 & 0.721 & 0.957 & 0.301 & 1436 & 1049 & 64 & 451 \\
CCID & 3000 & Logistic Regression & 2:1 & 0.735 & 0.761 & 0.846 & 0.624 & 1269 & 564 & 231 & 936 \\
CCID & 3000 & Logistic Regression & 1:1 & 0.762 & 0.746 & 0.700 & 0.823 & 1050 & 265 & 450 & 1235 \\
CCID & 3000 & Logistic Regression & 1:2 & 0.728 & 0.666 & 0.543 & 0.914 & 814 & 129 & 686 & 1371 \\
CCID & 3000 & Logistic Regression & 1:5 & 0.669 & 0.532 & 0.377 & 0.960 & 566 & 60 & 934 & 1440 \\
CCID & 3000 & Logistic Regression & 1:10 & 0.615 & 0.392 & 0.249 & 0.981 & 373 & 28 & 1127 & 1472 \\
CCID & 3000 & Logistic Regression & 1:50 & 0.526 & 0.101 & 0.053 & 0.999 & 80 & 2 & 1420 & 1498 \\
CCID & 3000 & Decision Tree & 50:1 & 0.566 & 0.677 & 0.909 & 0.223 & 1363 & 1166 & 137 & 334 \\
CCID & 3000 & Decision Tree & 10:1 & 0.649 & 0.683 & 0.759 & 0.539 & 1138 & 691 & 362 & 809 \\
CCID & 3000 & Decision Tree & 5:1 & 0.651 & 0.657 & 0.667 & 0.635 & 1001 & 548 & 499 & 952 \\
CCID & 3000 & Decision Tree & 2:1 & 0.657 & 0.627 & 0.577 & 0.736 & 866 & 396 & 634 & 1104 \\
CCID & 3000 & Decision Tree & 1:1 & 0.661 & 0.602 & 0.514 & 0.808 & 771 & 288 & 729 & 1212 \\
CCID & 3000 & Decision Tree & 1:2 & 0.659 & 0.576 & 0.463 & 0.855 & 695 & 217 & 805 & 1283 \\
CCID & 3000 & Decision Tree & 1:5 & 0.657 & 0.597 & 0.507 & 0.807 & 761 & 289 & 739 & 1211 \\
CCID & 3000 & Decision Tree & 1:10 & 0.645 & 0.525 & 0.393 & 0.898 & 589 & 153 & 911 & 1347 \\
CCID & 3000 & Decision Tree & 1:50 & 0.565 & 0.273 & 0.164 & 0.965 & 246 & 52 & 1254 & 1448 \\
CCID & 3000 & MLP & 50:1 & 0.500 & 0.667 & 1.000 & 0.000 & 1500 & 1500 & 0 & 0 \\
CCID & 3000 & MLP & 10:1 & 0.677 & 0.736 & 0.900 & 0.453 & 1350 & 820 & 150 & 680 \\
CCID & 3000 & MLP & 5:1 & 0.721 & 0.742 & 0.805 & 0.637 & 1207 & 545 & 293 & 955 \\
CCID & 3000 & MLP & 2:1 & 0.738 & 0.724 & 0.690 & 0.785 & 1035 & 322 & 465 & 1178 \\
CCID & 3000 & MLP & 1:1 & 0.725 & 0.678 & 0.581 & 0.869 & 872 & 197 & 628 & 1303 \\
CCID & 3000 & MLP & 1:2 & 0.705 & 0.624 & 0.493 & 0.917 & 740 & 125 & 760 & 1375 \\
CCID & 3000 & MLP & 1:5 & 0.700 & 0.602 & 0.457 & 0.943 & 685 & 85 & 815 & 1415 \\
CCID & 3000 & MLP & 1:10 & 0.600 & 0.329 & 0.229 & 0.971 & 344 & 43 & 1156 & 1457 \\
CCID & 3000 & MLP & 1:50 & 0.504 & 0.016 & 0.008 & 0.999 & 12 & 1 & 1488 & 1499 \\
CCID & 3000 & XGBoost & 50:1 & 0.541 & 0.685 & 0.997 & 0.085 & 1496 & 1373 & 4 & 127 \\
CCID & 3000 & XGBoost & 10:1 & 0.715 & 0.766 & 0.933 & 0.497 & 1399 & 754 & 101 & 746 \\
CCID & 3000 & XGBoost & 5:1 & 0.744 & 0.768 & 0.849 & 0.638 & 1274 & 543 & 226 & 957 \\
CCID & 3000 & XGBoost & 2:1 & 0.770 & 0.760 & 0.726 & 0.815 & 1089 & 278 & 411 & 1222 \\
CCID & 3000 & XGBoost & 1:1 & 0.754 & 0.715 & 0.618 & 0.891 & 927 & 164 & 573 & 1336 \\
CCID & 3000 & XGBoost & 1:2 & 0.718 & 0.635 & 0.492 & 0.944 & 738 & 84 & 762 & 1416 \\
CCID & 3000 & XGBoost & 1:5 & 0.718 & 0.637 & 0.494 & 0.942 & 741 & 87 & 759 & 1413 \\
CCID & 3000 & XGBoost & 1:10 & 0.632 & 0.440 & 0.290 & 0.974 & 435 & 39 & 1065 & 1461 \\
CCID & 3000 & XGBoost & 1:50 & 0.516 & 0.068 & 0.035 & 0.997 & 53 & 4 & 1447 & 1496 \\
CCID & 3000 & LightGBM & 50:1 & 0.544 & 0.686 & 0.996 & 0.092 & 1494 & 1362 & 6 & 138 \\
CCID & 3000 & LightGBM & 10:1 & 0.717 & 0.767 & 0.932 & 0.502 & 1398 & 747 & 102 & 753 \\
CCID & 3000 & LightGBM & 5:1 & 0.755 & 0.775 & 0.842 & 0.668 & 1263 & 498 & 237 & 1002 \\
CCID & 3000 & LightGBM & 2:1 & 0.766 & 0.750 & 0.703 & 0.829 & 1055 & 257 & 445 & 1243 \\
CCID & 3000 & LightGBM & 1:1 & 0.750 & 0.705 & 0.599 & 0.901 & 898 & 148 & 602 & 1352 \\
CCID & 3000 & LightGBM & 1:2 & 0.717 & 0.630 & 0.483 & 0.951 & 724 & 74 & 776 & 1426 \\
CCID & 3000 & LightGBM & 1:5 & 0.717 & 0.636 & 0.495 & 0.939 & 742 & 91 & 758 & 1409 \\
CCID & 3000 & LightGBM & 1:10 & 0.637 & 0.451 & 0.299 & 0.975 & 448 & 37 & 1052 & 1463 \\
CCID & 3000 & LightGBM & 1:50 & 0.513 & 0.052 & 0.027 & 1.000 & 40 & 0 & 1460 & 1500 \\
CCID & 3000 & FT-Transformer & 50:1 & 0.570 & 0.694 & 0.975 & 0.164 & 1463 & 1254 & 37 & 246 \\
CCID & 3000 & FT-Transformer & 10:1 & 0.653 & 0.731 & 0.939 & 0.367 & 1409 & 949 & 91 & 551 \\
CCID & 3000 & FT-Transformer & 5:1 & 0.717 & 0.747 & 0.833 & 0.601 & 1250 & 598 & 250 & 902 \\
CCID & 3000 & FT-Transformer & 2:1 & 0.745 & 0.745 & 0.747 & 0.743 & 1120 & 386 & 380 & 1114 \\
CCID & 3000 & FT-Transformer & 1:1 & 0.751 & 0.743 & 0.719 & 0.783 & 1079 & 326 & 421 & 1174 \\
CCID & 3000 & FT-Transformer & 1:2 & 0.731 & 0.685 & 0.594 & 0.868 & 891 & 198 & 609 & 1302 \\
CCID & 3000 & FT-Transformer & 1:5 & 0.707 & 0.621 & 0.481 & 0.933 & 722 & 100 & 778 & 1400 \\
CCID & 3000 & FT-Transformer & 1:10 & 0.673 & 0.547 & 0.397 & 0.949 & 596 & 76 & 904 & 1424 \\
CCID & 3000 & FT-Transformer & 1:50 & 0.546 & 0.175 & 0.099 & 0.992 & 149 & 12 & 1351 & 1488 \\
CCID & 3000 & ResNet & 50:1 & 0.506 & 0.669 & 0.997 & 0.015 & 1495 & 1478 & 5 & 22 \\
CCID & 3000 & ResNet & 10:1 & 0.595 & 0.705 & 0.961 & 0.228 & 1442 & 1158 & 58 & 342 \\
CCID & 3000 & ResNet & 5:1 & 0.675 & 0.736 & 0.902 & 0.449 & 1353 & 827 & 147 & 673 \\
CCID & 3000 & ResNet & 2:1 & 0.736 & 0.755 & 0.813 & 0.659 & 1219 & 512 & 281 & 988 \\
CCID & 3000 & ResNet & 1:1 & 0.759 & 0.756 & 0.746 & 0.773 & 1119 & 341 & 381 & 1159 \\
CCID & 3000 & ResNet & 1:2 & 0.742 & 0.729 & 0.700 & 0.785 & 1050 & 323 & 450 & 1177 \\
CCID & 3000 & ResNet & 1:5 & 0.716 & 0.650 & 0.534 & 0.897 & 801 & 154 & 699 & 1346 \\
CCID & 3000 & ResNet & 1:10 & 0.614 & 0.373 & 0.257 & 0.971 & 385 & 43 & 1115 & 1457 \\
CCID & 3000 & ResNet & 1:50 & 0.506 & 0.028 & 0.015 & 0.997 & 22 & 4 & 1478 & 1496 \\
CCID & 3000 & EBM & 50:1 & 0.537 & 0.683 & 0.998 & 0.077 & 1497 & 1385 & 3 & 115 \\
CCID & 3000 & EBM & 10:1 & 0.703 & 0.751 & 0.894 & 0.512 & 1341 & 732 & 159 & 768 \\
CCID & 3000 & EBM & 5:1 & 0.735 & 0.757 & 0.825 & 0.645 & 1237 & 532 & 263 & 968 \\
CCID & 3000 & EBM & 2:1 & 0.748 & 0.739 & 0.712 & 0.784 & 1068 & 324 & 432 & 1176 \\
CCID & 3000 & EBM & 1:1 & 0.740 & 0.705 & 0.621 & 0.860 & 931 & 210 & 569 & 1290 \\
CCID & 3000 & EBM & 1:2 & 0.717 & 0.653 & 0.532 & 0.903 & 798 & 146 & 702 & 1354 \\
CCID & 3000 & EBM & 1:5 & 0.719 & 0.638 & 0.498 & 0.939 & 747 & 91 & 753 & 1409 \\
CCID & 3000 & EBM & 1:10 & 0.650 & 0.487 & 0.333 & 0.967 & 499 & 49 & 1001 & 1451 \\
CCID & 3000 & EBM & 1:50 & 0.523 & 0.086 & 0.045 & 1.000 & 68 & 0 & 1432 & 1500 \\
CCID & 3000 & APLR & 50:1 & 0.505 & 0.667 & 0.994 & 0.015 & 1491 & 1477 & 9 & 23 \\
CCID & 3000 & APLR & 10:1 & 0.602 & 0.710 & 0.973 & 0.231 & 1459 & 1153 & 41 & 347 \\
CCID & 3000 & APLR & 5:1 & 0.644 & 0.728 & 0.955 & 0.333 & 1433 & 1001 & 67 & 499 \\
CCID & 3000 & APLR & 2:1 & 0.738 & 0.764 & 0.846 & 0.631 & 1269 & 554 & 231 & 946 \\
CCID & 3000 & APLR & 1:1 & 0.763 & 0.751 & 0.717 & 0.809 & 1076 & 287 & 424 & 1213 \\
CCID & 3000 & APLR & 1:2 & 0.730 & 0.669 & 0.549 & 0.911 & 823 & 134 & 677 & 1366 \\
CCID & 3000 & APLR & 1:5 & 0.662 & 0.521 & 0.369 & 0.956 & 553 & 66 & 947 & 1434 \\
CCID & 3000 & APLR & 1:10 & 0.565 & 0.239 & 0.139 & 0.991 & 208 & 14 & 1292 & 1486 \\
CCID & 3000 & APLR & 1:50 & 0.500 & 0.000 & 0.000 & 1.000 & 0 & 0 & 1500 & 1500 \\
CCID & 3000 & CORELS & 50:1 & 0.500 & 0.000 & 0.000 & 1.000 & 0 & 0 & 1500 & 1500 \\
CCID & 3000 & CORELS & 10:1 & 0.500 & 0.000 & 0.000 & 1.000 & 0 & 0 & 1500 & 1500 \\
CCID & 3000 & CORELS & 5:1 & 0.500 & 0.000 & 0.000 & 1.000 & 0 & 0 & 1500 & 1500 \\
CCID & 3000 & CORELS & 2:1 & 0.637 & 0.711 & 0.890 & 0.385 & 1335 & 923 & 165 & 577 \\
CCID & 3000 & CORELS & 1:1 & 0.710 & 0.663 & 0.577 & 0.843 & 865 & 235 & 635 & 1265 \\
CCID & 3000 & CORELS & 1:2 & 0.714 & 0.637 & 0.503 & 0.925 & 755 & 113 & 745 & 1387 \\
CCID & 3000 & CORELS & 1:5 & 0.687 & 0.574 & 0.423 & 0.951 & 634 & 73 & 866 & 1427 \\
CCID & 3000 & CORELS & 1:10 & 0.500 & 0.000 & 0.000 & 1.000 & 0 & 0 & 1500 & 1500 \\
CCID & 3000 & CORELS & 1:50 & 0.500 & 0.000 & 0.000 & 1.000 & 0 & 0 & 1500 & 1500 \\
CCID & 3000 & MHL & 50:1 & 0.671 & 0.705 & 0.788 & 0.553 & 1182 & 670 & 318 & 830 \\
CCID & 3000 & MHL & 10:1 & 0.634 & 0.695 & 0.834 & 0.435 & 1251 & 848 & 249 & 652 \\
CCID & 3000 & MHL & 5:1 & 0.646 & 0.671 & 0.727 & 0.565 & 1090 & 652 & 410 & 848 \\
CCID & 3000 & MHL & 2:1 & 0.653 & 0.697 & 0.797 & 0.509 & 1195 & 736 & 305 & 764 \\
CCID & 3000 & MHL & 1:1 & 0.705 & 0.727 & 0.785 & 0.625 & 1177 & 562 & 323 & 938 \\
CCID & 3000 & MHL & 1:2 & 0.639 & 0.703 & 0.853 & 0.426 & 1279 & 861 & 221 & 639 \\
CCID & 3000 & MHL & 1:5 & 0.684 & 0.709 & 0.765 & 0.602 & 1148 & 597 & 352 & 903 \\
CCID & 3000 & MHL & 1:10 & 0.610 & 0.688 & 0.860 & 0.360 & 1290 & 960 & 210 & 540 \\
CCID & 3000 & MHL & 1:50 & 0.710 & 0.677 & 0.609 & 0.811 & 914 & 283 & 586 & 1217 \\
\end{longtable}
\endgroup

% Auto-generated from hl-data-process/contrast0/contrast0_summary.csv.
% Regenerate with hl-data-process/generate_results_latex_tables.py; do not edit numeric values manually.
% The internal CSV dataset identifier YHD is rendered as CCID to match the manuscript terminology.
\begin{table}[!t]
\centering
\small
\caption{Mean test performance of the final MHL rule programs across random seeds 36, 40, and 42 for seven LLM backend configurations. The LLMs serve as rule-generation and rule-revision backends rather than direct classifiers.}
\label{tab:results-2-6-backends}
\begin{tabular}{@{}llrrrr@{}}
\toprule
Backend & Dataset & ACC & F1 & Sens. & Spec. \\
\midrule
DeepSeek-V4-Pro-High & UKB & 0.549 & 0.637 & 0.799 & 0.300 \\
DeepSeek-V4-Pro-Max & UKB & 0.561 & 0.663 & 0.860 & 0.263 \\
DeepSeek-V4-Flash-High & UKB & 0.548 & 0.652 & 0.851 & 0.245 \\
DeepSeek-V4-Flash-Max & UKB & 0.566 & 0.636 & 0.775 & 0.357 \\
Qwen 3.7-Max & UKB & 0.590 & 0.639 & 0.730 & 0.451 \\
Gemini 3.1-Pro & UKB & 0.606 & 0.645 & 0.715 & 0.497 \\
GPT-5.5 & UKB & 0.545 & 0.664 & 0.899 & 0.192 \\
DeepSeek-V4-Pro-High & CCID & 0.688 & 0.721 & 0.807 & 0.569 \\
DeepSeek-V4-Pro-Max & CCID & 0.675 & 0.719 & 0.828 & 0.523 \\
DeepSeek-V4-Flash-High & CCID & 0.661 & 0.707 & 0.817 & 0.506 \\
DeepSeek-V4-Flash-Max & CCID & 0.620 & 0.703 & 0.896 & 0.343 \\
Qwen 3.7-Max & CCID & 0.703 & 0.728 & 0.797 & 0.608 \\
Gemini 3.1-Pro & CCID & 0.707 & 0.731 & 0.795 & 0.619 \\
GPT-5.5 & CCID & 0.665 & 0.721 & 0.867 & 0.463 \\
\bottomrule
\end{tabular}
\end{table}

% Auto-generated from hl-data-process/continuous_learning/continuous_learning_summary.csv.
% Regenerate with hl-data-process/generate_results_latex_tables.py; do not edit numeric values manually.
\begin{table}[!t]
\centering
\small
\caption{Mean MIMIC test performance across random seeds 36, 40, and 42 at the Stage 1 direct-training, Stage 2 continual-learning, and Stage 2 direct-training endpoints. Only the two Stage 2 endpoints share the same test split; Stage 1 is contextual.}
\label{tab:results-2-7-continual-learning}
\begin{tabular}{@{}llrrrr@{}}
\toprule
Model & Endpoint & ACC & F1 & Sens. & Spec. \\
\midrule
MLP & Stage 1 direct ($n=1{,}000$) & 0.656 & 0.655 & 0.655 & 0.657 \\
MLP & Stage 2 continual ($n=40$) & 0.655 & 0.651 & 0.645 & 0.664 \\
MLP & Stage 2 direct ($n=40$) & 0.576 & 0.559 & 0.548 & 0.603 \\
XGBoost & Stage 1 direct ($n=1{,}000$) & 0.669 & 0.669 & 0.671 & 0.666 \\
XGBoost & Stage 2 continual ($n=40$) & 0.653 & 0.655 & 0.661 & 0.645 \\
XGBoost & Stage 2 direct ($n=40$) & 0.598 & 0.576 & 0.548 & 0.648 \\
LightGBM & Stage 1 direct ($n=1{,}000$) & 0.658 & 0.659 & 0.659 & 0.656 \\
LightGBM & Stage 2 continual ($n=40$) & 0.648 & 0.652 & 0.661 & 0.635 \\
LightGBM & Stage 2 direct ($n=40$) & 0.570 & 0.578 & 0.594 & 0.545 \\
FT-Transformer & Stage 1 direct ($n=1{,}000$) & 0.662 & 0.659 & 0.655 & 0.669 \\
FT-Transformer & Stage 2 continual ($n=40$) & 0.654 & 0.652 & 0.645 & 0.664 \\
FT-Transformer & Stage 2 direct ($n=40$) & 0.585 & 0.587 & 0.607 & 0.564 \\
ResNet & Stage 1 direct ($n=1{,}000$) & 0.657 & 0.657 & 0.658 & 0.658 \\
ResNet & Stage 2 continual ($n=40$) & 0.646 & 0.659 & 0.687 & 0.603 \\
ResNet & Stage 2 direct ($n=40$) & 0.595 & 0.581 & 0.568 & 0.623 \\
EBM & Stage 1 direct ($n=1{,}000$) & 0.666 & 0.666 & 0.666 & 0.665 \\
EBM & Stage 2 continual ($n=40$) & 0.632 & 0.637 & 0.655 & 0.608 \\
EBM & Stage 2 direct ($n=40$) & 0.607 & 0.603 & 0.598 & 0.615 \\
MHL & Stage 1 direct ($n=1{,}000$) & 0.593 & 0.677 & 0.855 & 0.331 \\
MHL & Stage 2 continual ($n=40$) & 0.589 & 0.685 & 0.891 & 0.287 \\
MHL & Stage 2 direct ($n=40$) & 0.591 & 0.665 & 0.812 & 0.370 \\
\bottomrule
\end{tabular}
\end{table}

\section{Prompt Templates}
\label{app:prompt-templates}

\subsection{Knowledge Probe Prompt}
\label{app:b-1}

Source function: \texttt{get\_knowledge\_probe\_prompt(...)}

\begin{PromptBlock}
You are a medical prior-knowledge assistant. Generate clinical prior knowledge for the given tabular medical features.
Write EVERYTHING in English only.
[Optional if {task_description} is non-empty]
Task description: {task_description}
Outcome/target column: {target}
You MUST return a Markdown table that includes the relationships between the features and the outcome, with exactly these columns:
| Feature | Univariate signal (summary) | Clinical rationale | Suggested threshold | Evidence confidence (high/medium/low) |
Requirements:
- You must provide a suggested threshold (if not applicable, write "no clear threshold" and explain why)
- Clinical rationale should be specific enough to justify rules
- Evidence confidence must be one of: high / medium / low
Feature list (json):
{features_json}
\end{PromptBlock}

\subsection{Initial Rule Generation Prompt}
\label{app:b-2}

Source function: \texttt{get\_rule\_generation\_prompt(...)}

\begin{PromptBlock}
You are a medical rule learning agent. Based on the input information, you will generate a pure-Python classification rule function.
Write EVERYTHING in English only.
Inputs include: univariate summary, medical knowledge table, and metric priority.
[Optional if {task_description} is non-empty]
Task description: {task_description}
{metric_desc}

[Univariate Summary]
{univariate_summary}

[Medical Knowledge Table]
{knowledge_table}

Return STRICT JSON with fields:
- version: "v0"
- error_analysis: the design rationale (in English)
- new_policy_code: full Python function definition; function name MUST be predict_v0
Function signature MUST be: def predict_v0(features: dict) -> int:
Return an integer class label. For binary tasks, return 0/1 matching the dataset label.
You may implement a score-based rule, a direct rule-based decision tree, or any deterministic rule set, as long as it is a pure-Python function.
Each if/elif/else branch MUST include an English comment briefly explaining the medical rationale or design intent.
The rule must be self-contained and use ONLY the Python standard library (no third-party packages).
\end{PromptBlock}

\subsection{Iteration Prompt}
\label{app:b-3}

Source function: \texttt{get\_iteration\_prompt(...)}

\begin{PromptBlock}
You are a medical rule learning agent. Update the current Python-based classification rule based on:
- Current full code (all historical versions)
- This round's training set error analysis
- Iteration trajectory (reasons for previous changes)
- Metric priority
- (If any) degradation warning: cases that regressed from correct to wrong after the last change

[Optional if {task_description} is non-empty]
Task description: {task_description}

{metric_desc}

[Current Full Code]
{current_code}

[Training Error Analysis]
{error_report}

[Iteration Trajectory]
{trajectory}

[Degradation Warning]
{degradation_warning}

Important: if a degradation warning exists, you MUST prioritize fixing regressed cases (previously correct, now wrong) and try not to introduce new regressions.
Each if/elif/else branch MUST include an English comment briefly explaining the medical rationale or design intent.
Extra constraint: do NOT collapse to predicting almost all 1s or 0s.
Minimal-change constraint: only make small adjustments this round (e.g., adjust 1-2 thresholds/weights, or add/remove no more than 2 rules). Do NOT rewrite the whole function.
Return STRICT JSON:
{
  "version": "{next_version}",
  "error_analysis": "...(English)...",
  "new_policy_code": "def predict_...\\n ..."
}
Keep changes minimal and comments clear.
The rule must be self-contained and use ONLY the Python standard library (no third-party packages).
\end{PromptBlock}

\subsection{Continual Learning \texttt{v0} Generation Prompt}
\label{app:b-4}

Source function: \texttt{get\_continuous\_v0\_generation\_prompt(...)}

\begin{PromptBlock}
You are a medical rule learning agent. I have already built a pure-Python classification rule function. However, feature drift has occurred. Based on the following summary of feature drift, together with the provided univariate summary, medical knowledge table, metric priority, and other inputs, update the classification rule function.
Write EVERYTHING in English only.
Inputs include: feature drift summary, updated univariate summary, updated medical knowledge table, metric priority, and the previous final model blueprint.
[Optional if {task_description} is non-empty]
Task description: {task_description}
{metric_desc}

[Feature Drift Summary]
- Dropped columns: {dropped_cols}
- Added columns: {added_cols}
- Renamed columns: {renamed_cols}
- Change note: {change_note}

[Optional if {univariate_summary} is non-empty]
[Updated Univariate Summary]
{univariate_summary}

[Optional if {knowledge_table} is non-empty]
[Updated Medical Knowledge Table]
{knowledge_table}

[Previous Final Model Blueprint]
{blueprint_code}

Return STRICT JSON with fields:
- version: "v0"
- error_analysis: the design rationale (in English)
- new_policy_code: full Python function definition; function name MUST be predict_v0
Function signature MUST be: def predict_v0(features: dict) -> int:
Return an integer class label. For binary tasks, return 0/1 matching the dataset label.
Do not reference dropped features.
You may use added or renamed features if useful.
Each if/elif/else branch MUST include an English comment briefly explaining the medical rationale or design intent.
The rule must be self-contained and use ONLY the Python standard library (no third-party packages).
\end{PromptBlock}

\bibliographystyle{unsrt}
\bibliography{references}

\end{document}